\documentclass[twoside,11pt]{article}

\usepackage{blindtext}

\usepackage[preprint]{jmlr2enew}

\usepackage{microtype}
\usepackage{array,multirow}
\usepackage{graphicx}
\usepackage{subfigure}
\usepackage{booktabs} 
\usepackage{placeins}
\usepackage{tabularx}
\usepackage{comment}

\usepackage{hyperref}
\usepackage{wrapfig}

\usepackage{amsmath}
\usepackage{amssymb}
\usepackage{mathtools}

\usepackage[capitalize,noabbrev]{cleveref}

\newcommand{\D}{\mathcal{D}}
\newcommand{\M}{\mathcal{M}}
\newcommand{\N}{\mathcal{N}}
\newcommand{\E}{\mathbb{E}}

\usepackage{xcolor}
\definecolor{dark-blue}{rgb}{0.15,0.15,0.4}

\hypersetup{
    colorlinks=true,
    linkcolor=blue,
    filecolor=magenta,      
    urlcolor=cyan,
    citecolor=dark-blue,
    pdfpagemode=FullScreen,
}

\usepackage[textsize=tiny]{todonotes}

\usepackage{tikz} 
\usepackage{lipsum}

\newcommand{\dataset}{{\cal D}}

\usepackage{lastpage}
\jmlrheading{23}{2023}{1-\pageref{LastPage}}{12/31; Revised xx/xx}{xx/xx}{21-0000}{Lotfi, Izmailov, Benton, Goldblum, and Wilson}

\ShortHeadings{Bayesian Model Selection, the Marginal Likelihood, and Generalization}{Lotfi, Izmailov, Benton, Goldblum, and Wilson}
\firstpageno{1}

\begin{document}

\title{Bayesian Model Selection, the Marginal Likelihood, and Generalization}

\author{\name Sanae Lotfi \email sl8160@nyu.edu \\
       \addr Center for Data Science, New York University
       \AND
       \name Pavel Izmailov \email pi390@nyu.edu \\
       \addr Courant Institute, New York University
       \AND
       \name Gregory Benton \email gwb260@nyu.edu \\
       \addr Courant Institute, New York University
       \AND
       \name Micah Goldblum \email goldblum@nyu.edu\\
       \addr Center for Data Science, New York University
       \AND
       \name Andrew Gordon Wilson \email andrewgw@cims.nyu.edu \\
       \addr Courant Institute and Center for Data Science, New York University}
\editor{}

\maketitle

\begin{abstract}
How do we compare between hypotheses that are entirely consistent with observations? The marginal likelihood (aka Bayesian evidence), which represents the probability of generating our observations from a prior, provides a distinctive approach to this foundational question, automatically encoding Occam's razor. Although it has been observed that the marginal likelihood can overfit and is sensitive to prior assumptions, its limitations for hyperparameter learning and discrete model comparison have not been thoroughly investigated. We first revisit the appealing properties of the marginal likelihood for learning constraints and hypothesis testing. We then highlight the conceptual and practical issues in using the marginal likelihood as a proxy for generalization. Namely, we show how marginal likelihood can be negatively correlated with generalization, with implications for neural architecture search, and can lead to both underfitting and overfitting in hyperparameter learning. We also re-examine the connection between the marginal likelihood and PAC-Bayes bounds and use this connection to further elucidate the shortcomings of the marginal likelihood for model selection. We provide a partial remedy through a conditional marginal likelihood, which we show is more aligned with generalization, and practically valuable for large-scale hyperparameter learning, such as in deep kernel learning. 
\end{abstract}


\section{Introduction}
\label{sec: intro}

The search for scientific truth is elusive. No matter how consistent a theory may be with all available data, it is always possible to propose an alternative theory that is equally consistent. Moreover, no theory is entirely correct: there will always be missed nuances, or phenomena we have not or cannot measure. To decide between different possible explanations, we heavily rely on a notion of \emph{Occam's razor} --- that the ``simplest'' explanation of data consistent with our observations is most likely to be true. For example, there are alternative theories of gravity to general relativity that are similarly consistent with observations, but general relativity is preferred because of its simplicity and intuitive appeal.

\citet{jeffreys1939theory}, and many follow up works, showed that Occam's razor is not merely an ad-hoc rule of thumb, but a rigorous quantifiable consequence of probability theory. \citet[][Chapter 28]{mackay2003information} arguably makes this point most clearly. Suppose we observe what appears to be a block behind a tree. If we had x-ray vision, perhaps we would see that there are in fact two blocks of equal height standing next to each other.
The two block model can generate many more observations, but as a consequence, has to assign these observations lower probabilities. For what we do observe, the one block hypothesis is significantly more likely (see Figure \ref{fig:mll_pitfalls}(c)), even if we believe each hypothesis is equally likely before we observe the data. This probability of generating a dataset from a prior model is called the \emph{marginal likelihood}, or \emph{Bayesian evidence}. The marginal likelihood is widely applied to hypothesis testing, and \emph{model selection}, where we wish to know which trained model is most likely to provide the best generalization. Marginal likelihood optimization has also been applied with great success for hyperparameter learning, where it is known as \emph{empirical Bayes}, often outperforming cross-validation.

There is a strong polarization in the way marginal likelihood is treated.
Advocates make compelling arguments about its philosophical benefits for hypothesis testing, 
its ability to learn constraints,
and its practical successes, especially in Gaussian process kernel learning --- often embracing the marginal likelihood as a nearly all-encompassing solution to model selection \citep[e.g.,][]{mackay1992thesis, minka2001automatic, rasmussen06, DKL}.
Critics tend to focus narrowly 
on its sensitivity to prior assumptions, without appreciating its many strengths \citep[e.g.,][]{domingos1999role, gelmanbayes2011, gelman2013bayesian}. There is a great need for a more comprehensive exposition, clearly demonstrating the limits of the marginal likelihood, while acknowledging its unique strengths, especially given the rise of the marginal likelihood in deep learning. 

Rather than focus on a specific feature of the marginal likelihood, such as its sensitivity to the prior in isolation, in this paper we aim to fundamentally re-evaluate whether the marginal likelihood is the right metric for predicting the generalization of trained models, and learning hyperparameters.
We argue that it does a good job of prior hypothesis testing, which is exactly aligned with the question it is designed to answer. 
However, we show that the marginal likelihood is only peripherally related to the question of which model we expect to generalize best after training, with significant implications for its use in model selection and hyperparameter learning. 

We first highlight the strengths of the marginal likelihood, and its practical successes, in Section \ref{sec: caseforit}. We then describe several practical and philosophical issues in using the marginal likelihood for selecting between trained models in Section \ref{sec: pitfalls}, and present a conditional marginal likelihood as a partial remedy for these issues. We exemplify these abstract considerations throughout the remainder of the paper, with several significant findings. We show that the marginal likelihood can lead to both underfitting and overfitting in data space, explaining the fundamental mechanisms behind each. 
In Section \ref{sec:la-in-dl}, we discuss practical approximations of the marginal likelihood, given its intractability in the general case. In particular, we discuss the Laplace approximation used for neural architecture search, the variational ELBO, and sampling-based approaches. 
We then highlight the advantages of these approximations, and how their drawbacks affect the relationship between the marginal likelihood and generalization. 
In Section \ref{sec:speed}, we re-examine the relationship between the marginal likelihood and training efficiency, where we show that a conditional marginal likelihood, unlike the marginal likelihood, is correlated with generalization for a range of datasizes. 
In Section \ref{sec: selection}, we demonstrate that the marginal likelihood can be negatively correlated with the generalization of trained neural network architectures. 
In Section \ref{sec: hypers}, we show that the conditional marginal likelihood provides particularly promising performance for deep kernel hyperparameter learning. 
In Section \ref{sec: pac-bayes-lml}, we revisit the connection between the marginal likelihood and PAC-Bayes generalization bounds in theory and practice. We show that while such intuitive and formal connection exists, it does not imply that the marginal likelihood should be used for hyperparameter tuning or model selection.
We also use this connection to understand the pitfalls of the marginal likelihood from a different angle.
We make our code \href{https://github.com/Sanaelotfi/Bayesian_model_comparison}{\underline{available here}}.

This paper extends a \href{https://arxiv.org/abs/2202.11678v2}{\underline{shorter version}} of this work, particularly with additional discussion and experiments regarding approximate inference, PAC-Bayes, and neural architecture search.

\section{Related Work}
\label{sec: related}

As as early as \citet{jeffreys1939theory}, it has been known that the log marginal likelihood (LML) encodes a notion of Occam's razor arising from the principles of probability, providing a foundational approach to hypothesis testing 
\citep{good1968corroboration, good1977explicativity, jaynes1979inference, gull1988bayesian, smith1980bayes, loredo1990laplace, berger1991minimal, jefferys1991sharpening, bayesfactors1995}.
In machine learning, Bayesian model selection was developed and popularized by the pioneering works of David MacKay \citep{mackay1992thesis, mackay1992practical, mackay1992evidence, mackay1992interpolation}. 
These works develop early Bayesian neural networks, and use a Laplace approximation of the LML for neural architecture design, and learning hyperparameters such  as weight-decay \citep{mackay1992practical, mackay1995probable}. 

In addition to the compelling philosophical arguments, the practical success of the marginal likelihood is reason alone to study it closely. For example, LML optimization is now the de facto procedure for kernel learning with Gaussian processes, working much better than other approaches such as standard cross-validation and covariogram fitting, and can be applied in many cases where these standard alternatives are simply intractable  \citep[e.g.,][]{rasmussen06, wilson2014thesis, lloyd2014automatic, DKL}.

Moreover, in variational inference, the evidence lower bound (ELBO) to the LML is often used for automatically setting hyperparameters \citep{hoffman2013stochastic, kingma2013auto, kingma2015variational,alemi2018fixing}. Notably, in variational auto-encoders (VAE), the whole decoder network (often, with millions of parameters) is treated as a model hyperparameter and is trained by maximizing the ELBO \citep{kingma2013}.

Recently, the Laplace approximation (LA) and its use in marginal likelihood model selection has quickly regained popularity in Bayesian deep learning \citep{kirkpatrick2017overcoming, ritter2018scalable, daxberger2021laplace, immer2021scalable, immerouderaa2022deepinv}.
Notably, \citet{immer2021scalable} use a scalable Laplace approximation of the marginal likelihood
to predict which architectures will generalize best, and for automatically setting hyperparameters in deep learning, in the vein of \citet{mackay1992thesis}, but with much larger networks.

\citet{mackay2003information} uses the Laplace approximation to make connections between the marginal likelihood and the minimum description length framework. \citet{mackay1995probable} also notes that structural risk minimization \citep{guyon1992structural} has the same scaling behaviour as the marginal likelihood. 
In recent years, PAC-Bayes \citep[e.g.,][]{alquier2021user} has provided a popular framework for generalization bounds on stochastic networks \citep[e.g.][]{dziugaite2017computing,zhou2018non,lotfi2022pac}.
Notably, \citet{germain2016pac} derive PAC-Bayes bounds that are tightly connected with the marginal likelihood.
We discuss these works in detail in Section \ref{sec: pac-bayes-lml},
where we use the PAC-Bayes bounds to provide further insights into the limitations of the marginal likelihood for model selection and hyperparameter tuning.

Critiques of the marginal likelihood often note its inability to manage improper priors for hypothesis testing, sensitivity to prior assumptions, lack of uncertainty representation over hyperparameters, and its potential misuse in advocating for models with fewer parameters \citep[e.g.,][]{domingos1999role, gelman2013bayesian, gelmanbayes2011, ober2021promises}. To address such issues, \citet{berger1996intrinsic} propose the \emph{intrinsic Bayes factor} to enable Bayesian hypothesis testing with improper priors. Decomposing the LML into a sum over the data, \citet{fong2020marginal} use a similar measure to help reduce sensitivity to prior assumptions when comparing trained models. \citet{lyle2020bayesian} also use this decomposition to suggest that LML is connected to training speed. \citet{rasmussen01} additionally note that the LML operates in function space, and can favour models with many parameters, as long as they do not induce a distribution over functions unlikely to generate the data. 

Our work complements the current understanding of the LML, and has many features that distinguish it from prior work: (1) 
We provide a comprehensive treatment of the strengths and weaknesses of the LML across hypothesis testing, model selection, architecture search, and hyperparameter optimization; (2) While it has been noted that LML model selection can be sensitive to prior specification, we argue that the LML is answering an entirely different question than ``will my trained model provide good generalization?'', even if we have a reasonable prior; (3) We differentiate between LML hypothesis testing of fixed priors, and predicting which trained model will generalize best; 
(4) We also show that LML optimization can lead to \emph{underfitting} or \emph{overfitting} in function space; 
(5) We show the recent characterization in \citet{lyle2020bayesian} that ``models which train faster will obtain a higher LML'' is not generally true, and revisit the connection between LML and training efficiency; 
(6) We show that in modern deep learning, the Laplace LML is not well-suited for architecture search and hyperparameter learning despite its recent use; 
(7) We study a conditional LML (CLML), related to the metrics in \citet{berger1996intrinsic} and \citet{fong2020marginal}, but with a different rationale and application.
We are the first to consider the CLML for hyperparameter learning, model selection for neural networks, approximate inference, and classification. 
We also do not consider prior sensitivity a drawback of the LML, but argue instead that the LML is answering a fundamentally different question than whether a trained model provides good generalization, and contrast this setting with hypothesis testing.
Compared to cross-validation, the CLML can be more scalable and can be conveniently used to learn thousands of hyperparameters.

\section{The Case for the Marginal Likelihood}
\label{sec: caseforit}

While we are primarily focused on exploring the limitations of the marginal likelihood, we emphasize that the marginal likelihood distinctively addresses foundational questions in hypothesis testing and constraint learning.
By encoding a notion of Occam's razor, the marginal likelihood can outperform cross-validation, without intervention and using training data alone. Since we can directly take gradients of the marginal likelihood with respect to hyperparameters on the training data, it can also be applied where standard cross-validation cannot, for computational reasons.

\textbf{Definition.}\quad
The \emph{marginal likelihood} is the probability that we would generate a dataset $\mathcal{D}$ with a model $\mathcal{M}$ if we randomly sample from a prior over its parameters $p(w)$: 
\begin{equation}
\label{eq:lml_definition}
p(\mathcal{D}|\mathcal{M}) = \int p(\mathcal{D}|\mathcal{M},w) p(w | \mathcal{M}) dw.
\end{equation}
It is named the \emph{marginal} likelihood, because it is a likelihood formed from marginalizing parameters $w$. It is also known as the \emph{Bayesian evidence}. Maximizing the marginal likelihood is sometimes referred to as \emph{empirical Bayes}, \emph{type-II maximum likelihood estimation}, or \emph{maximizing the evidence}. 
We can also decompose the marginal likelihood as 
\begin{equation}
    \label{eq:lml_product}
    p(\mathcal{D}|\mathcal{M}) =  \prod^n_i p(\mathcal{D}_i | \mathcal{D}_{<i}, \mathcal{M}),
\end{equation}
where it can equivalently be understood as how good the model is at predicting each data point in sequence given every data point before it.

\textbf{Occam factors.}\quad 
In the definition of the marginal likelihood in Eq.~\eqref{eq:lml_definition}, the argument of the integral is the posterior $p(w|\mathcal{D},\mathcal{M})$ up to a constant of proportionality. If we assume the posterior is relatively concentrated around $\hat{w} = \text{argmax}_w p(w|\mathcal{D},\mathcal{M})$, then we can perform a rectangular approximation of the integral, as the height of the posterior times its width, $\sigma_{w|\mathcal{D}}$, to find
\begin{equation}
    \label{eq:occam_factors}
    p(\mathcal{D} | \mathcal{M} ) \approx p(\mathcal{D}|\hat{w}, \mathcal{M}) \cdot \frac{\sigma_{w|\mathcal{D}}}{\sigma_w},
\end{equation}
where $p(\mathcal{D}|\hat{w}, \mathcal{M})$ is the data fit and $\frac{\sigma_{w|\mathcal{D}}}{\sigma_w}$ is the \emph{Occam factor} --- the width of the posterior over the width of the prior. If the posterior contracts significantly from the prior, there will be a large Occam penalty, leading to a low LML.

\begin{figure*}[t]
\centering
    \begin{tabular}{ccc}
        \hspace{-0.2cm}\includegraphics[height=0.17\textwidth, trim={0 -0.5cm 0 0},clip]{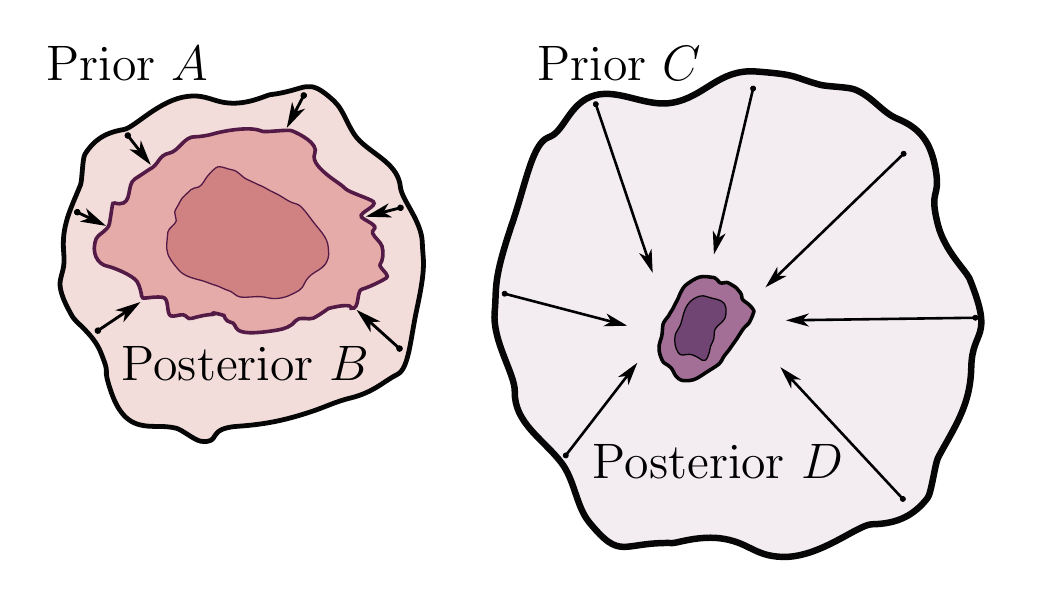}
        &
        \begin{tikzpicture}
            \node (pic1)[right=0.cm] {\includegraphics[height=0.17\textwidth, trim={0 -0.5cm 0 0},clip]{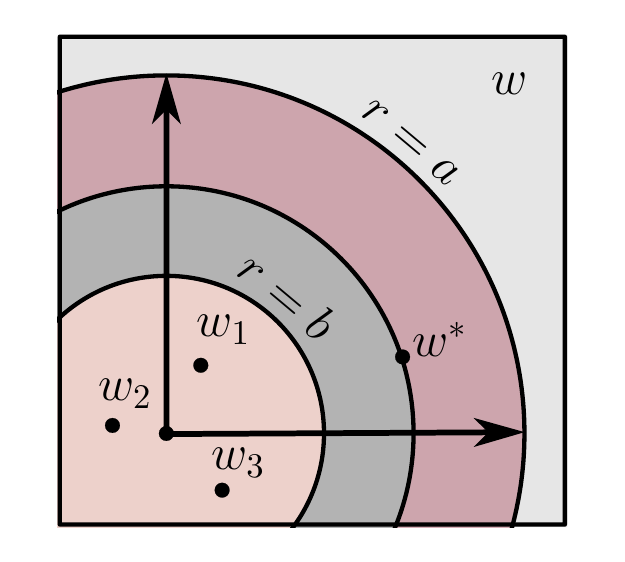}};
            \node (pic2)[right=4.cm] {\includegraphics[height=0.17\textwidth, trim={0 -0.5cm 0 0},clip]{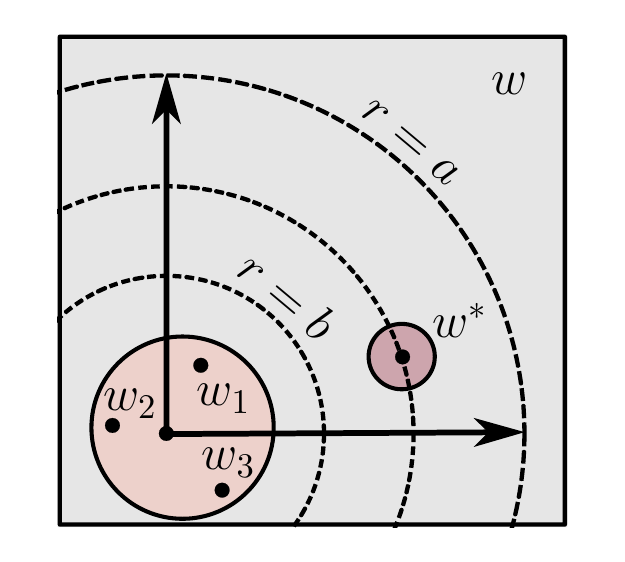}};
            \draw[-latex,thick] ([xshift=-3mm] pic1.east) -- ([xshift=3mm] pic2.west) 
            node[midway,below]{\scriptsize Observe $\D$};
        \end{tikzpicture}
        &
        \hspace{-0.4cm}\includegraphics[height=0.18\textwidth]{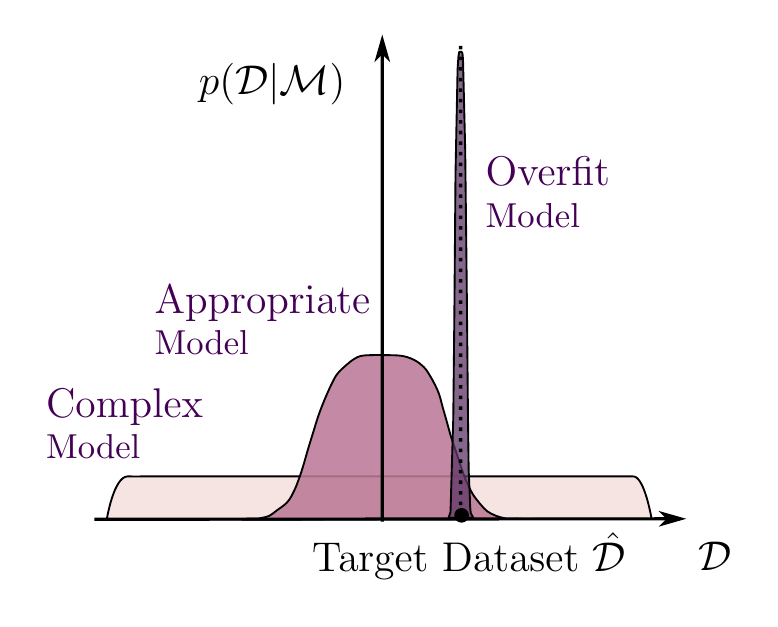}
      \\
        \hspace{-0.2cm}{\small (a) Posterior contraction} &
        \hspace{-0.3cm}{\small \begin{tabular}{c} (b) Marginal likelihood \\ underfitting\end{tabular}} &  
        \hspace{-0.4cm}{\small \begin{tabular}{c} (c) Marginal likelihood \\ overfitting\end{tabular}}
    \\[-0.3cm]
    \end{tabular}
\caption{
\textbf{Pitfalls of marginal likelihood.}
\textbf{(a)}: 
Prior $B$ is vague, but contains easily identifiable solutions and quickly collapses to posterior $D$ after observing a small number of datapoints.
Prior $A$ describes the data better than prior $B$, but posterior $D$ describes the data better than posterior $B$.
The marginal likelihood will prefer model $A$, but model $C$ generalizes better.
\textbf{(b)}: Example of misalignment between marginal likelihood and generalization. 
The marginal likelihood will pick prior scale $b$, and not include the best solution $w^*$, leading to suboptimal generalization performance.
\textbf{(c)}: The complex model spreads its mass thinly on a broad support, while the appropriate model concentrates its mass on a particular class of problems. The overfit model is a $\delta$-distribution on the target dataset $\hat \D$.
}
\label{fig:mll_pitfalls}
\end{figure*}

\textbf{Occam's Razor.}\quad
The marginal likelihood automatically encapsulates a notion of Occam's razor, as in Figure \ref{fig:mll_pitfalls}(c).
If a model can only generate a small number of datasets, it will generate those datasets with high probability, since the marginal likelihood is a normalized probability density.
By the same reasoning, a model which can generate many datasets cannot assign significant probability density to all of them.
For a given dataset, the marginal likelihood will automatically favour the most constrained model that is consistent with the data.
For example, suppose we have $f_1(x,w) = w_1 x$, and $f_2(x,w) = \sum_{i=1}^{100} w_i x^i$, with $p(w) = \mathcal{N}(0,I)$ in both cases, and data given by a straight line with a particular slope.
Both models have parameters consistent with the data, yet the first model is significantly more likely to generate this dataset from its prior over functions.

\textbf{Hypothesis Testing.}\quad
The marginal likelihood provides an elegant mechanism to select between fixed hypotheses, even if each hypothesis is entirely consistent with our observations, and the prior odds of these hypotheses are equal.
For example, in the early twentieth century, it was believed that the correct explanation for the irregularities in Mercury's orbit was either an undiscovered planet, orbital debris, or a modification to Newtonian gravity, but not general relativity.
Since the predictions of general relativity are unable to explain other possible orbital trajectories, and thus easy to falsify, but consistent with Mercury's orbit, \citet{jefferys1991sharpening} show it has a significantly higher marginal likelihood than the alternatives.
We emphasize here we are comparing fixed \emph{prior} hypotheses. We are not interested in how parameters of general relativity update based on orbital data, and then deciding whether the updated general relativity is the correct description of orbital trajectories.

\textbf{Hyperparameter Learning.}\quad
In practice, the LML is often used to learn hyperparameters of the prior to find $\text{argmax}_{\theta} p(\mathcal{D}|\theta)$ where  $p(\mathcal{D}|\theta) = \int p(D|w) p(w|\theta) dw $. 
Gaussian processes (GPs) provide a particularly compelling demonstration of LML hyperparameter learning. The LML does not prefer a small RBF length-scale that would optimize the data fit. 
Instead, as we show empirically in Figure \ref{fig:gp_results} (Appendix), the LML chooses a value that would make the distribution over functions likely to generate the training data. 
We note that the LML can be used to learn many such kernel parameters \citep{rasmussen06, wilsonadams2013, DKL}. Since we can take gradients of the LML with respect to these hypers using only training data, 
the LML can also be used where cross-validation would suffer from a curse of dimensionality. 

\textbf{Constraint Learning.}\quad
Typical learning objectives like maximum likelihood are never incentivized to select for constraints, because a constrained model will be a special case of a more flexible model that is more free to increase likelihood.
The LML, on the other hand, can provide a consistent estimator for such constraints, automatically selecting the most constrained solution that fits the data, and collapsing to the true value of the constraint in the limit of infinite observations, from training data alone. Bayesian PCA is a clear example of LML constraint learning  \citep{minka2001automatic}.
Suppose the data are generated from a linear subspace, plus noise. While maximum likelihood always selects for the largest possible subspace dimensionality, and cross-validation tends to be cumbersome and inaccurate, the LML provides a consistent and practically effective estimator for the true dimensionality. Another clear example is in automatically learning symmetries, such as rotation invariance \citep{van2018learning,immerouderaa2022deepinv}.

\section{Pitfalls of the Marginal Likelihood}
\label{sec: pitfalls}

We now discuss general conceptual challenges in working with the marginal
likelihood, and present the conditional marginal likelihood as a partial remedy. The remainder of this paper concretely exemplifies each of these challenges.

\subsection{Marginal Likelihood is not Generalization}
\label{sec: notgen}

The marginal likelihood answers the question \emph{``what is the probability that a prior model generated the training data?''}. This question is subtly different from asking \emph{``how likely is the posterior, conditioned on the training data, to have generated withheld points drawn from the same distribution?''}.
Although the marginal likelihood is often used as a proxy for generalization \citep[e.g.][]{mackay1992thesis,  immer2021scalable, daxberger2021laplace}, it is the latter question we wish to answer in deciding whether a model will provide good generalization performance.

Indeed, if after observing data, prior $A$ leads to posterior $B$, and prior $C$ leads to posterior $D$, it can be the case that the same data are \emph{less} probable under $B$ than $D$, and also that $D$ provides better generalization on fresh points from the same distribution, \emph{even if the prior $A$ explains the data better than $C$}.
Consider, for example, the situation where we have a prior over a diffuse set of solutions which are easily identifiable from the data.
We will then observe significant posterior contraction, as many of these solutions provide poor likelihood.
While the marginal likelihood will be poor, the posterior could be perfectly reasonable for making predictions:
in the product decomposition of the marginal likelihood in Section \ref{sec: caseforit}, the first terms will have low probability density, even if the posterior updates quickly to become a good description of the data.
A different prior, which allocates significant mass to moderately consistent solutions, could then give rise to a much higher marginal likelihood, but a posterior which provides poorer generalization. We illustrate this effect in Figure \ref{fig:mll_pitfalls}(a) and provide concrete examples in Section \ref{sec:speed}.

There are several ways of understanding why the marginal likelihood will be poor in this instance:
(1) the diffuse prior is unlikely to generate the data we observe, since it allocates significant mass to generating other datasets;
(2) we pay a significant Occam factor penalty, which is the width of the posterior over the width of the prior, in the posterior contraction;
(3) in the product decomposition of the marginal likelihood in Section \ref{sec: caseforit}, the first terms will have low probability density, even if the posterior updates quickly to become a good description of the data.

\textbf{Model Selection.}\quad In hypothesis testing, our interest is in evaluating priors, whereas in \emph{model selection} we wish to evaluate posteriors. In other words, in model selection we are not interested in a fixed hypothesis class $A$ corresponding to a prior (such as the theory of general relativity in the example of Section~\ref{sec: caseforit}), but instead the posterior $B$ that arises when $A$ is combined with data. Marginal likelihood is answering the question most pertinent to hypothesis testing, but is not generally well-aligned with model selection. We provide several examples in Sections \ref{sec:speed}, \ref{sec: selection}.

\subsection{Marginal Likelihood Optimization and Overfitting}
\label{sec: mltraining}

Marginal likelihood optimization for hyperparameter learning, also known as \emph{type-II maximum likelihood} or \textit{empirical Bayes}, is a special case of model selection.
In this setting, we are typically comparing between many models --- often a continuous spectrum of models --- corresponding to different hyperparameter settings. 
In practice the marginal likelihood can be effective for tuning hyperparameters, as discussed in Section~\ref{sec: caseforit}. 
However, marginal likelihood optimization can be prone to both underfitting and overfitting.

\textbf{Overfitting by ignoring uncertainty.}\quad We can overfit the marginal likelihood, as we can overfit the likelihood.
Indeed, a likelihood for one model can always be seen as a marginal likelihood for another model.
For example, suppose we include 
in our search space 
a prior model concentrated around a severely overfit maximum likelihood solution.
Such a model would be ``simple'' in that it is extremely constrained --- it can essentially only generate the dataset under consideration --- and would thus achieve high marginal likelihood, but would provide poor generalization (Figure \ref{fig:mll_pitfalls}(c)).

\begin{figure}[t]
\centering
\includegraphics[height=0.22\textwidth]{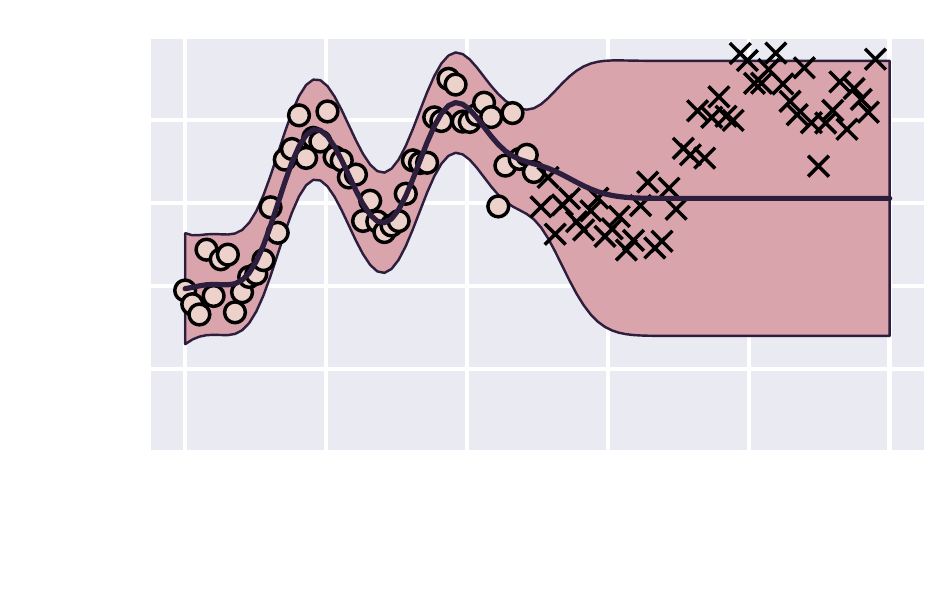}
\quad
\includegraphics[height=0.22\textwidth]{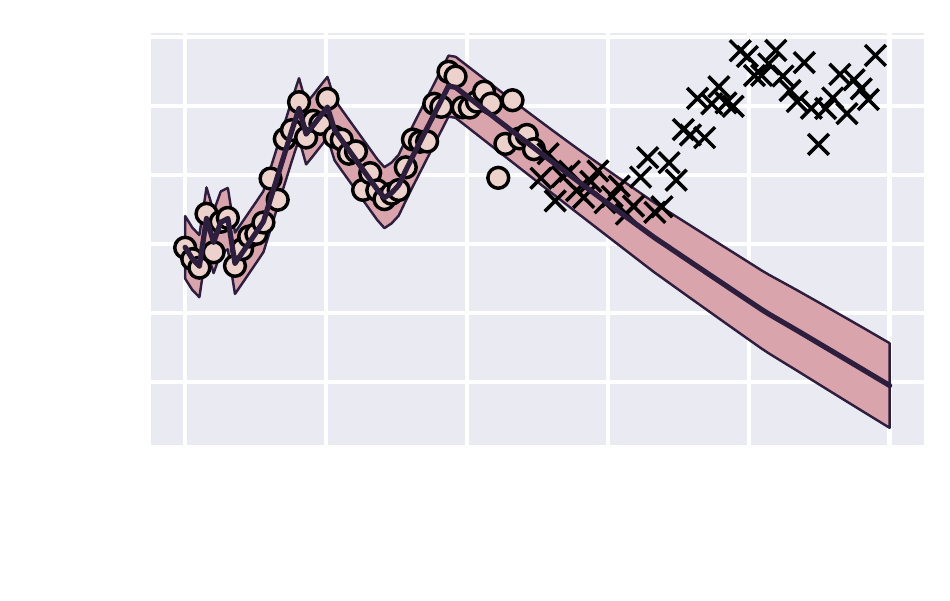}
\vspace{-0.5cm}
\caption{
\textbf{LML overfitting in Gaussian Processes.}
\textbf{Left:} A fit of a GP regression model with a constant prior mean.
\textbf{Right:} a prior mean parameterized with an MLP and trained via marginal likelihood optimization.
Train data is shown with circles, test data is shown with crosses and the shaded region visualizes the $2\sigma$-predictive region of the GP.
Given enough flexibility with the prior mean, the marginal likelihood overfits the data, providing poor overconfident predictions outside of the train region.
}
\label{fig:lml_overfitting_example_main}
\end{figure}

As an example,
we parameterize the mean function in an RBF GP with a small multi-layer perceptron (MLP) and learn the parameters of the MLP by optimizing the LML.
We show the results in Figure \ref{fig:lml_overfitting_example_main}, where the learned mean function overfits the train data, leading to poor and overconfident predictions outside of the train region. We note that the mean of a GP does not appear in the Occam factor of the marginal likelihood. Therefore the Occam factor does not directly influence neural network hyperparameter learning in this instance, which is different from deep kernel learning \citep{wilson2016deep}. We provide additional details in Appendix~\ref{sec:app-overfit-gp}.

While it may not appear surprising that we can overfit the marginal likelihood,
the narratives relating the marginal likelihood to Occam's razor often give the impression that it is safe to expand our model search, and that by favouring a ``constrained model'', we are protected from conventional overfitting.
For example, \citet{iwata2017improving} proposed a model analogous to the example in Figure \ref{fig:lml_overfitting_example_main}.
where a neural network serves as a mean function of a GP and argued that
``since the proposed method is based on Bayesian inference, it can help alleviate overfitting".
Furthermore, \citet[][Chapter 3.4]{mackay1992thesis} argues that if the correlation between the marginal likelihood and generalization is poor for a set of models under consideration, then we should expand our search to find new model structures that can achieve better marginal likelihood.
While a mismatch between generalization and marginal likelihood can indicate that we should revisit our modelling assumptions, this advice could easily lead to overfitting.

\textbf{Underfitting in hyperparameter selection.}\quad
The above example involves overfitting that arises by ignoring uncertainty. The marginal likelihood also has a bias towards underfitting. This bias arises because supporting a good solution could involve also supporting many solutions that are unlikely to generate the training data from the prior. As an example, consider a zero-centred Gaussian prior on a set of parameters, $p(w) = \N(0,\sigma^2 I)$.
Now suppose the parameters $w^{\star}$ that provide the best generalization have large norm, $\|w^{\star}\|$, but there are several settings of the parameters that provide moderate fits to the data with smaller norms $b \ll \|w^{\star}\|$.
Further suppose that parameters with norms $b< \|w \| < \|w^\star\|$ provide very poor fits to the data. The marginal likelihood will not favour a large value of $\sigma$ that makes $w^{\star}$ likely under the prior --- even though such a value could lead to a posterior with much better generalization, as in Figure \ref{fig:mll_pitfalls}(b). With more data, the likelihood signal for $w^{\star}$ will dominate, and the underfitting bias disappears. 

\subsection{The Conditional Marginal Likelihood}
\label{sec:conditional_mll}

Using the product decomposition of the marginal likelihood 
in Eq.~\eqref{eq:lml_product}, 
we can write the LML as
\begin{equation}
    \label{eq:mll_expanded}
    \log p(\D \vert \M) = 
    \sum_{i=1}^n \log p(\D_{i} \vert \D_{<i}, \M).
\end{equation}
Each term $\log p(\D_{i} \vert \D_{<i}, \M)$ is the predictive log-likelihood of the data point $\D_i$ under the Bayesian model average after observing the data $\D_{<i}$.
The terms for $i$ close to $n$ are clearly indicative of generalization of the model to new test data:
we train on the available data, and test on the remaining, unseen data.
On the other hand, the terms corresponding to small $i$ have an equally large effect on the marginal likelihood, but may have little to do with generalization.

Inspired by the reasoning above, we consider the \textit{conditional log marginal likelihood (CLML)}:
\begin{equation}
    \label{eq:clml}
    \sum_{i=m}^n \log p(\D_{i} \vert \D_{<i}, \M) = \log p(\D_{\ge m} \vert \D_{<m}),
\end{equation}
where $m \in \{1, \ldots, n\}$ is the cut-off number, and $\D_{\ge m}$ is the set of datapoints $\D_m, \ldots, \D_n$.
In CLML, we simply drop the first $m-1$ terms of the LML decomposition,
to obtain a metric that is more aligned with generalization.
In Appendix \ref{sec:app_clml_details}, we provide further details on the CLML, including a permutation-invariant version, and study how performance varies with the choice of $m$ in Figure \ref{fig:dkl-m} (Appendix~\ref{sec:ablation-clml}). We note the CML
can be written as $\int p(\mathcal{D}_{m:n} | w) p(w | \mathcal{D}_{1:m-1}) dw$, and thus can be more easily estimated by Monte
Carlo sampling than the LML, since samples from the posterior over $m-1$ points will typically have much greater likelihood
than samples from the prior (see Section \ref{sec:lml_sampling} for a discussion of sampling-based estimates of the LML).

Variants of the CLML were considered in \citet{berger1996intrinsic} as an \emph{intrinsic Bayes factor} for handling improper uniform priors in hypothesis testing, and \citet{fong2020marginal} to show a connection with cross-validation and reduce the sensitivity to the prior. Our rationale and applications are different, motivated by understanding how the marginal likelihood can be fundamentally misaligned with generalization. We do not consider prior sensitivity a deficiency of the marginal likelihood, since the marginal likelihood is evaluating the probability the data were generated from the prior. We also are the first to consider the CLML for neural architecture comparison, hyperparameter learning, approximate inference, and transfer learning.
We expect the CLML to address the issues we have presented in this section, with the exception of overfitting, since CLML optimization is still fitting to withheld points.
For hyperparameter optimization, we expect the CLML to be at its best relative to the LML for small datasets.
As in Figure \ref{fig:mll_pitfalls}(b), the LML suffers because it has to assign mass to parameters that are unlikely to generate the data in order to reach parameters that are likely to generate the data.
But as we get more data, the likelihood signal for the good parameters becomes overwhelming, and the marginal likelihood selects a reasonable value.
Even for small datasets, the CLML is more free to select parameters that provide good generalization, since it is based on the posterior $p(w \vert \D_{<m})$ that is re-centred from the prior, as shown in Figure \ref{fig:mll_pitfalls}(b).

\subsection{Marginal Likelihood is Not Aligned with Posterior Model Averaging}
\label{sec:lml_vs_bma}

In Bayesian inference, we are concerned with the performance of the \emph{Bayesian model average (BMA)} in which we integrate out the parameters according to the posterior, to form the \emph{posterior predictive} distribution:
\begin{equation}
    \label{eq:bma}
    p(d^* \vert \dataset, \M) =
    \int_{w} p(d^* \vert w) p(w \vert \dataset) dw,
\end{equation}
where $d^*$ is a test datapoint. In other words, rather than use a single setting of parameters, we \emph{combine} the predictions of models corresponding to every setting of parameters, weighting these models by their posterior probabilities. 

The marginal likelihood, according to its definition in Eq.~\eqref{eq:lml_definition}, measures the expected performance of the prior model average on the training data, i.e. it integrates out the parameters according to the prior.
It is thus natural to assume that the marginal likelihood is closely connected to the BMA performance.
Indeed, \citet{mackay1992thesis} argues that ``the evidence is a measure of plausibility of the entire posterior ensemble''.

However, \emph{the marginal likelihood is not aligned with posterior model averaging}. 
Consider the following representation of the marginal likelihood, using the standard derivation of the evidence lower bound (ELBO, see Section \ref{sec:app_vi}):
\begin{equation}
\begin{split}
    \label{eq:lml_elbo_derivation}
    \log p(\dataset \vert \M) =
    \log p(\dataset)
    &= 
    \int_{w}  \log \left [ p(\dataset) \right ] p(w \vert \dataset) dw   =
    \\
    \int_{w}  \log \left [ 
    \frac{ p(w, \dataset)}{p(w \vert \dataset)} 
    \right ] p(w \vert \dataset) dw &=
    \int_{w}  \log \left [ 
    \frac{ p(\dataset \vert w) p(w)}{p(w \vert \dataset)} 
    \right ] p(w \vert \dataset) dw =
    \\
    \int_{w}  \log \left [ 
    p(\dataset \vert w) 
    \right ] p(w \vert \dataset) dw
    &-
    \int_{w}  \log \left [ 
    \frac{p(w \vert \dataset)}{p(w)}
    \right ] p(w \vert \dataset) dw = 
    \\
    \underbrace{\mathbb{E}_{w \sim p(w \vert \dataset)}\big[\log p(\dataset \vert w)\big]}_{\text{expected sample likelihood}}
    &-
    \underbrace{\mathbb{KL}(p(w \vert \dataset) \vert\vert p(w))}_{\text{Occam factor}},
\end{split}
\end{equation}
where $\mathbb{KL}$ represents the Kullback–Leibler divergence.
From Eq.~\eqref{eq:lml_elbo_derivation}, we can see that the marginal likelihood is in fact more closely related to the average performance (train likelihood) of individual samples from the posterior, and not the Bayesian model average.

This discrepancy is practically important for several reasons.  
First, those using the marginal likelihood will typically be using the posterior predictive for making predictions, and unconcerned with the average performance of individual posterior samples. Second, the discrepancy between these measures is especially pronounced in deep learning.

For example, \citet{wilson2020bayesian} argue that Bayesian marginalization is especially useful in flexible models such as Bayesian neural networks, where different parameters, especially across different modes of the posterior, can correspond to functionally different solutions, so that the ensemble of these diverse solutions provides strong performance.
However, this functional diversity does not affect the marginal likelihood in Eq.~\eqref{eq:lml_elbo_derivation}, as the marginal likelihood is only concerned with the average performance of a random sample from the posterior and the degree of posterior contraction.

In particular, using a prior that only provides support for a single mode of the BNN posterior may significantly affect the BMA performance, as it would limit the functional diversity of the posterior samples, but it would not hurt the marginal likelihood as long as the average training likelihood of a posterior sample within that mode is similar to the average sample likelihood across the full posterior.
We provide an illustration of this behaviour in Figure \ref{fig:periodic-laplacem}, where the marginal likelihood has no preference between a prior that leads to a unimodal posterior, and a prior that leads to a highly multimodal posterior.
We discuss this example in more detail in the next section on approximations of the marginal likelihood.

\section{Marginal Likelihood Approximations}
\label{sec:la-in-dl}
\label{sec:lmlapprox}

Outside of a few special cases, such as Gaussian process regression, the marginal likelihood is intractable.
Because  the marginal likelihood is integrating with respect to the prior, and we thus cannot effectively perform simple Monte Carlo, it is also harder to approximate than the posterior predictive distribution.
Moreover, modern neural networks contain millions of parameters, leaving few practical options. 
In this section, we discuss several strategies for approximating the marginal likelihood, including the Laplace approximation, variational ELBO, and sampling-based methods, and their limitations.
For an extensive review of how to compute approximations to the marginal likelihood, see \citet{llorente2020marginal}.

\subsection{Laplace Approximation}
\label{sec:laplace_approx}

The \textit{Laplace approximation} (LA) for model selection in Bayesian neural networks was originally proposed by \citet{mackay1992thesis},
and has recently seen a resurgence of popularity \citep[e.g.][]{immer2021scalable}. Moreover, several generalization metrics, such as the Bayesian Information Criterion (BIC), can be derived by further approximating the Laplace approximate marginal likelihood \citep{bishop06}.

The Laplace approximation represents the parameter posterior with a Gaussian distribution centred at a local optimum of the posterior aka the maximum a posteriori (MAP) solution,
$w_{\text{MAP}}$, with covariance matrix $\Sigma$ given by the inverse Hessian at $w_{\text{MAP}}$: 
\begin{equation}
\label{eq:laplace_distr}
q(w) = \mathcal{N}(w_{\text{MAP}}, \Sigma),\quad
\Sigma^{-1} = \nabla^2_w \log \left(p(\dataset \vert w) p(w)\right)~
\big\vert_{w = w_{\text{MAP}}}.
\end{equation}
The covariance matrix $\Sigma$ captures the sharpness of the posterior.
The marginal likelihood estimate \citep[see e.g.][]{bishop06} is then given by
\begin{equation}
\label{eq:laplace_approx}
\log p(\dataset \vert \mathcal{M}) \approx
\log p(\dataset \vert w_{\text{MAP}}) 
+ \log p(w_{\text{MAP}}) 
+ \frac D 2 \log (2 \pi) + \frac 1 2 \log\left ( \det \Sigma \right ),
\end{equation}
where $D$ is the dimension of the parameter vector $w$.

 \begin{figure}[t]
 \centering
     \includegraphics[height=0.22\textwidth]{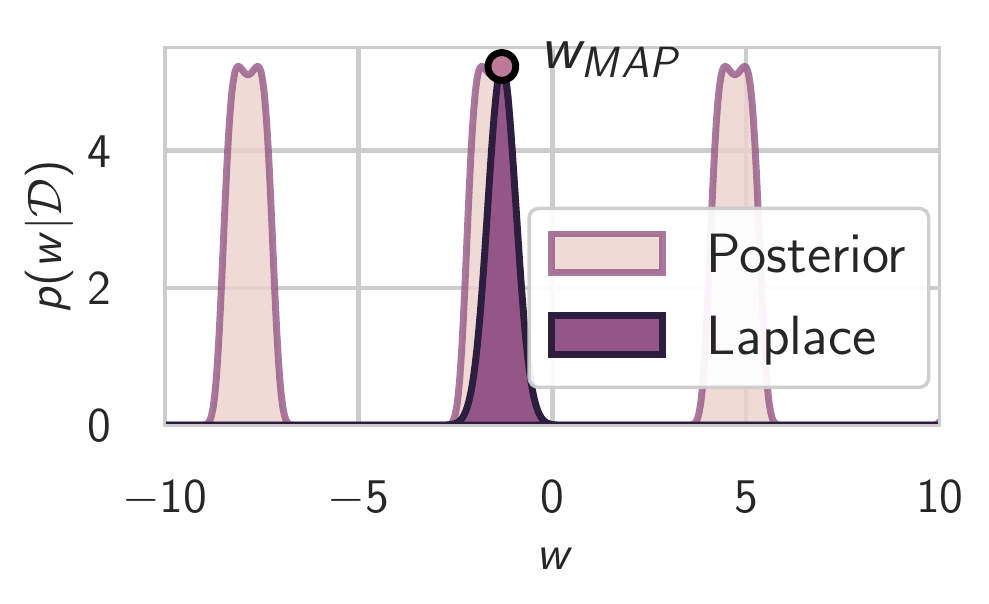}
     \quad\quad
     \includegraphics[height=0.22\textwidth]{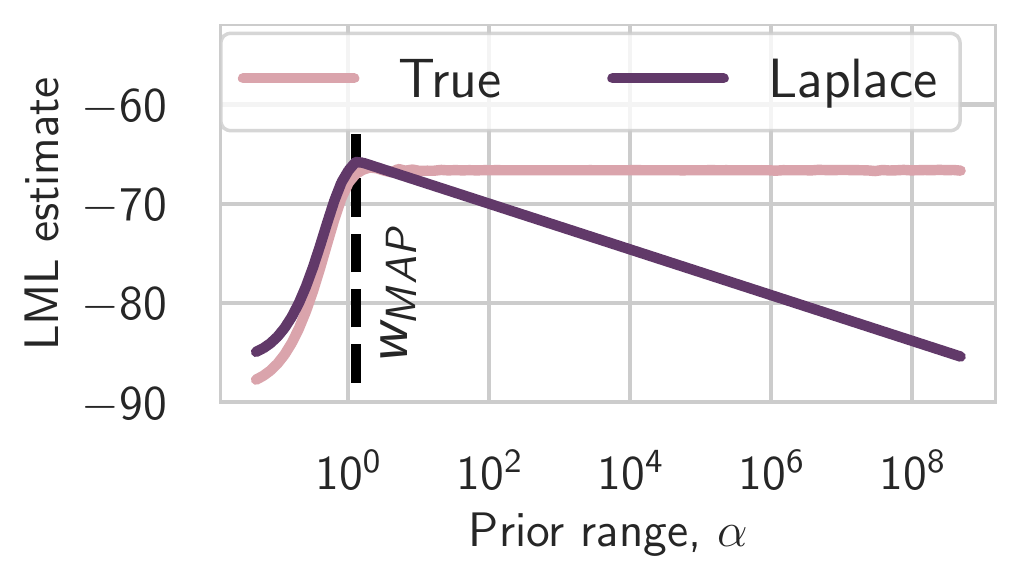}
     \vspace{-0.3cm}
 \caption{
 \textbf{Laplace approximation.}
 Model $x \sim \N(\sin(w), 1)$ with a uniform prior $w \sim U[-\alpha, \alpha]$.
 \textbf{(Left):} Posterior density and a Laplace approximation to the posterior (scaled for visualization);
 \textbf{(Right):} True marginal likelihood and the Laplace estimate as a function of $\alpha$.
 As Laplace only captures a single mode, the Laplace estimate of marginal likelihood decreases linearly with $\alpha$
 while the true marginal likelihood is roughly constant.
 }
 \label{fig:periodic-laplacem}
 \end{figure}

\textbf{Drawbacks of the Laplace approximation.}\quad
The actual posterior distribution for a modern neural network is highly multimodal.
By representing only a single mode, the Laplace approximation provides a poor representation of the true Occam factor in Eq.~\eqref{eq:occam_factors}, which is the posterior volume divided by the prior volume.
As a consequence, the Laplace marginal likelihood will overly penalize diffuse priors that capture multiple reasonable parameter settings across different modes. We provide an example in Figure~\ref{fig:periodic-laplacem}.

We generate data from $x \sim \N(\sin(w), 1)$ with uniform prior $w \sim U[-\alpha, \alpha]$, then estimate the posterior on $w$ and evaluate the marginal likelihood to estimate the parameter $\alpha$.
The posterior is periodic with a period of $2\pi$. 
Consequently, as we increase $\alpha$, the marginal likelihood will be roughly constant for $\alpha > w_{MAP}$, where $w_{MAP}$ is the lowest norm maximum a posteriori solution, as the ratio of the posterior volume to the prior volume (Occam factor) is roughly constant in this regime.
We visualize the posterior and the true LML in Figure~\ref{fig:periodic-laplacem}.
However, the Laplace approximation only captures a single mode of the posterior, and thus greatly underestimates the posterior volume.
As a result, the Laplace marginal likelihood estimate decreases linearly with $\alpha$.
This toy example shows that Laplace can be problematic for tuning the prior scale in Bayesian neural networks, where covering multiple diverse modes is 
beneficial for generalization.

There are several additional drawbacks to the Laplace approximation:
\begin{itemize}
\item The Laplace approximation is \emph{highly local}: it only depends on the value and curvature of the unnormalized posterior log-density $\log(p(\dataset \vert w) p(w))$ at the MAP solution. The curvature at that point may describe the structure of the posterior poorly, even within a single basin of attraction, causing the Laplace approximation to differ dramatically from the true marginal likelihood.
\item In practice, computing the covariance matrix $\Sigma \in \mathbb{R}^{D\times D}$ in Eq.~\eqref{eq:laplace_distr} is intractable for large models such as Bayesian neural networks, as it amounts to computing and inverting the $D \times D$ Hessian matrix of the loss function.
Consequently, practical versions of the Laplace approximation construct approximations to $\Sigma$, e.g. in diagonal \citep{mackay1992practical, kirkpatrick2017overcoming} or 
block-Kronecker factorized (KFAC)
\citep{ritter2018scalable, martens2015optimizing}
form.
These approximations further separate the Laplace estimates of the marginal likelihood from the true marginal likelihood.

\item As evident from Eq. \eqref{eq:laplace_approx}, the Laplace approximation of the marginal likelihood also highly penalizes models with many parameters $D$, even though such models could be simple \citep{maddox2020rethinking}. Unlike the standard marginal likelihood, which operates purely on the properties of functions, the marginal likelihood is sensitive to the number of parameters, and is not invariant to (non-linear) reparametrization. Recent work aims to help mitigate this issue \citep{antoran2022adapting}.
\end{itemize}

In Sections~\ref{sec:speed} and \ref{sec:cifar-exps} we show examples of misalignment between the Laplace marginal likelihood and generalization in large Bayesian neural networks.

\textbf{Information criteria.} \quad
While the Laplace approximation provides a relatively cheap estimate of the marginal likelihood, it still requires estimating the Hessian of the posterior density, which may be computationally challenging. 
We can further simplify the approximation in Eq.~\eqref{eq:laplace_approx} by dropping all the terms 
which do not scale with the number of datapoints $n$,
arriving at the Bayesian information criterion (BIC) \citep{schwarz1978estimating}:
\begin{equation}
    \label{eq:bic}
    \log p(\dataset \vert \mathcal{M}) \approx
\log p(\dataset \vert w_{\text{MLE}}) 
- \frac D 2 \log n,
\end{equation}
where $n$ is the number of datapoints in $\dataset$ and $w_{\text{MLE}}$ is the maximum likelihood estimate (MLE) of the parameter $w$, which replaces the MAP solution.
For a detailed derivation, see 
Chapter 4.4.1 of \citet{bishop2006pattern}.
The BIC is cheap and easy to compute, compared to the other approximations of the marginal likelihood.
However, it is also a crude approximation, as it removes all information about the model except for the number of parameters $D$ and the value of the maximum likelihood, completely ignoring the prior and the amount of posterior contraction. In practice, BIC tends to be more dominated by the maximum likelihood term than other more faithful marginal likelihood approximations, causing it to prefer overly unconstrained models \citep[e.g.,][]{minka2001automatic}.
Other related information criteria include AIC \citep{akaike1974new}, DIC \citep{spiegelhalter2002bayesian}, and WAIC \citep{watanabe2010asymptotic}.
For a detailed discussion of these criteria, see e.g., \citet{gelman2014understanding}.

\subsection{Variational Inference and ELBO}
\label{sec:app_vi}

In variational inference (VI), the evidence lower bound (ELBO), a lower bound on log-marginal likelihood, is often used for automatically setting hyperparameters \citep{hoffman2013stochastic, kingma2013auto, kingma2015variational,alemi2018fixing}.
In variational auto-encoders (VAE), the whole decoder network (often, with millions of parameters) is treated as a model hyper-parameter and is trained by maximizing the ELBO \citep{kingma2013}.

The ELBO is given by
\begin{equation}
    \label{eq:elbo}
    \log p(\D \vert \M) \ge
    \underbrace{\mathbb{E}_{q(w)} \log p(\D \vert w)}_{\text{data fit}} - 
    \underbrace{\mathbb{KL}(q(w) \vert\vert p(w))}_{\text{complexity penalty}} \,,
\end{equation}
where $q(w)$ is an approximate posterior. Note that the ELBO generalizes the decomposition of marginal likelihood in Eq.~\eqref{eq:lml_elbo_derivation}:
the inequality in Eq.~\eqref{eq:elbo} becomes an equality if $q(w) = p(w \vert \dataset)$.

In VI for Bayesian neural networks, the posterior is often approximated with a unimodal Gaussian distribution with a diagonal covariance matrix.
For a complex model, the ELBO will suffer from some of the same drawbacks described in Section \ref{sec:laplace_approx} and the example in Figure \ref{fig:periodic-laplacem}.
However, the ELBO is not as locally defined as the Laplace approximation, as it takes into account the average performance of the samples from the posterior and not just the local curvature of the posterior at the MAP solution.
Consequently, the ELBO can be preferable for models with highly irregular posteriors. 
Moreover, the ELBO in principle allows for non-Gaussian posterior approximations \citep[e.g.][]{rezende2015variational}, making it more flexible than the Laplace approximation. However, the KL term can be exactly evaluated if the prior $p(w)$ and the approximate posterior $q(w)$ are both Gaussian, and must typically be approximated otherwise. Similarly, the ELBO is in principle invariant to reparametrization, unlike Laplace, but in practice one often works with a parametrization where $q(w)$ and $p(w)$ are Gaussian to retain tractability.

On the downside, the ELBO generally requires multiple epochs of gradient-based optimization to find the optimal variational distribution $q$, while the Laplace approximation can be computed as a simple post-processing step for any pretrained model. Moreover, optimizing the ELBO can generally suffer from the same overfitting behaviour as the marginal likelihood in general (see Section \ref{sec: mltraining}).
Indeed, if we can set the prior $p(w)$ to be highly concentrated on a solution that is overfit to the training data, e.g. by tuning the mean of the prior to fit the data as we did in the example in Figure~\ref{fig:lml_overfitting_example_main}, we can set the posterior to match the prior $q(w) = p(w)$ achieving very low ELBO, without improving generalization.

\subsection{Sampling-Based Methods}
\label{sec:lml_sampling}

Another important group of methods for estimating the marginal likelihood are based on sampling.
In the likelihood weighting approach, we form a simple Monte Carlo approximation to the integral in Eq.~\eqref{eq:lml_definition}:
\begin{equation}
    \label{eq:likweight}
    p(\dataset \vert \mathcal{M})
    =
    \mathbb{E}_{w \sim p(w)}
    p(\dataset \vert w)
    \approx
    \frac 1 m
    \sum_{i=1}^m p(\dataset \vert w_i),
    \quad
    w_i \sim p(w),~i=1,\ldots, m.
\end{equation}
While Eq.~\eqref{eq:likweight} provides an unbiased estimate for the marginal likelihood, its variance can be very high.
Indeed, for complex models such as Bayesian neural networks, we are unlikely to encounter parameters that are consistent with the data by randomly sampling from the prior with a computationally tractable number of samples.
Consequently, we will not achieve a meaningful estimate of the marginal likelihood.

In order to reduce the variance of the simple Monte Carlo estimate in  Eq.~\eqref{eq:likweight}, we can use importance sampling, where the samples come from a \textit{proposal distribution} $q(w)$, rather than the prior.
Specifically, in the simple importance sampling approach, the marginal likelihood is estimated as
\begin{equation}
    \label{eq:sis}
    p(\dataset \vert \mathcal{M})
    =
    \int \frac{p(\dataset \vert w) p(w)}{q(w)} q(w) dw
    = 
    \mathbb{E}_{w \sim q(w)} 
    \frac{p(\dataset \vert w) p(w)}{q(w)} 
    \approx
    \frac 1 m
    \sum_{i=1}^m \frac{p(\dataset \vert w_i) p(w_i)}{q(w_i)},
\end{equation}
where $\{w_i\}_{i=1}^m$ are sampled from an arbitrary proposal distribution $q$. In particular, if we use the true posterior as the proposal distribution $q(w) = p(w \vert \dataset) = \frac{p(\dataset \vert w) p(w)}{p(\dataset \vert \M)}$, we have 
$\frac{p(\dataset \vert w_i) p(w_i)}{q(w_i)} = \frac{p(\dataset \vert w_i) p(w_i)}{p(w_i \vert \dataset)} = p(\dataset \vert \M)$. 
Generally, we do not have access to the posterior in a closed-form, so we have to use approximations to the posterior in place of the proposal distribution $q(w)$, retaining a high variance of the LML estimate.

Multiple approaches that aim to reduce the variance of the sampling-based estimates of the marginal likelihood have been developed.
\citet{llorente2020marginal} provide an extensive discussion of many of these methods.
Notably, annealed importance sampling (AIS) \citep{neal2001annealed} constructs a sequence of distributions transitioning from the prior to the posterior so that the difference between each consecutive pair of distributions is not very stark.
\citet{grosse2015sandwiching} derive both lower and upper bounds on the marginal likelihood based on AIS, making it possible to guarantee the accuracy of the estimates.

While AIS and related ideas provide an improvement over the simple Monte Carlo in Eq.~\eqref{eq:likweight}, these approaches are still challenging to apply to large models such as Bayesian neural networks.
In particular, these methods typically require full gradient evaluations in order to perform Metropolis-Hastings correction steps, and using stochastic gradients is an open problem \citep{zhang2021differentiable}.
Moreover, these methods are generally not differentiable, and do not provide an estimate of the gradient of the marginal likelihood with respect to the prior parameters.
This limitation prevents the sampling-based methods from being generally used for hyperparameter learning, which is a common practice with the Laplace and variational approximations.
Several works attempt to address this limitation
\citep[e.g.][]{tomczak2020marginal, zhang2021differentiable}.
However, in general sampling-based approaches are yet to be applied successfully to estimating and optimizing the marginal likelihood in high-dimensional large-scale Bayesian models containing millions of parameters, such as Bayesian neural networks.

\begin{figure*}[t]
\centering
\begin{tabular}{ccc}
        \includegraphics[height=0.25
        \textwidth]{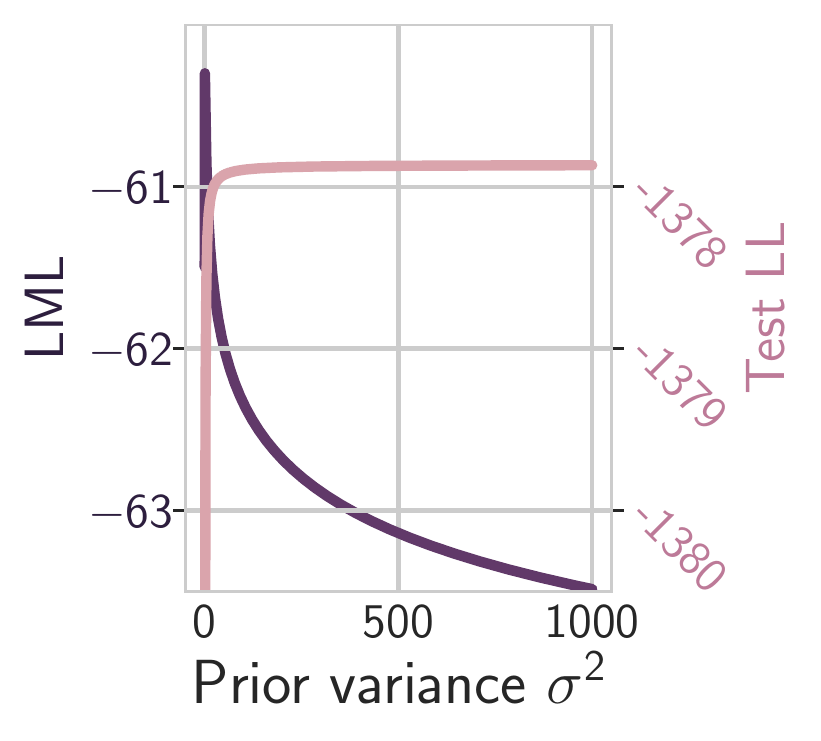}
        &
        \includegraphics[height=0.25
        \textwidth]{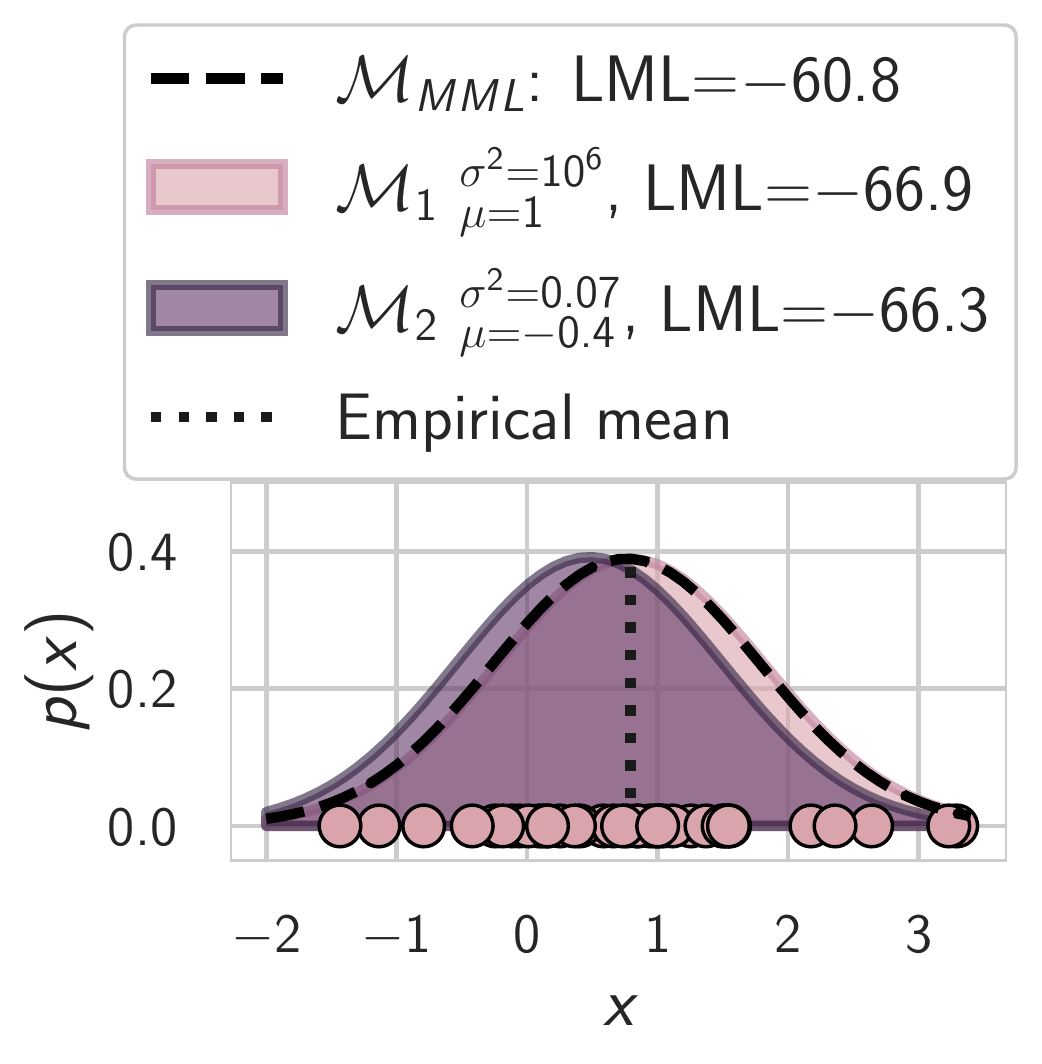}
        &
        \includegraphics[height=0.25\textwidth]{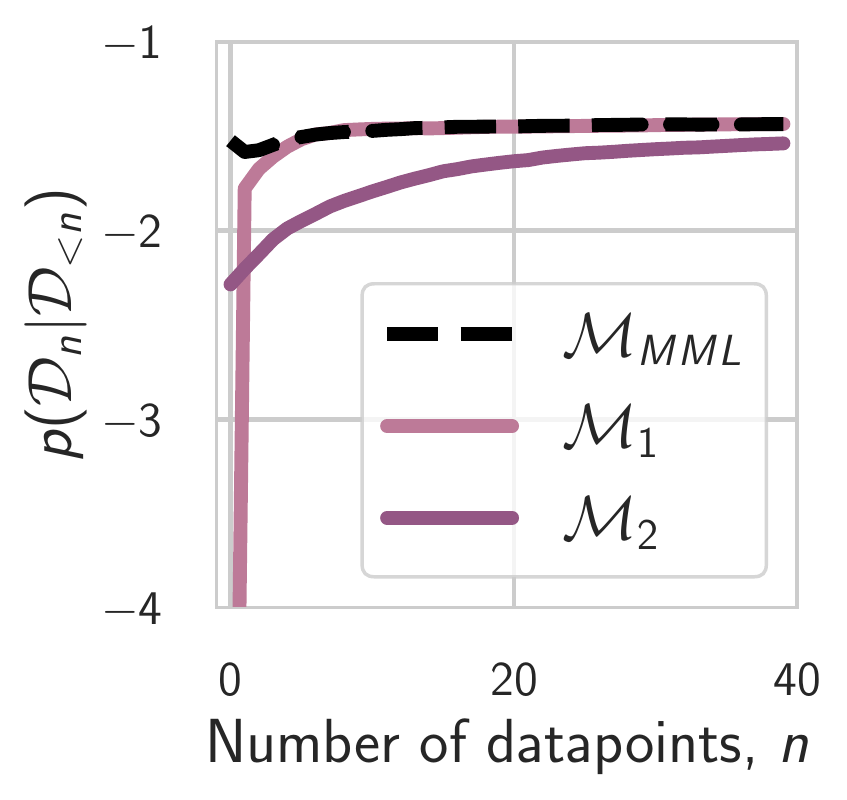}
        \\
        {\small (a) Fixed $\mu=0$} & 
        {\small (b) LML prefers $\mathcal{M}_2$} &
        {\small (c) Density learning curves} \\[5mm]
        \multicolumn{3}{c}{
        \begin{tabular}{ccc}
            \includegraphics[height=0.25\textwidth]{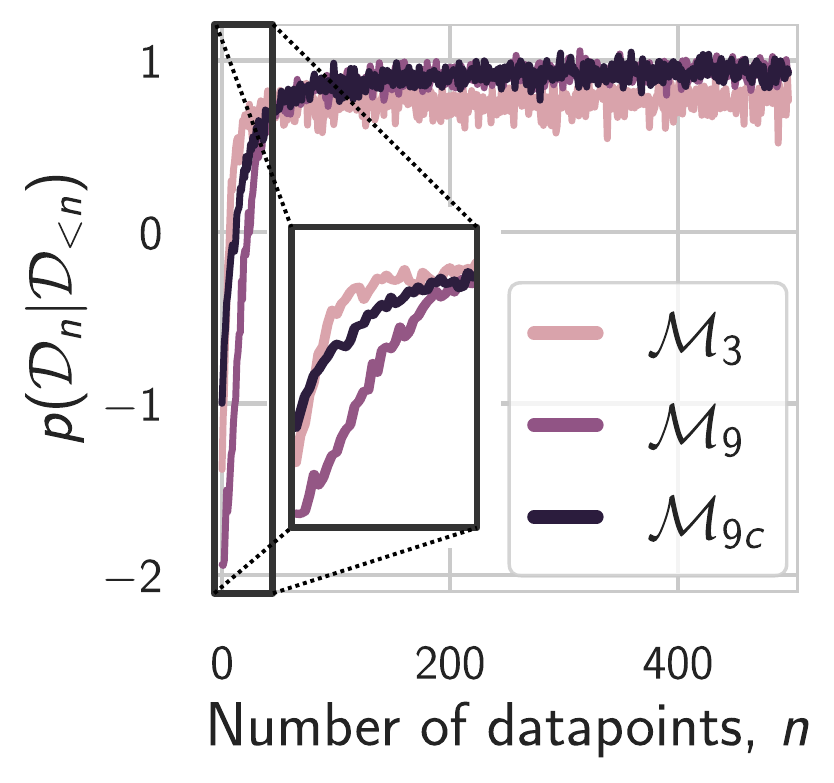}
            &\quad\quad&
            \includegraphics[height=0.25\textwidth]{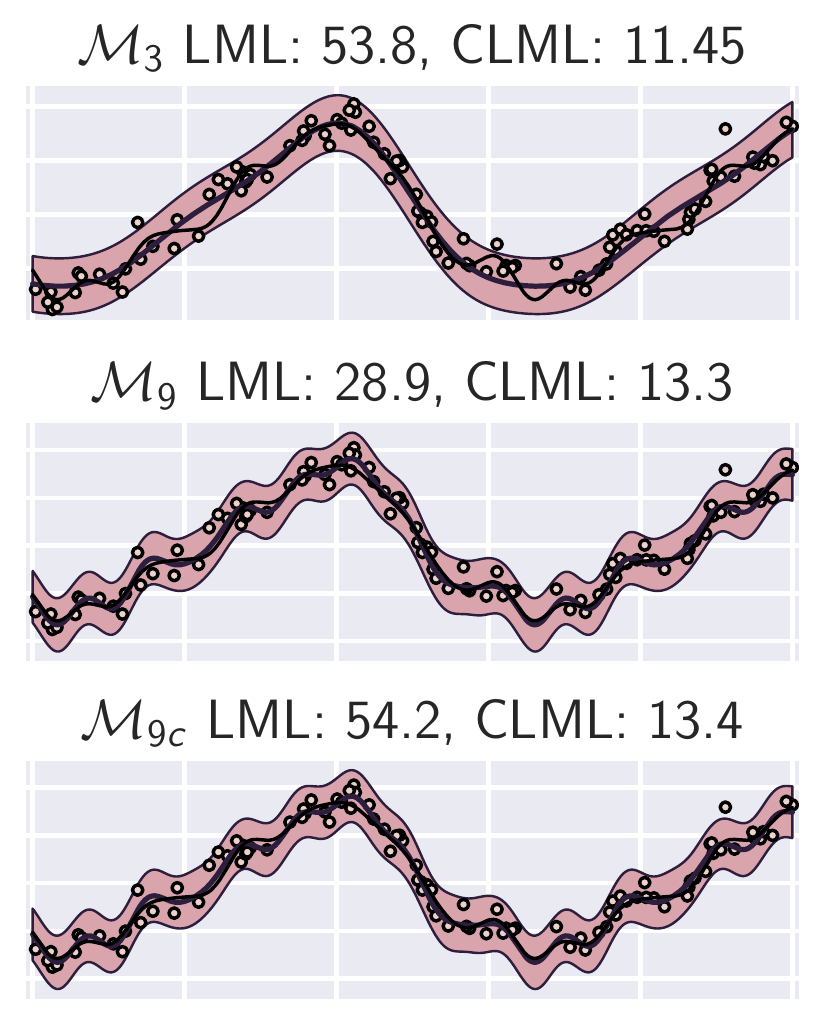}
            \\
            {\small (d) Fourier learning curves} &&
            {\small (e) Fourier model fits}
        \end{tabular}
        }
    \end{tabular}
\caption{
\textbf{Training speed and learning curves.}
\textbf{(a):}
The prior variance continues to affect the marginal likelihood as its value increases whereas the test predictive distribution becomes insensitive to its value starting at a certain threshold.
\textbf{(b):}
While $\mathcal{M}_1$ and $\mathcal{M}_{MML}$ provide identical fits of the data, LML favors $\mathcal{M}_{MML}$.
Moreover, LML prefers the model $\mathcal{M}_2$ with a poor data fit to $\mathcal{M}_1$.
\textbf{(c)}: 
$\mathcal{M}_1$  trains faster than $\mathcal{M}_2$, but has a worse LML than  $\mathcal{M}_2$. 
\textbf{(d)}: 
$\M_9$ provides a better fit after observing $60$ datapoints, but the LML prefers $\M_3$ until $n=297$.
The model $\M_{9c}$ provides a near-identical fit to $\M_9$ after observing $50$ datapoints, but is preferred by the LML.
\textbf{(e)}: 
Data fit for $\M_3, \M_9$ and $\M_{9c}$.
$\M_3$ undefits, while the other two models get identical fits.
}
\label{fig:learning-curves-sec5}
\end{figure*}

\section{Training Speed and Learning Curves}
\label{sec:speed}

The remainder of this paper will now concretely exemplify and further elucidate many of the conceptual issues we have discussed, regarding the misalignment between the marginal likelihood and generalization.

How a model updates based on new information is a crucial factor determining its generalization properties. We will explore this behaviour with \emph{learning curves} --- graphs showing how $\log p(\D_n \vert \D_{<n})$ changes as a function of $n$.
The LML can be thought of as the area under the learning curve 
\citep{lyle2020bayesian}.
We will see that the first few terms in the learning curve corresponding to small $n$ often decide which model is preferred by the LML.
These terms are typically maximized by small, inflexible models, biasing the LML towards underfitting.
We illustrate this behaviour in Figure \ref{fig:mll_pitfalls}(a): the marginal likelihood penalizes models with vague priors, even if after observing a few datapoints the posterior collapses, and generalizes well to the remaining datapoints. 

\textbf{Density Estimation}.\quad
Consider the process where $x$ is generated from a Gaussian distribution $\mathcal{N}(u, 1)$ and the mean parameter is in turn generated from a Gaussian distribution $u \sim \mathcal{N}(\mu, \sigma^2)$.
Figure~\ref{fig:learning-curves-sec5}(a) shows the LML and the test predictive log likelihood as a function of the prior variance $\sigma^2$. 
The posterior over $u$ and the predictive distribution are stable above a threshold of the prior variance $\sigma^2$, as the likelihood of the training data constrains the model and outweighs the increasingly weak prior.
However, as we increase  $\sigma^2$, the training data becomes increasingly unlikely according to the \textit{prior}, so the marginal likelihood sharply decreases with $\sigma^2$.
We provide analytical results in Appendix~\ref{sec:app_density_estimation}. 

A direct consequence of this behaviour is that two models may have the same generalization performance but very different values of the marginal likelihood;
or worse, the marginal likelihood might favor a model with a poor generalization performance. 
We can see this effect in Figure~\ref{fig:learning-curves-sec5}(b), where the predictive distributions of $\mathcal{M}_1$ and the maximum marginal likelihood (MML) model  $\mathcal{M}_{MML}$ almost coincide, but the LML values are very different. Moreover, we can design a third model, $\mathcal{M}_{2}$, with a prior variance $0.07$ and prior mean $2$ which leads to a poor fit of the data but achieves higher marginal likelihood than $\M_1$.
This simple example illustrates the general point presented in Section \ref{sec: notgen}: LML measures the likelihood of the data according to the prior, which can be very different from the generalization performance of the corresponding posterior.

In Figure \ref{fig:learning-curves-sec5}(c) we show $\log p(\D_n \vert \D_{<n})$ as a function of $n$, averaged over $100$ orderings of the data.
We see that $\mathcal{M}_1$ trains faster than $\mathcal{M}_2$ --- where the \textit{training speed} is defined by~\citet{lyle2020bayesian} as ``the number of data points required by a model to form an accurate posterior" --- but achieves a lower LML, contradicting recent claims that ``models which train faster will obtain a higher LML''~\citep{lyle2020bayesian}.
These claims seem to implicitly rely on the assumption that all models start from the same $\log p(\D_1 | \mathcal{M})$, which is not true in general as we  demonstrate in Figure \ref{fig:learning-curves-sec5}(c).

\textbf{Fourier Model.}\quad Consider the Fourier model 

$f(x, a, b) = \sum_{d = 1}^D a_d \sin(d \cdot x) + b_d \cos(d \cdot x),$
where $\{a_d, b_d\}_{d=1}^D$ are the parameters of the model, and $D$ is the order of the model.
To generate the data, we use a model of order $D = 9$.
We sample the model parameters $\hat a_d, \hat b_d \sim \N(0, (1 / d^2)^2)$.
We sample 100 data points $x \sim \mathrm{Uniform}[0, 1]$, and compute the corresponding $y = f(x, \hat a, \hat b) + \epsilon$, with noise $\epsilon \sim \N(0, 0.1^2)$.
We then compare an order-9 model $\M_9$ and an order-3 $\M_3$ model on this dataset using LML and CLML.
For both models, we use the prior
$p(a_d) = p(b_d) = \N(0, 1)$.
Note that the $\M_9$ model includes ground truth, while the $\M_3$ model does not.
 We show the fit for both models in Figure \ref{fig:learning-curves-sec5}(e) (top and middle).
$\M_9$ provides a much better fit of the true function, while $\M_3$ finds an overly simple solution.
However, the LML strongly prefers the simpler $\M_3$ model, which achieves a value of $53.8$ compared to $28.9$ for the model $\M_9$.
We additionally evaluate the CLML using $200$ random orders and conditioning on $m = 85$ datapoints. CLML strongly prefers the flexible $\M_9$ model with a value of $28.9$ compared to $11.45$ for $\M_3$.

We can understand the behaviour of LML and CLML by examining the decomposition of LML into a sum over data in Eq.~\eqref{eq:mll_expanded} and Figure \ref{fig:mll_pitfalls}(a).
In Figure \ref{fig:learning-curves-sec5}(d) we plot the terms $\log p(\D_n \vert \D_{<n})$ of the decomposition as a function of $n$, averaged over $200$ orderings of the data.
For $n > 50$ observed datapoints, the more flexible model $\M_9$ achieves a better generalization log-likelihood $\log p(\D_n \vert \D_{<n})$.
However, for small $n$ the simpler $\M_3$ model achieves better generalization, where the difference between $\M_3$ and $\M_9$ is more pronounced.
As a result, LML prefers the simpler $\M_3$ for up to $n=297$ datapoints!
For $n \in [50, 296]$ the LML picks the model with suboptimal generalization performance. 
We can achieve the best of both worlds with the corrected model $\M_{9c}$ with the parameter prior $a_d, b_d \sim \N(0, (1 / d^2)^2)$:
strong generalization performance both for small and large training dataset sizes $n$.
These results are qualitatively what we expect: for small datasizes, the prior, and thus the LML, are relatively predictive of generalization. For intermediate size data, the first terms in the LML decomposition have a negative effect on how well LML predicts generalization. For asymptotically large data sizes, the first terms have a diminishing effect, and the LML becomes a consistent estimator for the true model if it is contained within its support.
For further details, please see Appendix~\ref{sec:fourier_model}.

\textbf{Neural Networks.}\quad We show the rank of $6$ different neural network architectures on their BMA test accuracy on CIFAR-10 for different dataset sizes in Figure~\ref{fig:nns-correlations}(b) (Appendix). 
We see that DenseNet121 and GoogLeNet train faster than ResNet-18 and VGG19, but rank worse with more data. 
In Figure \ref{fig:nns-correlations}(a) (Appendix), we show the correlation of the BMA test log-likelihood with the LML is positive for small datasets and negative for larger datasets, whereas the correlation with the CLML is consistently positive. As above, the LML will asymptotically choose the correct model if it is in the considered options as we increase the datasize, but for these architectures we are nowhere near any regime where these asymptotic properties could be realized.
Finally, Figure \ref{fig:nns-correlations}(a) (Appendix) shows that the Laplace LML heavily penalizes the number of parameters, as in Section~\ref{sec:la-in-dl}. We provide additional details in Appendix~\ref{sec:app-training-speed}. 

\textbf{Summary.}\quad
In contrast with \citet{lyle2020bayesian}, we find that models that train faster do not necessarily have higher marginal likelihood, or better generalization. Indeed, the opposite can be true: fast training is associated with rapid posterior contraction, which can incur a significant Occam factor penalty (Section~\ref{sec: caseforit}), because the first few terms in the LML expansion are very negative. We also show that, unlike the LML, the CLML is positively correlated with generalization in both small and large $n$ regimes, and that it is possible for a single model to do well in both regimes.

\begin{figure*}[h!]
\centering
    \begin{tabular}{cc}
        \hspace{-0.3cm}\includegraphics[height=0.2\textwidth]{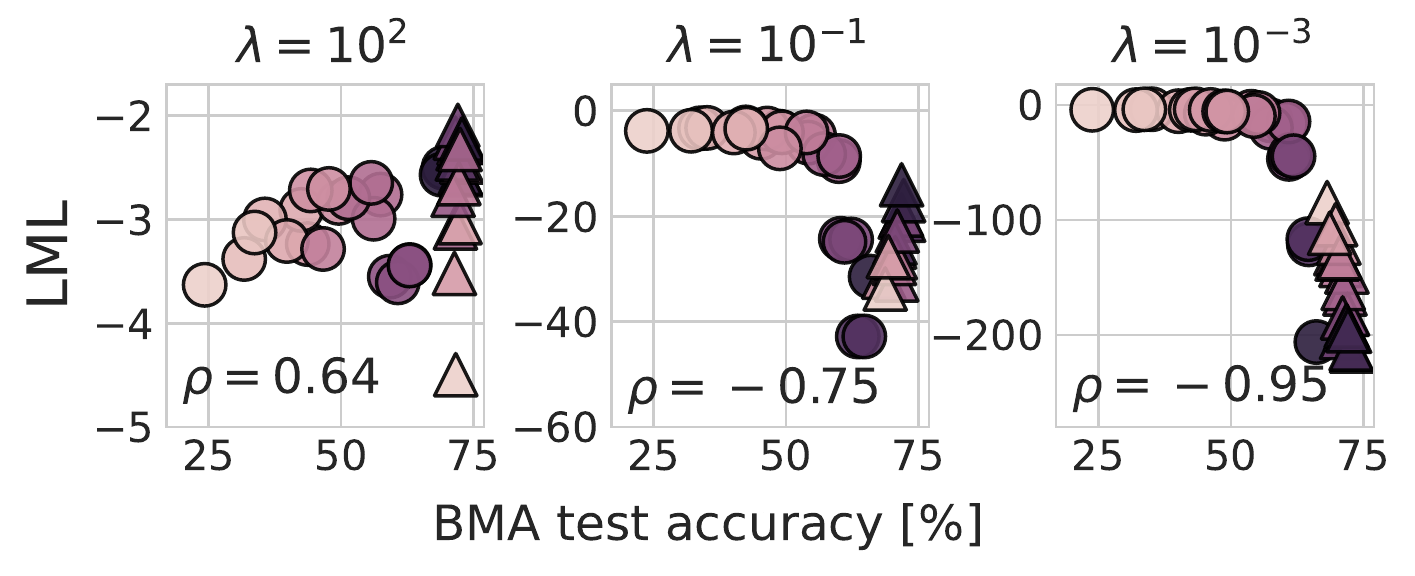}
        &
        \hspace{0.cm}\includegraphics[height=0.2\textwidth]{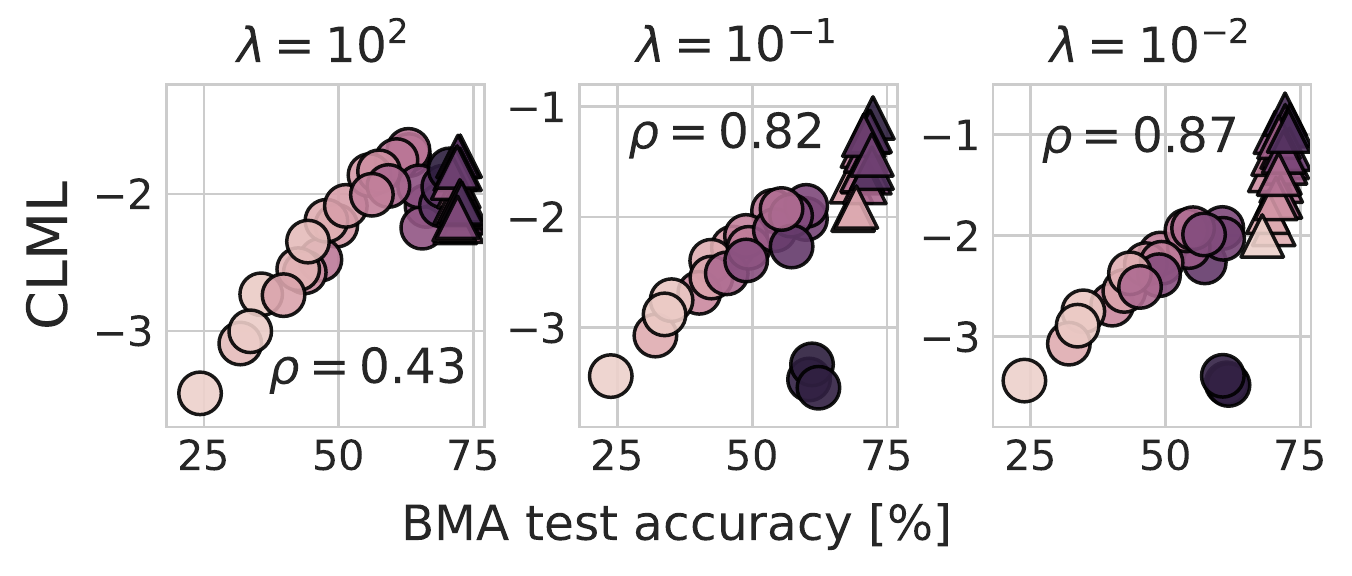}
        \\[-0.1cm]
        \hspace{-0.3cm}{\small (a) LML vs BMA accuracy} & 
        \hspace{-0.cm}{\small (b) CLML vs BMA accuracy}
        \\[0.2cm]
        \multicolumn{2}{c}{
            \hspace{-0.32cm}\includegraphics[height=0.2\textwidth]{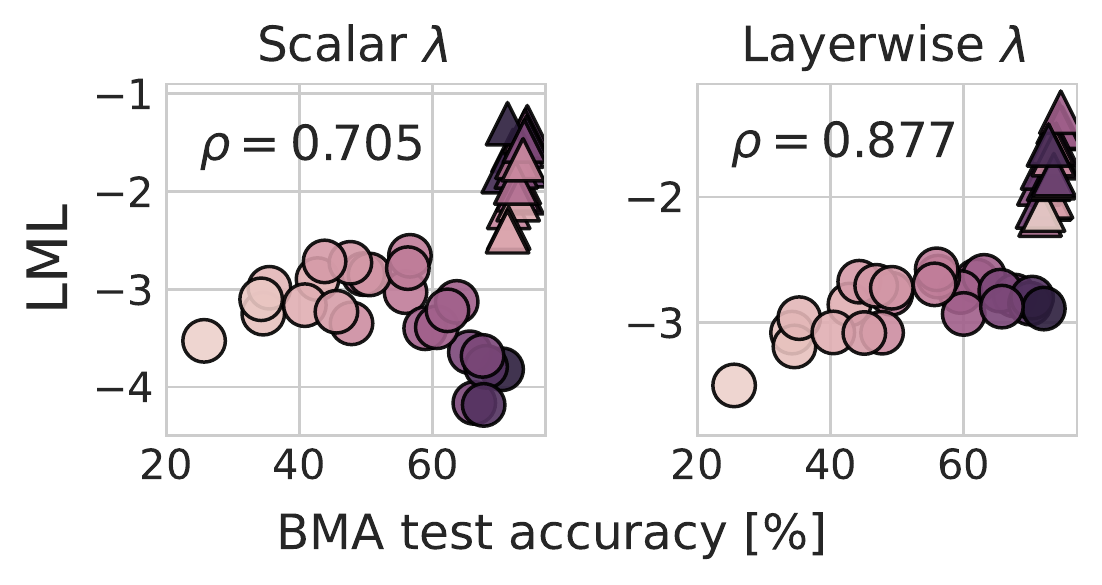}
            \quad
            \includegraphics[width=0.12\textwidth, trim={1.cm -2.0cm 0 0},clip]{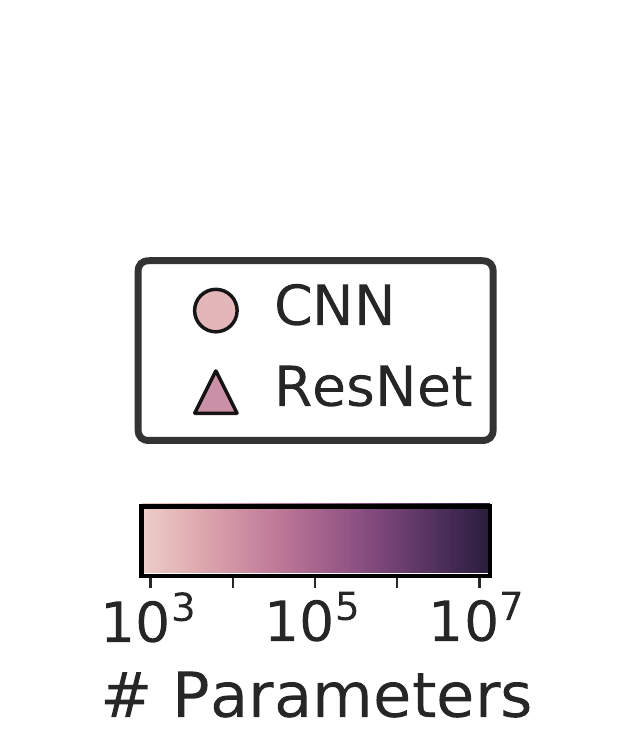}
        }
       \\[-0.1cm]
        \multicolumn{2}{c}{
            \hspace{-2.5cm}{\small (c) Optimized $\lambda$}
        }
        \\[-0.2cm]
    \end{tabular}
\caption{
\textbf{Neural hyperparameter optimization for CIFAR-100.}
The correlation (Spearman's $\rho$) between the model rankings and generalization. For panels \textbf{(a), (b)}, we consider a fixed prior precision $\lambda = 10^2$, $10^{-1}$, and  $10^{-3}$. 
\textbf{(a)} Correlation between the Laplace BMA test accuracy and the LML.
\textbf{(b)}: Correlation between the BMA test accuracy and the CLML.
\textbf{(c)}: 
Correlation between the BMA test accuracy and the LML for optimized global (left) and layer-wise (right) prior precision. 
The correlation between the LML and test accuracy highly depends on the value of the prior precision.
The CLML does not suffer from this sensitivity to the prior precision. 
}
\label{fig:laplace}
\end{figure*}

\section{Model Selection and Architecture Search}
\label{sec: selection}
\label{sec:cifar-exps}

In Section~\ref{sec: notgen}, we discussed how the marginal likelihood is answering a fundamentally different question than ``will my trained model provide good generalization?''.
In model selection and architecture search, we aim to find the model with the best predictive distribution, not the prior most likely to generate the training data.
Here, we consider neural architecture selection. In a way, the marginal likelihood for neural architecture search has come full circle: it was the most prominent example of the marginal likelihood in seminal work by \citet{mackay1992thesis}, and it has seen a resurgence of recent popularity for this purpose with the Laplace approximation \citep{immer2021scalable, daxberger2021laplace}.
We investigate the correlation between LML and generalization performance across $25$ convolutional (CNN) and residual (ResNet) architectures of varying depth and width on CIFAR-10 and CIFAR-100,
following the setup of \citet{immer2021scalable}.
See Appendix~\ref{sec:app_cifar_search} for more details.

First, we investigate the correlation between the Laplace marginal likelihood and BMA test accuracy, when the prior precision (aka weight decay) $\lambda$ is fixed. 
Figure~\ref{fig:laplace}(a) shows the results for fixed prior precision $\lambda = 10^2$, $10^{-1}$, and $10^{-3}$.
In each panel, we additionally report the Spearman’s correlation coefficient $\rho$~\citep{spearman1961proof} between the model rankings according to the BMA test accuracy and the LML.
LML is positively correlated with the BMA test accuracy when the prior precision is high, $\lambda = 10^2$, but the correlation becomes increasingly negative as $\lambda$ decreases. 
While the prior precision has little effect on the BMA test accuracy, it has a significant effect on the approximation of the LML values 
and model ranking!
As discussed in Section~\ref{sec: notgen}, the marginal likelihood heavily penalizes vague priors, especially in large, flexible models.
Moreover, as discussed in Section~\ref{sec:la-in-dl}, the Laplace approximation is especially sensitive to the prior variance, and the number of parameters in the model.

By the same rationale, we expect the conditional marginal likelihood to help alleviate this problem, since it evaluates the likelihood of the data under the posterior, rather than the prior.
Moreover, CLML is evaluated in \textit{function space} rather than in parameter space (see Appendix \ref{sec:app_clml_details} for details), and consequently is not sensitive to the number of parameters in the model, unlike the Laplace approximation (Section~\ref{sec:laplace_approx}).  Indeed, in Figure~\ref{fig:laplace}(b) the CLML exhibits a positive correlation with the generalization performance for both large and small values of the prior precision. In Appendix~\ref{sec:app_cifar_search}, we show that unlike the LML, the CLML is positively correlated with BMA accuracy, BMA log-likelihood, MAP accuracy and MAP log-likelihood across a wide range of prior precision values both on CIFAR-10 and CIFAR-100.

\textbf{Prior precision optimization.}\quad
In Figure~\ref{fig:laplace}(c), we show that optimizing the global or layer-wise prior precision leads to a positive correlation between the LML and the BMA test accuracy, following the online procedure in \citet{immer2021scalable}.
This optimization selects high-precision priors, leading to a positive correlation between the LML estimate and the test performance.
Notably, optimizing a separate prior scale for each layer leads to higher correlation, an observation that was also made in Chapter 3.4 of \citet{mackay1992thesis}.
In particular, if we only optimize a global prior precision, the correlation between the LML and the BMA test accuracy is negative for the CNN models, and we only recover a positive correlation by including the ResNet models.

\begin{figure*}[t]
\centering
    \begin{tabular}{ccc}
        \hspace{-0.3cm}\includegraphics[height=0.16\textwidth]{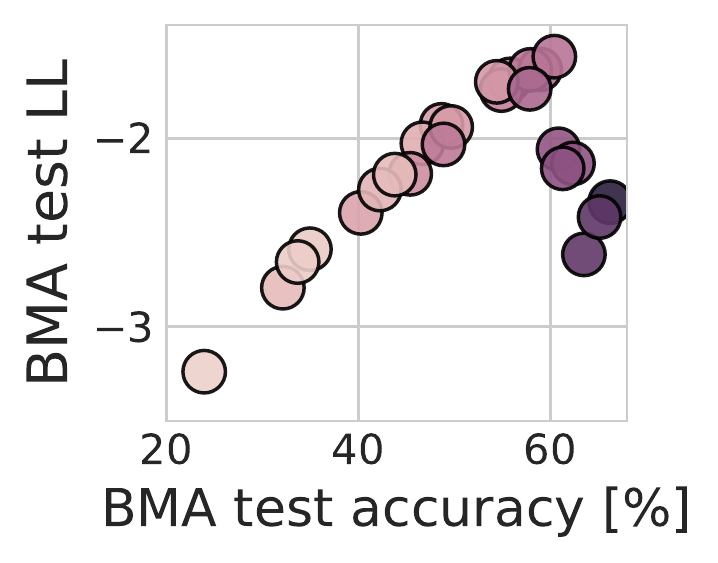}
        &
        \hspace{-0.32cm}\includegraphics[height=0.18\textwidth]{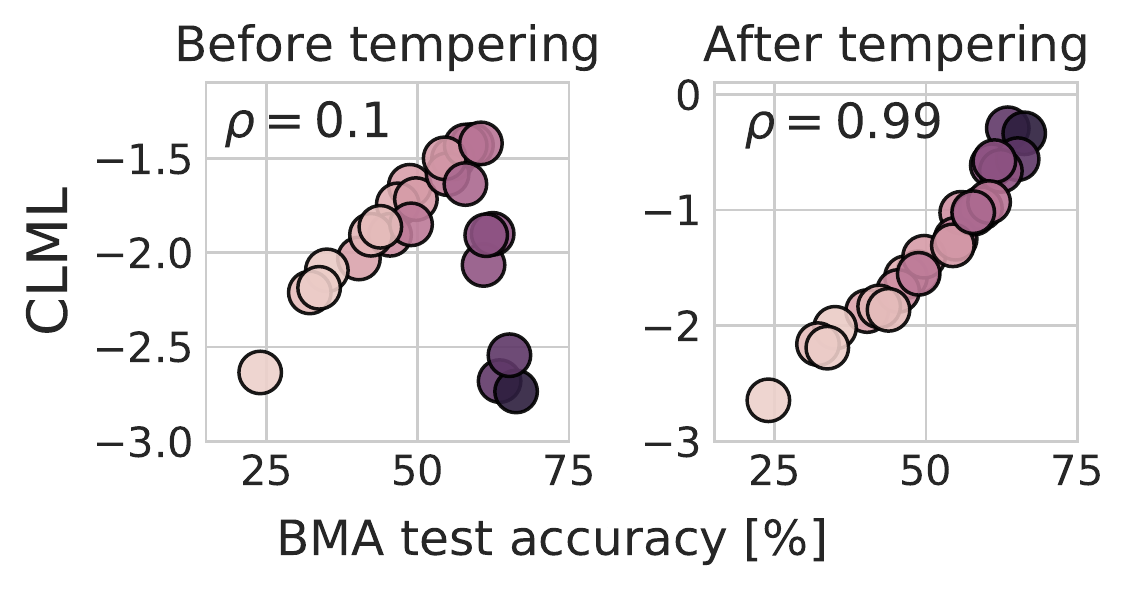}
        &       \hspace{-0.16cm}\includegraphics[height=0.16\textwidth, trim={0cm -2cm 0 0},clip]{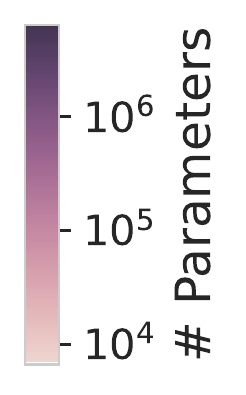}
      \\[-0.1cm]
        \hspace{-0.3cm}{\small (a) BMA LL vs BMA accuracy} & 
        \hspace{-0.cm}{\small (b) CLML vs BMA accuracy} &
        \\[-0.2cm]
    \end{tabular}
\caption{
\textbf{The effect of calibration on neural architecture search.}
For both panels \textbf{(a), (b)}, we consider a fixed prior precision $\lambda = 10^{-1}$ and focus on convolutional neural networks (CNNs), which exhibit outliers in Figure \ref{fig:laplace}. 
\textbf{(a)} Correlation between the Laplace BMA test accuracy and the Laplace BMA test log-likelihood (LL).
The models with a large number of parameters have a relatively good BMA test accuracy but a bad BMA test log-likelihood, and are therefore mis-calibrated. 
\textbf{(b)}: Correlation between the BMA test accuracy and the CLML for CNNs before and after we perform temperature scaling in order to calibrate the models. 
Calibrating the models significantly improves the correlation between the CLML and the BMA test accuracy, especially for large models. 
}
\label{fig:tempering}
\end{figure*}

\textbf{On the effect of calibration.}\quad Although the conditional marginal likelihood correlates much better than the marginal likelihood with the BMA accuracy, we notice in Figure \ref{fig:laplace}(b) that large CNN models appear to represent outliers of this trend.
To further investigate this behaviour, we plot the BMA test likelihood as a function of the BMA test accuracy in Figure \ref{fig:tempering}(a). 
We observe in this figure that the largest CNN models have higher accuracy but are poorly calibrated compared to other models.
These findings are compatible with the conclusions of \citet{guo2017calibration} which argue that larger vision models are more miscalibrated than smaller models. Incidentally, while Bayesian methods can help improve calibration \citep{wilson2020bayesian}, it appears the Laplace approximation is still clearly susceptible to overconfidence with large CNNs. 
Prompted by this observation, we calibrate these models via temperature scaling \citep{guo2017calibration} and find that the correlation between the CLML and BMA test accuracy for these well-calibrated models improves in Figure \ref{fig:tempering}(b).

To understand why model calibration is important for the alignment of CLML and BMA test accuracy and likelihood, let us examine the definition of CLML in Eq.~\eqref{eq:clml}.
The CLML represents the \textit{joint likelihood} of the held-out datapoints $\dataset_{\ge m}$ for the model conditioned on the datapoints $\dataset_{<m}$.
In particular, for miscalibrated models the test likelihood is not predictive of accuracy: highly accurate models can achieve poor likelihood due to making overconfident mistakes \citep{guo2017calibration}.

Moreover, the test likelihood can also be misaligned with the CLML for miscalibrated models. 
This difference is caused by the discrepancy between the marginal and joint predictive distributions:
the CLML evaluates the joint likelihood of $\dataset_{\ge m}$, while the test likelihood only depends on the marginal predictive distribution on each test datapoint.
The difference between the marginal and joint predictive likelihoods is discussed in detail in \citet{osbandneural} and \citet{wen2021predictions}.
We provide further intuition for this discrepancy in Appendix \ref{sec:app_calibration}.
In particular, Figure \ref{fig:clml-valid} shows that while the CLML and BMA validation loss correlate positively with the BMA test accuracy, the MAP validation loss correlates negatively with the BMA test accuracy.
One possible explanation for the difference between the correlation factors for the BMA validation loss in contrast with the MAP validation loss is that the BMA solutions generally tend to be less overconfident and better calibrated than MAP solutions, hence the positive correlation with the BMA test accuracy. 
This discrepancy between the marginal and joint predictive distributions also explains the difference between the CLML and standard cross-validation.

\begin{figure*}[t]
\centering
    \begin{tabular}{ccc}
        \hspace{-0.cm}\includegraphics[height=0.18\textwidth]{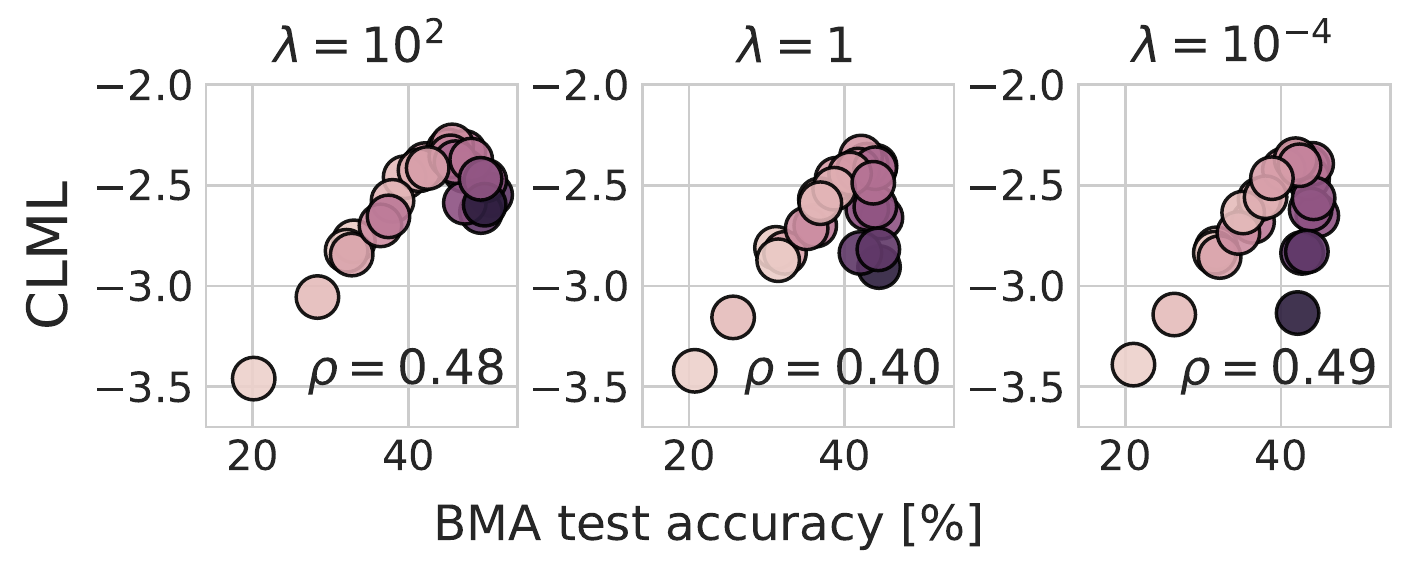}
        &
        \hspace{-0.32cm}\includegraphics[height=0.18\textwidth]{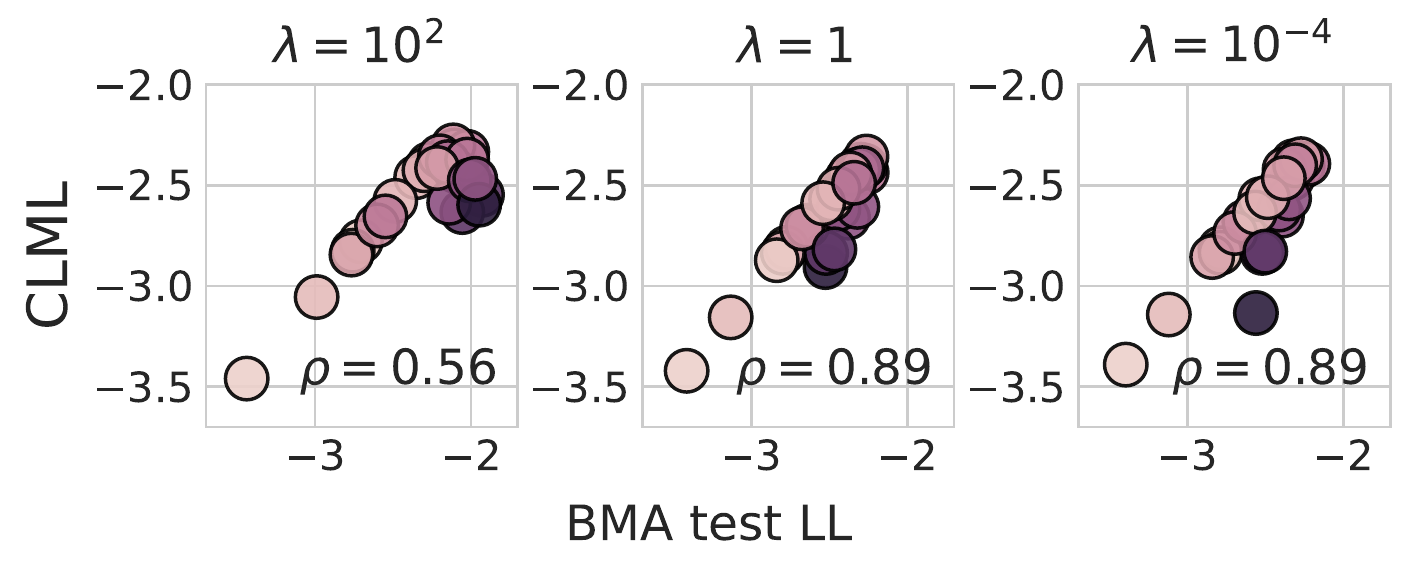}
        &       \hspace{-0.16cm}\includegraphics[height=0.16\textwidth, trim={0cm -2cm 0 0},clip]{figures/clml_function_space/cbar.pdf}
      \\[-0.1cm]
        \hspace{-0.cm}{\small (a) CLML vs BMA accuracy} & 
        \hspace{-0.cm}{\small (b) CLML vs BMA LL} &
        \\[-0.2cm]
    \end{tabular}
\caption{
\textbf{Estimating CLML with MCMC.}
CLML estimates produced with SGLD plotted against \textbf{(a)}: BMA test accuracy and \textbf{(b)}: BMA test likelihood for CNN models of varying width and depth, with different fixed values of prior precision $\lambda \in \{10^2, 1, 10^{-4}\}$.
Each plot also shows the corresponding Spearman's correlation coefficient $\rho$.
With SGLD CLML estimates we observe similar results to the results for Laplace estimates in Figure \ref{fig:laplace}: CLML correlates with both the BMA test accuracy and likelihood with the exception of the largest models which are poorly calibrated.
}
\label{fig:sgld_clml}
\end{figure*}

\textbf{Estimating CLML with MCMC.}\quad 
A key practical advantage of the CLML compared to the LML is that we can reasonably estimate the CLML directly with posterior samples produced by MCMC (see also the discussion in Appendix \ref{sec:app_clml_details}), which means working in function-space and avoiding some of the drawbacks (such as parameter counting properties) of the Laplace approximation. MCMC is difficult to directly apply to estimate the LML, because simple Monte Carlo integration to compute the LML would require sampling from a typically uninformative prior, leading to a high variance estimate. The CLML, on the other hand, can be viewed as a marginal likelihood formed on a subset of the data using an informative prior, corresponding to a posterior formed with a different subset of the data (Section~\ref{sec:conditional_mll}).

In Figure \ref{fig:sgld_clml}, we use the approximate posterior samples produced by the SGLD method \citep{welling2011bayesian} to estimate the CLML, BMA test accuracy, and log-likelihood.
We follow the setup from \citet{kapoor2022uncertainty} and use a cosine annealing learning rate schedule with initial learning rate $10^{-7}$ and momentum $0.9$.
We also remove data augmentation and use posterior temperature $T=1$, since data augmentation does not have a clear Bayesian interpretation \citep[][e.g.,]{wenzel2020good,fortuin2021bayesian,izmailov2021bayesian,kapoor2022uncertainty}.

We achieve results consistent with our observations using the Laplace estimates of CLML in Figure \ref{fig:laplace}: the CLML is closely aligned with both BMA accuracy and log-likelihood, with the exception of the largest models which are poorly calibrated.
These results suggest that the CLML can be estimated efficiently with Monte Carlo methods, which are significantly more accurate than the alternatives such as the Laplace approximation for models with complex posteriors such as Bayesian neural networks \citep{izmailov2021bayesian}.

\textbf{Summary.}\quad
Claims that ``the marginal likelihood can be used to choose between two discrete model alternatives after training" and that ``we only need to choose the model with a higher LML value"~\citep{immer2021scalable} do not hold universally: we see in Figure~\ref{fig:laplace}(a) that the marginal likelihood can be negatively correlated with generalization in practice!
In Figure \ref{fig:laplace}(c), we have seen that this correlation can be fixed by optimizing the prior precision, but in general there is no recipe for how many prior hyperparameters we should be optimizing to ensure a positive correlation.
For example, in Figure \ref{fig:laplace}(c) optimizing the global prior precision leads to a positive correlation for ResNet models but not for CNNs.
The CLML on the other hand consistently provides a positive correlation with the generalization performance.

\section{Hyperparameter Learning}\label{sec: hypers}

We want to select hyperparameters that provide the best possible generalization.
We have argued that LML optimization is not always aligned with generalization. As in Section~\ref{sec: mltraining}, there are two ways LML optimization can go awry.
The first is associated with overfitting through ignoring uncertainty. The second is associated with underfitting as a consequence of needing to support many unreasonable functions. 
CLML optimization can help address this second issue, but not the first, since it still ignores uncertainty in the hyperparameters.

We provide examples of both issues in GP kernel hyperparameter learning. Curiously, overfitting the marginal likelihood through ignoring uncertainty can lead to \emph{underfitting} in function space, which is not a feature of standard maximum likelihood overfitting. We then demonstrate that the CLML provides a highly practical mechanism for deep kernel hyperparameter learning, significantly improving performance over LML optimization. The performance gains can be explained as a consequence of the second issue, where we accordingly see the biggest performance gains on smaller datasets, as we predict in the discussion in Section~\ref{sec: mltraining}.

\begin{figure}[t]
\centering
    \begin{tabular}{ccc}
        \includegraphics[height=0.25\textwidth]{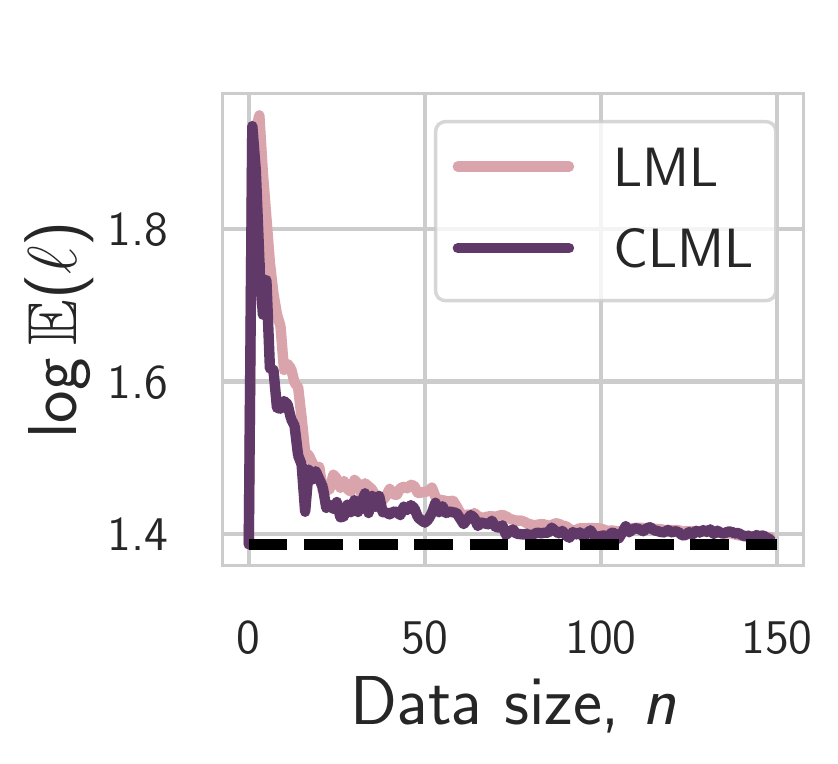}
        &\quad\quad&
        \includegraphics[height=0.25\textwidth]{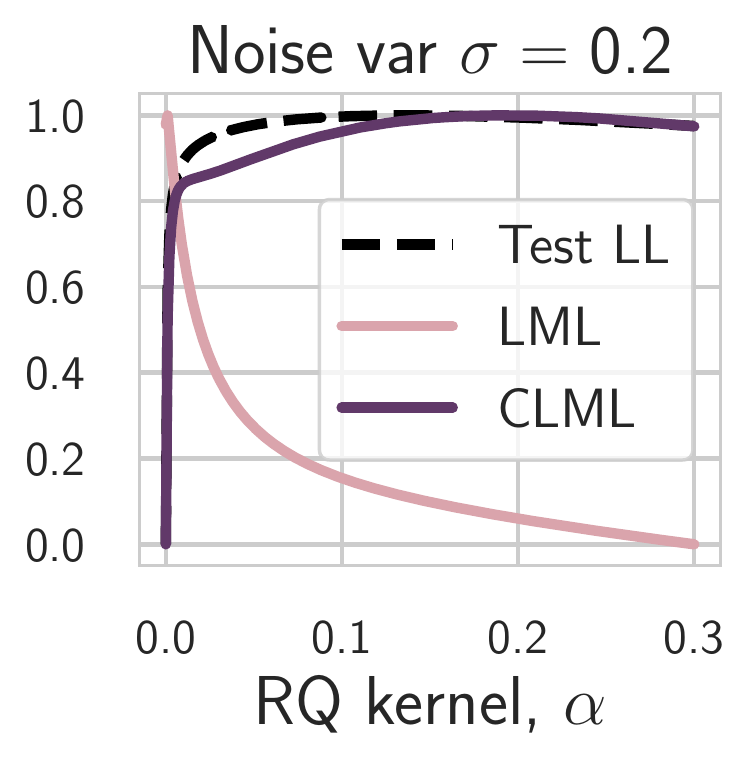}
      \\
        {\small (a) Underfitting bias} && 
        {\small (b) Underfitting with the RQ kernel}
    \end{tabular}
\caption{
\textbf{LML for hyperparameter learning in Gaussian processes.}
\textbf{(a):} The log-lengthscale learned by LML and CLML in a GP regression model averaged over $100$ datasets generated from a 
GP model with a lengthscale of $4$.
Unlike the train likelihood, LML has a bias towards underfitting, consistently overestimating the lengthscale, particularly for small $n < 20$.
\textbf{(b):} Test log-likelihood, LML and CLML as a function of the $\alpha$ hyper-parameter in the rational quadratic kernel with noise variance $\sigma^2 = 0.2$. The CLML is closely aligned with test log likelihood, unlike the LML.}
\label{fig:gp_resultsm}
\end{figure}

\subsection{Two issues with LML Optimization}
\label{sec: twoissues}

Using Gaussian process (GP) kernel learning, we provide illustrative examples of two conceptually different ways LML optimization can select hyperparameters that provide poor generalization, discussed in Section~\ref{sec: mltraining}. 

If we severely overfit the GP LML by optimizing with respect to the covariance matrix itself, subject to no constraints, the solution is the empirical covariance of the data, which is degenerate and biased. Figure~\ref{fig:gp_resultsm}(a) shows RBF kernel learning inherits this bias by over-estimating the length-scale parameter, which pushes the eigenvalues of the covariance matrix closer to the degenerate unconstrained solution. As we observe more data, the RBF kernel becomes increasingly constrained, and the bias disappears \citep{wilson2015human}. 
This finding is curious in that it shows how ignoring uncertainty in LML can lead to \emph{underfitting} in data space, since a larger length-scale will lead to a worse fit of the data. This behaviour is not a feature of standard maximum likelihood overfitting, and also not a property of the LML overfitting in the example of Figure~\ref{fig:lml_overfitting_example_main}. But since it is overfitting arising from a lack of uncertainty representation, the CLML suffers from the same issue.

In our next experiment, we generate data from a GP with a rational quadratic (RQ) kernel. Figure \ref{fig:gp_resultsm}(b) shows that if we overestimate the observation noise, then the LML is completely misaligned with the shape of the test log-likelihood as a function of the $\alpha$ hyper-parameter of the RQ kernel, whereas the CLML is still strongly correlated with the test likelihood. 
We see here the underfitting bias of Figure~\ref{fig:mll_pitfalls}(b), where supporting an $\alpha$ of any reasonable size leads to a prior over functions unlikely to generate the training data. In Appendix~\ref{sec:app_gp}, we show that under the ground truth observation noise both LML and CLML provide adequate representations of the test log-likelihood in this instance. Indeed, the CLML is additionally \emph{more robust to misspecification} than the LML.

We provide further details in Appendix~\ref{sec:app_gp}.

\begin{figure*}[t]
\centering
    \begin{tabular}{ccc}
        \multicolumn{3}{c}{
            \includegraphics[height=0.2\textwidth]{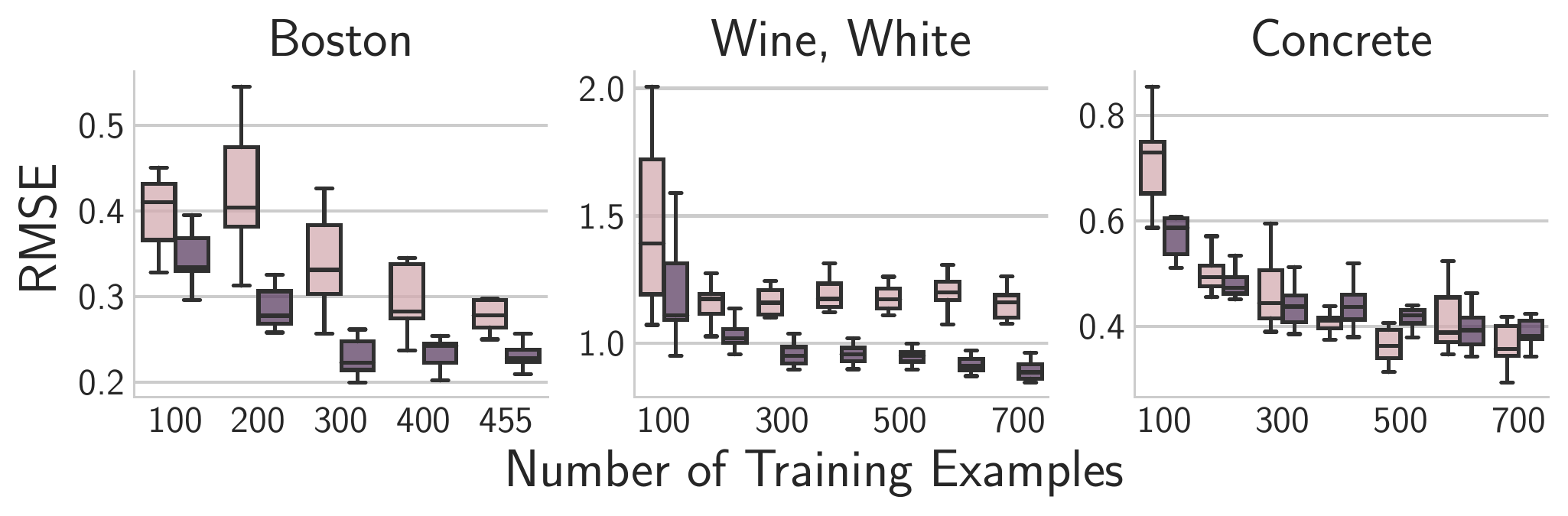}
        }
        \\
        \multicolumn{3}{c}{
        {\small (a) Deep Kernel Learning Regression}}
        \\[0.5cm]
        \includegraphics[height=0.2\textwidth,trim={0.cm -.7cm 0 0},clip]{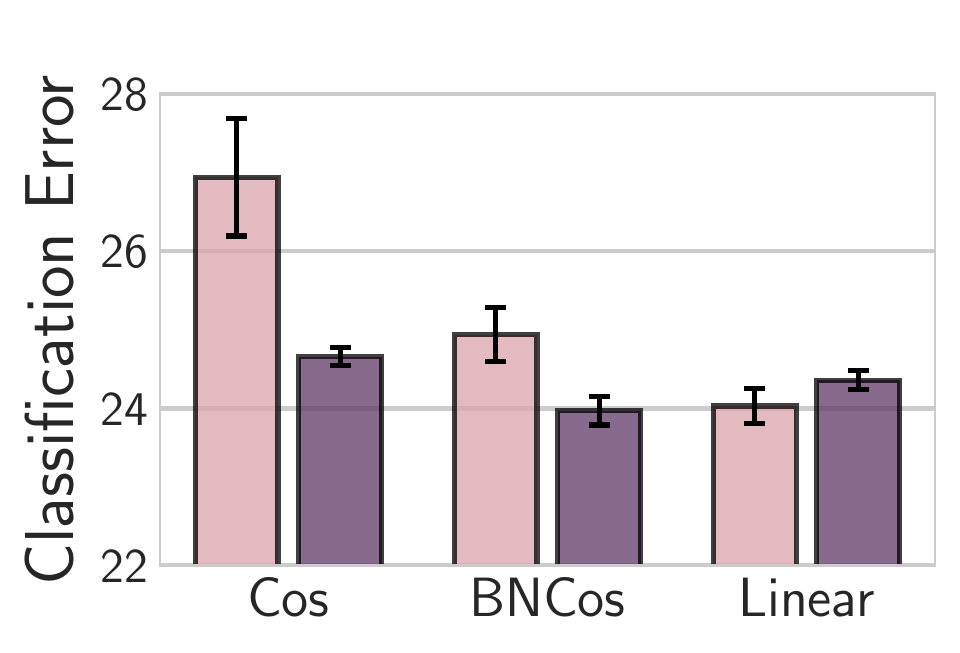}
        &
        \includegraphics[height=0.2\textwidth,trim={0.cm -.7cm 0 0},clip]{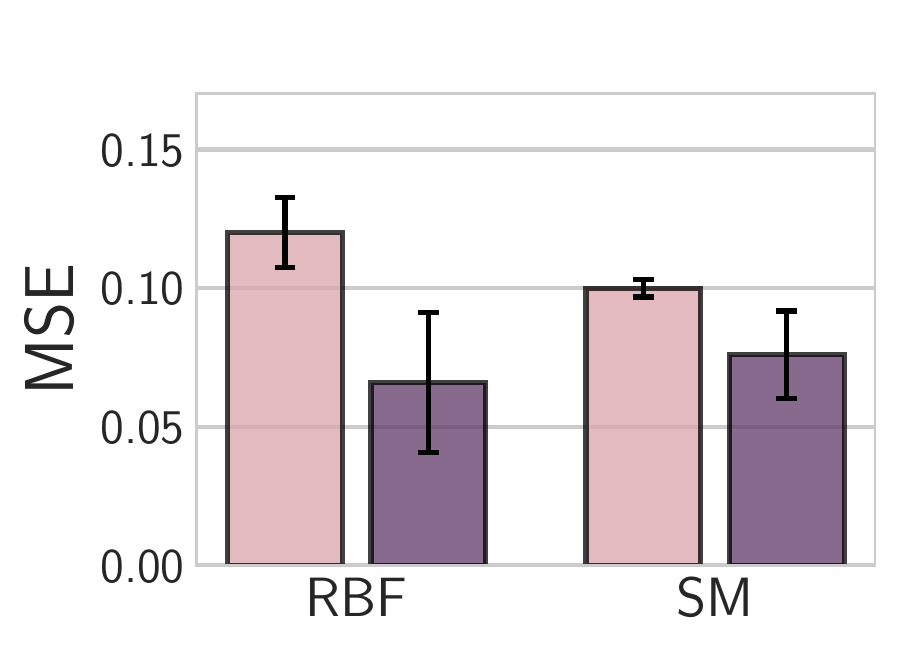}
        &
        \hspace{-0.2cm}\includegraphics[height=0.17\textwidth,trim={0.cm -2.0cm 0 0},clip]{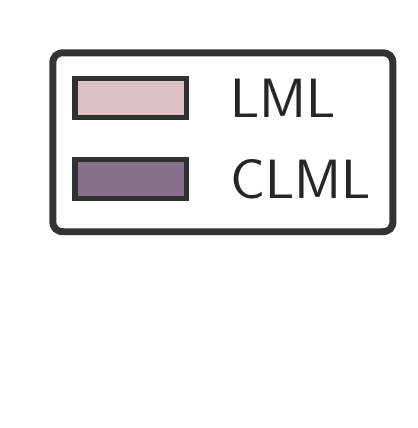}
      \\
        \hspace{0.cm}{\small (b) Transfer to Omniglot} & 
        \hspace{0.cm}{\small (c) Transfer to QMUL}
        &
        \\[-0.1cm]
    \end{tabular}
\caption{
\textbf{Deep kernel learning (DKL).}
\textbf{(a):} DKL trained with LML and CLML optimization with the cutoff $m$ at $90\%$ of the training data. CLML optimization outperforms LML optimization in low data regimes.
\textbf{(b):} Classification error ($\pm$ stderr) for Deep Kernel Transfer (DKT) models trained on the Omniglot dataset and evaluated on EMNIST. For both Cosine similarity and BatchNorm Cosine similarity kernels, CLML optimization outperforms LML optimization.
\textbf{(c):} MSE for DKT ($\pm$ stderr) applied to the QMUL dataset shift problem of recognizing head poses. 
For both RBF and SM kernels, CLML optimization outperforms LML optimization.
In all panels results are collected over $10$ random initializations. Test likelihood results (Appendix) are qualitatively similar.
}
\label{fig:dkl_results}
\end{figure*}

\subsection{Deep Kernel Learning}
\label{sec: dklfull}

Deep kernel learning (DKL) \citep{wilson2016deep} presents a scenario in which a large number of hyperparameters are tuned through marginal likelihood optimization. 
While it has been noted that DKL can overfit through ignoring hyperparameter uncertainty \citep{ober2021promises}, in this section we are primarily concerned with the underfitting described in Section~\ref{sec: mltraining}, where the CLML will lead to improvements.

Here we showcase CLML optimization as a practical tool in both UCI regression tasks and transfer learning tasks from \citet{patacchiola2020bayesian}.
In UCI regression tasks, we examine the performance of LML vs CLML in terms of test performance when training with limited amounts of training data.
In Figure \ref{fig:dkl_results} we see a common trend: when we are restricted to a small number of training examples, LML optimization is outperformed by CLML optimization.
As the number of training examples increases, the gap between LML and CLML optimized models closes.
We provide further details, with complete results including a comparison of negative log-likelihoods in Appendix \ref{app:dkl}.

In transfer learning tasks we are typically concerned with how well our method performs on unseen data, which may be from a different distribution than the training data, rather than how well aligned our prior is to the training data. 
In Figure \ref{fig:dkl_results} we reproduce the Deep Kernel Transfer (DKT) transfer learning experiments from \citet{patacchiola2020bayesian}, replacing LML optimization with CLML optimization. 
In these experiments DKL models are trained on one task with either LML or CLML optimization, and then evaluated on a separate but related task. 
Figure \ref{tbl:omni-emnist}(a), and Table \ref{tbl:omni-emnist} (Appendix), shows a comparison of methods on a transfer learning task in which we train on the Omniglot dataset and test on the EMNIST dataset. In both experiments CLML optimization provides a clear improvement over LML optimization.
Figure \ref{fig:dkl_results}(b), and Table \ref{tbl:qmul} (Appendix), shows a comparison of methods on the QMUL head pose regression problem for estimating the angular pose of gray-scale images of faces, where the individuals in the test set are distinct from those in the training set leading to dataset shift.
For experimental details see \citet{patacchiola2020bayesian}.

\section{The Marginal Likelihood and PAC-Bayes Generalization Bounds}
\label{sec: pac-bayes-lml}

PAC-Bayes provides a compelling approach for constructing state-of-the-art generalization bounds for deep neural networks \citep{mcallester1999pac,dziugaite2017computing, zhou2018non, alquier2021user, lotfi2022pac}. Like the marginal likelihood, PAC-Bayes bounds depend on how well the model can fit the data, and the amount of posterior contraction: models where the posterior differs significantly from the prior are heavily penalized (see Section \ref{sec:speed}). 
In fact, the similarity between PAC-Bayes and the marginal likelihood can be made precise: 
\citet{germain2016pac} show that for models where the likelihood is bounded, it is possible to derive PAC-Bayes generalization bounds that are monotonically related to the marginal likelihood \citep{mcallester1998some, germain2016pac}.
In other words, it is possible to construct formal generalization bounds based on the value of the marginal likelihood, with higher values of marginal likelihood implying stronger guarantees on generalization.

This result may initially appear at odds with our argument that the marginal likelihood is not the right tool for predicting generalization. In this section, we reconcile these two observations, and consider to what extent one can take comfort from the connection with PAC-Bayes in using the marginal likelihood for model comparison and hyperparameter tuning.

In Section \ref{sec: pacbayes-and-lml}, we provide a brief introduction to PAC-Bayes bounds and their connection to the marginal likelihood.
In Section \ref{sec: overfit-pacbayes}, we explain how PAC-Bayes generalization bounds provide insight into the underfitting and overfitting behaviour of the marginal likelihood.
In Section \ref{sec:data-dep-priors}, we discuss the connection between conditional marginal likelihood and data-dependent priors in PAC-Bayes generalization bounds.
In Section~\ref{sec: generalization-model-comparison},
we show that the PAC-Bayes bounds are typically not prescriptive of model construction.
In Section~\ref{sec: pacinpractice}, we show that the connections between state-of-the-art PAC-Bayes bounds in deep learning both to generalization and to marginal likelihood are limited.
Finally, we summarize our observations in Section \ref{sec:pac_summary}.

\subsection{PAC-Bayes and its Relation to the Marginal Likelihood}
\label{sec: pacbayes-and-lml}

Generalization bounds are often based on the following idea: if we select the parameters $w$ of a model from a fixed set of possible values $\Omega$, then the difference between the performance of the model with weights $w$ on the training data and the test data can be bounded by a term that depends on the size of the set $\Omega$ \citep[see e.g., Chapter 2 of][]{mohri2018foundations}.
In particular, if the set $\Omega$ of possible parameters is small, and we find a value $w \in \Omega$ that performs well on the training data, we can expect it to perform well on the test data.
At the same time, if the set $\Omega$ is infinite, then we cannot provide strong guarantees on the test performance.

PAC-Bayes generalizes this idea: we put a prior distribution $p(w)$ on the possible parameter values, and we provide guarantees for the expected performance of a random sample $w \sim q$ from an arbitrary posterior distribution $q$ (not necessarily the Bayesian posterior) over the parameters. Then, we can provide non-trivial generalization bounds even if the set of possible parameter values $\Omega$ is infinite, as long as the distribution $q$ does not differ too much from the prior:
if we come up with a distribution $q$ which is similar to a fixed prior, and such that on average samples from this distribution perform well on the training data, we can expect these samples to perform well on test data.

Formally, suppose we have a model, defined by a likelihood $p(d \vert w)$, where $d$ are data, and a prior $p(w)$ over the parameters $w$.
Suppose that we are given a dataset of $n$ points $d_i$ sampled randomly from the data distribution $p_\dataset$.
Let $\mathcal R_\dataset(w)$ denote the risk (average loss) on the training dataset $\dataset$ for the model with weights $w$, and
let $\mathcal R_{p_\dataset}(w)$ denote the true risk, i.e. the expected loss on test datapoints sampled from the same distribution as $\dataset$:
\begin{equation}
    \label{eq:risk}
    \mathcal R_\dataset(w) = \frac 1 n \sum_{i=1}^n \ell(d_i, w), \quad
    \mathcal R_{p_\dataset}(w) = \mathbb{E}_{d^* \sim p_\dataset} \ell(d^*, w), \quad
\end{equation}
for some loss function $\ell$.
We are especially interested in the case when the loss is given by the negative log-likelihood $\ell(d_i, w) = - \log p(d_i \vert w)$.

PAC-Bayes bounds are typically structured as a sum of the expected loss (negative log-likelihood) of a posterior sample on the training data and a complexity term  which measures the amount of posterior contraction.
Here the term ``posterior'' refers to an arbitrary distribution $q(w)$ over the parameters, and not necessarily the Bayesian posterior.

For example, the early bound introduced in \citet{mcallester1999pac} can be written as follows.
For any distribution $q$ over the parameters $w$, with probability at least $1-\delta$ over the training sample $\dataset$, we can bound the true risk:
\begin{equation}
\label{eq:mcallester}
    \underbrace{\mathbb{E}_{w \sim q} \mathcal{R}_{p_\dataset}(w)}_{\text{true risk}}
    \le
     \underbrace{\mathbb{E}_{w \sim q} \mathcal R_\dataset(w)}_{\text{risk on training data}}
    +
    \underbrace{\sqrt{\frac{\mathbb{KL}(q(w) \vert\vert p(w)) + \log(n / \delta) +2}{2n - 1}}}_{\text{complexity penalty}},
\end{equation}
where $\mathbb{KL}$ denotes the Kullback–Leibler divergence.
In particular, the complexity term heavily penalizes cases where $q(w)$ differs significantly from the prior $p(w)$, such as when there is significant posterior contraction.
Multiple variations of the bound in Eq.~\eqref{eq:mcallester} follow the same general form \citep{langford2001bounds,maurer2004note, catoni2007pac, thiemann2017strongly}.

As shown in Section \ref{sec:lml_vs_bma} and Eq.~\eqref{eq:lml_elbo_derivation}, the log marginal likelihood can be written in a similar form:
\begin{equation}
\label{eq:lml_elbo}
    -\log p(\dataset \vert \mathcal{M})
    = 
    \underbrace{\mathbb{E}_{w \sim p(w \vert \dataset)}\big[ -\log p(\dataset \vert w)\big]}_{\text{risk on training data}}
    +
    \underbrace{\mathbb{KL}(p(w \vert \dataset) \vert\vert p(w))}_{\text{complexity penalty}},
\end{equation}
where the negative log-likelihood plays the role of the loss, and the Bayesian posterior $p(w \vert \mathcal D)$ replaces $q$.
Eq.~\eqref{eq:lml_elbo} is a special case of the ELBO in Eq.~\eqref{eq:elbo} where the posterior $p(w \vert \dataset)$ takes place of the variational distribution, in which case the ELBO equals the marginal likelihood.

Eq. \eqref{eq:mcallester} and \eqref{eq:lml_elbo} above formalize the intuitive connection between the marginal likelihood and PAC-Bayes bounds.
Indeed, the difference between the log marginal likelihood in Eq.~\eqref{eq:lml_elbo} and the PAC-Bayes bound in Eq.~\eqref{eq:mcallester} with negative log-likelihood loss for $q = p(w \vert \dataset)$ is then only in the specific form of complexity penalty:
the complexity penalty in the marginal likelihood is $\mathbb{KL}(p(w \vert \dataset) \vert\vert p(w))$, while in the PAC-Bayes bound of \citet{mcallester1999pac} the complexity penalty is $\sqrt{\frac{\mathbb{KL}(p(w \vert \dataset) \vert\vert p(w)) + \log(n / \delta) +2}{2n - 1}}$.

In some cases, it is possible to construct PAC-Bayes bounds that explicitly depend on the value of the marginal likelihood. 
\citet{germain2016pac} show that for models with bounded likelihood, the PAC-Bayes bound of \citet{catoni2007pac} on the generalization of a sample from the Bayesian posterior is a monotonic function of the marginal likelihood.
Specifically, they show that if the log-likelihood is bounded as $a \le -\log p(d \vert w) \le b$ for all $d, w$, then with probability at least $1 - \delta$ over the dataset $\dataset$ of $n$ datapoints sampled from $p_\dataset$, we have
\begin{equation}
\label{eq:germain}
    \mathbb{E}_{w \sim p(w \vert \dataset)} \mathbb{E}_{d^* \sim p_\dataset} 
    \left[- \log p(d^* \vert w)\right]
    \le
    a + \frac{b-a}{1 - e^{a-b}} 
    \left[
    1 - e^{a} \cdot \sqrt[n]{p(\dataset \vert \mathcal{M}) \delta}
    \right].
\end{equation}
We note that \citet{germain2016pac} also derive other variants of this bound applicable to some unbounded likelihoods.
\citet{mcallester1998some} also derives PAC-Bayes bounds which explicitly depend on the marginal likelihood in a different setting.

Note that the right hand side of the bound in Eq. \eqref{eq:germain} is a monotonic function of the marginal likelihood $p(\dataset \vert \mathcal{M})$.
Eq.~\eqref{eq:germain} implies that model selection based on the value of the marginal likelihood is, in this case, equivalent to model selection based on the value of a PAC-Bayes generalization bound.
However, we have observed how the marginal likelihood is in many ways misaligned with generalization. 
In the following subsections, we reconcile these observations and provide insight into the limitations of marginal likelihood from the perspective of PAC-Bayes bounds.

\textbf{What do PAC-Bayes bounds guarantee?}\quad
We note that the PAC-Bayes bounds, e.g., in Eq.~\eqref{eq:mcallester} and Eq.~\eqref{eq:germain}, provide performance guarantees for the expected performance of a random posterior sample.
In particular, these guarantees are not concerned with the performance of the Bayesian model average, where the parameters are integrated out.
This observation supports our argument in Section~\ref{sec:lml_vs_bma}, where we argue that the marginal likelihood targets the average posterior sample rather than the BMA performance.
We note that \citet{morningstar2022pacm} discuss the distinction between targeting the BMA performance and average sample performance from the perspective of the PAC-Bayes and variational inference.
In particular, they propose PAC$^m$-Bayes bounds, which explicitly target the predictive performance of the BMA.

\subsection{Underfitting, Overfitting, and PAC-Bayes}
\label{sec: overfit-pacbayes}

In Section \ref{sec: pitfalls}, we identified underfitting and overfitting as two pitfalls of the marginal likelihood. We now revisit these limitations, but from the perspective of the connections between the marginal likelihood and PAC-Bayes generalization bounds in Section~\ref{sec: pacbayes-and-lml} and  Eq. \eqref{eq:germain}.

\paragraph{Diffuse Priors and Underfitting.}
In Section \ref{sec: pitfalls}, we have seen that model selection based on the marginal likelihood can overly penalize diffuse priors, which can often contract to posteriors that provide good generalization. Penalizing these priors can also lead to underfitting, where overly simple solutions are favoured in preference to a prior that can support much better solutions. While the exact complexity penalties in the marginal likelihood and PAC-Bayes differ, PAC-Bayes will have the same behaviour: for models with diffuse priors, the penalty in the PAC-Bayes bound of Eq.~\eqref{eq:mcallester}, $\sqrt{\mathbb{KL}(q(w) \vert\vert p(w))}$, will necessarily be large even for posteriors $q$ which achieve low values of risk on the training data, including the Bayes posterior $p(w \vert \mathcal{D})$.

\paragraph{Overfitting.} Above we have seen that both the marginal likelihood and PAC-Bayes bounds can be unreliable for comparing models with poor marginal likelihood. We would expect a model with poor marginal likelihood to have a loose PAC-Bayes bound, and a loose bound simply does not say anything about generalization. On the other hand, it may be tempting to assume, based on the connection with PAC-Bayes, that models with ``good'' marginal likelihood will provide good generalization, since the bounds can monotonically improve with improved marginal likelihood. 
However, we have seen in Section~\ref{sec: mltraining} that optimizing the marginal likelihood to learn hyperparameters can lead to overfitting, where models with arbitrarily high marginal likelihood can generalize poorly.

To reconcile these observations, we note that we cannot simply optimize PAC-Bayes bounds (e.g., Eq.~\eqref{eq:mcallester} or Eq.~\eqref{eq:germain}) with respect to the prior and expect to have the same guarantees on generalization, as those bounds do not hold simultaneously for all priors $p(w)$.

Indeed, optimization is a form of model selection, and in order to perform model selection while preserving generalization guarantees, we need to expand the guaranteed generalization error using a union bound, relying on the property that the probability of a union of events is bounded above by the sum of their probabilities. Using this property, we have that a PAC-Bayes generalization bound holds simultaneously for $k$ models under consideration with probability at least $1-k \delta$, where $\delta$ is as described in Eq. \eqref{eq:mcallester} for bounding the generalization error of each model individually. Note that even though $1-k\delta$ may be much lower than $1-\delta$, we can choose a very low value of $\delta$, for example by dividing it by $k$ so that the $k$ generalization bounds hold simultaneously with high probability.  By decreasing $\delta$ by a factor of $k$, we obtain a looser bound.

We can understand how much looser the bound becomes when we are comparing $k$ models. Noting that the logarithmic term involving $\delta$ from Eq. \eqref{eq:mcallester} becomes $\log(\frac{nk}{\delta}) = \log(\frac{n}{\delta})+\log(k)$, we see we pay a cost equivalent to adding exactly $\log(k)$ to the KL-divergence $\mathbb{KL}(q(w) \vert\vert p(w))$ by bounding $k$ models simultaneously if we fix the probability $1-\delta$ with which the bound holds for each individual model simultaneously.  

Even though the logarithm may scale slowly in the number of models we compare, this term accumulates. If we tune real-valued prior hyperparameters using gradient-based optimizers on the marginal likelihood, as is common practice \citep[e.g.,][]{mackay1992thesis, rasmussen06, wilsonadams2013, hensman2013uai, DKL, molchanov2017variational, daxberger2021laplace,
immer2021scalable, immerouderaa2022deepinv, immer2022invariance, schwobel2022last}, we are searching over an uncountably infinite continuous space of models, and lose any guarantee that the resulting model which maximizes the marginal likelihood will generalize at all.

\textbf{Does good marginal likelihood imply good generalization?} Not in general. The bound in Eq.~\eqref{eq:germain} implies that under certain conditions we can provide formal guarantees on generalization based on the value of the marginal likelihood.
The specific value of the marginal likelihood required for such guarantees will depend on the model and size of the dataset.
However, the bound in Eq.~\eqref{eq:germain} will only hold if we fix the model a priori without looking at the data, and then evaluate the marginal likelihood for this model.
In particular, we cannot optimize the prior or model class as suggested for example in \citet[][Chapter 3.4]{mackay1992thesis} to find models with high marginal likelihood and still expect generalization guarantees to hold.

\subsection{PAC-Bayes Bounds with Data-dependent Priors and the CLML}
\label{sec:data-dep-priors}

We have considered the conditional log marginal likelihood (CLML) as an alternative to the LML. As argued in Section~\ref{sec: mltraining}, the CLML is intuitively more aligned with generalization, and indeed outperforms the LML across several experimental settings in Sections \ref{sec:speed}, \ref{sec:cifar-exps} and \ref{sec: hypers}. 

The conditional log marginal likelihood (CLML) in Eq.~\eqref{eq:clml} is equivalent to replacing the prior distribution $p(w)$ with the \emph{data-dependent} prior $p(w | \mathcal{D}_{< m})$, and computing the marginal likelihood on the remainder of the training data. The CLML is related to the PAC-Bayes bounds with data-dependent priors in the same way as the marginal likelihood is related to PAC-Bayes bounds with conventional fixed priors:
\begin{equation}
    \mathbb{E}_{w \sim p(w \vert \dataset)} \mathbb{E}_{d^* \sim p_\dataset} 
    \left[- \log p(d^* \vert w)\right]
    \le
    a + \frac{b-a}{1 - e^{a-b}} 
    \left[
    1 - e^{a} \cdot (p(\dataset_{\geq m} \vert \dataset_{< m}, \mathcal{M} )\delta)^{\frac{1}{n-m+1}}
    \right]. 
\label{eq:bayes-clml}
\end{equation}
The data-dependent bound of Eq.~\eqref{eq:bayes-clml} is found by replacing the prior $p(w)$ in Eq.~\eqref{eq:germain} with the data-dependent prior $p(w | \mathcal{D}_{< m})$ and subtracting the $m - 1$ datapoints used to form the prior from the size of the observed dataset $n$.

Data-dependent PAC-Bayes bounds are often tighter than what can be obtained with the optimal data-independent prior \citep{dziugaite2021role}, providing further motivation for the CLML.

\subsection{Prescriptions for Model Construction}
\label{sec: generalization-model-comparison}

While PAC-Bayes generalization bounds can be used to provide precise theoretical guarantees on generalization, they are not typically prescriptive of model construction.
In particular, achieving state-of-the-art generalization bounds typically involves severe model compression with pruning, quantization, and restricting the parameters to a subspace of the parameter space \citep{zhou2018non, lotfi2022pac}.
These interventions reduce the complexity penalty in Eq.~\eqref{eq:mcallester} and lead to significantly improved generalization bounds, but they \textit{hurt} generalization. Conversely, increasing the number of parameters in deep models typically improves generalization, but loosens the PAC-Bayes bounds. 

It is not uncommon to see claims that the marginal likelihood should be prescriptive of model construction. For example, \citet{mackay1992thesis}
argues that poor correlation between generalization and marginal likelihood implies we should re-think our model construction to achieve better marginal likelihood. In particular, he observes that there is a weak correlation between marginal likelihood and generalization for networks that have a global prior over all parameters, but both improved marginal likelihood and generalization if we have priors with different variance scales in each layer. In recent work, the marginal likelihood is similarly used for architecture search, and for learning invariances \citep{van2018learning, immer2021scalable, immerouderaa2022deepinv}. However, we have seen many examples of how better marginal likelihood can lead to worse generalization, especially if we expand our model search broadly, or optimize hyperparameters, leading to overfitting (Section \ref{sec: mltraining}, \ref{sec:speed}, \ref{sec: selection}). Moreover, it may be reasonable to design models specifically to encourage posterior contraction --- e.g., specify priors over easily identifiable solutions, enabling fast contraction around a desirable solution. Such models would have poor marginal likelihood, but would be explicitly designed for good generalization.

In short, neither PAC-Bayes nor the marginal likelihood can be relied upon as a prescriptive guide to model construction. Indeed, more often than not PAC-Bayes prescribes the opposite of what we know to provide good generalization in deep learning, and the bounds can even have a negative correlation with generalization across architectures \citep{maddox2020rethinking}. The connection with PAC-Bayes thus provides the opposite of a reassurance that the marginal likelihood would provide reliable prescriptions.

However, there is work investigating PAC bounds that are more informative about modelling decisions \citep{lyle2020benefits, lotfi2022pac, behboodi2022pac}. 
And it may be reasonable in some cases to cautiously use the marginal likelihood as a guide, as it does provide a consistent estimator for constraints, and may not be at major risk of overfitting if we carefully limit the number of models we are comparing. One should just proceed with extreme care, and not generally read too much into what is prescribed by the marginal likelihood.

\subsection{Connections in Practice}
\label{sec: pacinpractice}

The theoretical relationships between the marginal likelihood and PAC-Bayes only precisely hold when the PAC-Bayes bounds use the Bayesian posterior $p(w \vert \dataset)$. In practice, state-of-the-art PAC-Bayes bounds in deep learning make use of posteriors that are wildly divergent from the Bayesian posterior. For example, it is standard to use a single Gaussian distribution as $q$, while the true Bayesian posterior is highly multimodal. Moreover, there is not even an attempt to have this Gaussian approximate the Bayesian posterior, in contrast to approximate Bayesian inference \citep[e.g.,][]{dziugaite2017computing, zhou2018non, jiang2019fantastic, lotfi2022pac}. The prior choices for PAC-Bayes are also often chosen somewhat artificially, for the sake of good bounds rather than good generalization. For instance, the first non-vacuous PAC-Bayes bounds for overparameterized neural networks \citep{dziugaite2017computing} were computed by choosing the prior distribution to be a multivariate normal in order to obtain the KL divergence in closed form. 

Furthermore, in practice we often approximate the marginal likelihood. Indeed, an approximation is unavoidable for Bayesian neural networks.  There are a variety of approximations, with different properties. As we have discussed in Section~\ref{sec:lmlapprox}, these approximations tend to further weaken the connection between marginal likelihood and generalization, and would increase the already significant discrepancy between the marginal likelihood and PAC-Bayes in practice. 

Moreover, those working with the marginal likelihood are typically interested in the predictive performance of the posterior weighted model average.
However, as we discuss in Sections~\ref{sec:lml_vs_bma} and \ref{sec: pacbayes-and-lml} respectively, the marginal likelihood and the PAC-Bayes bounds target the expected generalization of a single sample from the posterior.
There are often major discrepancies between these quantities. For example, a multimodal mixture of Gaussians posterior confers significant generalization benefits for Bayesian neural networks, but has a negligible effect on the PAC-Bayes bounds \citep{wilson2020bayesian}. This result is intuitive: the solutions with different basins of the posterior are similarly good, but \emph{complementary}. These solutions can therefore be combined to great effect, but if we are only taking one sample, it won't matter so much which basin it is from.

While the marginal likelihood and PAC-Bayes are closely related in theory, the relationship in practice is much less clear. It is also worth emphasizing that although the marginal likelihood and PAC-Bayes share some limitations, the way the marginal likelihood and PAC-Bayes are used in practice are very different. Generally, PAC-Bayes is used more cautiously than the marginal likelihood. It is not assumed, for instance, that PAC-Bayes is prescriptive of model construction, or would be closely aligned with generalization in optimizing for hyperparameters. 

\subsection{Summary}
\label{sec:pac_summary}

We conclude our discussion of the PAC-Bayes bounds with a summary of the points presented in this section. 

\begin{itemize}
    \item There exist both intuitive and formal connections between the marginal likelihood and PAC-Bayes generalization bounds.
    In particular, \citet{germain2016pac} bound the expected test performance of a random posterior sample with a monotonic transformation of the marginal likelihood, under certain conditions.
    \item While the connection between PAC-Bayes bounds and the marginal likelihood exists, it does not justify the use of the marginal likelihood for model selection with the goal of generalization. Indeed, PAC-Bayes bounds are not intended to be used in the ways the marginal likelihood is used for these purposes.
    \item In particular, the PAC-Bayes bounds and the marginal likelihood both heavily penalize diffuse priors leading to the underfitting behaviour: selecting overly simple models to avoid posterior contraction.
    \item Moreover, the PAC-Bayes bounds provide insight into the marginal likelihood overfitting behaviour: in order to perform model selection based on the PAC-Bayes bounds, we need to pay a penalty based on the logarithm of the size of the model class.
    Consequently, we lose generalization guarantees based on the marginal likelihood if we optimize the hyper-parameters of the prior.
    \item In principle, ``high'' values of the marginal likelihood can provide certain guarantees on generalization, but only in the case when the model is chosen and fixed a priori, without looking at the data.
    \item The state-of-the-art PAC-Bayes bounds in deep learning are not prescriptive for model construction, further suggesting caution in using the marginal likelihood as a guide for model construction.
    \item The LML is related to the PAC-Bayes bounds in the same way as the CLML is related to PAC-Bayes bounds with data-dependent priors, which are known to be significantly tighter.
    This connection provides additional theoretical motivation for CLML.
    \item Despite their relationship, the LML and PAC-Bayes are focused on different quantities: the LML is computing performance of a Bayesian model average, while PAC-Bayes measures the performance of a single posterior sample.
    \item In practice, the marginal likelihood and PAC-Bayes may only be loosely connected. The formal connection only exists when PAC-Bayes uses the Bayesian posterior, which is not the posterior used in practice. The marginal likelihood is also typically approximated in practice.
\end{itemize}

\section{Discussion}
\label{sec: discussion}

While the marginal likelihood provides a powerful mechanism for hypothesis testing, and can be practical for hyperparameter tuning, we show that it is in many ways misaligned with generalization. These results are particularly timely in light of recent work proposing the marginal likelihood for model selection and hyperparameter tuning in deep learning. 
We show that a conditional marginal likelihood retains the convenient properties of the marginal likelihood, but helps resolve the misalignment between the marginal likelihood and generalization, but is still susceptible to overfitting.
We find that the conditional marginal likelihood provides particularly compelling performance in learning deep kernel hyperparameters, especially on smaller datasets, and transfer learning problems.  

\paragraph{To what extent does approximate inference affect our results?}
As we discuss in Section~\ref{sec:la-in-dl}, approximating the marginal likelihood can further weaken its ability to predict generalization. However, almost all of our experiments use the exact LML and CLML: the density model, Fourier features, Gaussian processes, and deep kernel learning. While the neural architecture search experiments in Section \ref{sec:cifar-exps} necessarily use approximate inference, the results are qualitatively similar to the exact experiments, and these results are also what we expect from Section~\ref{sec: pitfalls}. 
A key advantage of working with the CLML is that it can be effectively approximated by sampling. However, what we observe about the LML behaviour stands on its own, independently of the CLML.

\paragraph{How does the correlation between LML and generalization vary with dataset size $n$?}
The relationship between the LML and generalization is non-monotonic with dataset size. 
For very small datasets, the first (and only) terms in the LML decomposition are typically predictive of generalization.
For intermediate datasets, these terms have a negative effect on the correlation with generalization, as the posterior can differ significantly from the prior.
Finally, for asymptotically large datasets, the first terms have a diminishing effect, and the LML becomes a consistent estimator for the correct model, assuming it is within the set of considered models.
We observe these results empirically in Figure~\ref{fig:learning-curves-sec5}(d), where LML picks the better generalizing model for $n < 50$ and $n > 296$. For $n$ in $[50, 296]$ it picks the wrong model.

\paragraph{Can we construct a model which performs well for both small and large $n$?}
While we are primarily concerned with model \emph{selection}, model \emph{construction} is intimately related. There is a conventional wisdom that one should use small models for small datasets, and large models for large datasets. We show in Figure~\ref{fig:learning-curves-sec5}(e) that this prescription is incorrect: we can achieve the best of both worlds, a model which is good in both small and large $n$ regimes, by combining flexibility with reasonable inductive biases, aligned with the discussion in \citet{wilson2020bayesian}.

\paragraph{Is the CLML ``just'' cross-validation?}
The LML itself is arguably a form of cross-validation, although it is not   
standard cross-validation \citep{fong2020marginal}. 
The CLML can be significantly more efficient and practical than standard cross-validation for gradient-based learning of many hyperparameters. However, our goal with the CLML was not to explore a measure that is starkly different from cross-validation, nor do we consider any arguable similarity with cross-validation a deficiency. Instead, we show how a minor modification to the LML can improve alignment with generalization, and be practical for hyperparameter learning. 

We also show in Appendix~\ref{sec:app_cifar_search} and Figure \ref{fig:clml-valid} that the CLML correlates better than negative of the MAP validation 
loss with the BMA test accuracy, while the BMA validation loss achieves a similar degree of correlation.
In Section \ref{sec:cifar-exps} and Appendix \ref{sec:app_calibration} we also discuss the relationship between the CLML and the validation likelihood in more detail. 
We show that these two metrics can deviate from each other significantly for models with miscalibrated uncertainty estimates.

\paragraph{Can we take comfort in using the marginal likelihood for model selection and hyperparameter learning, due to its connection with PAC-Bayes?} Generally no. We summarize our observations about these connections in Section~\ref{sec:pac_summary}. 

\paragraph{Should we use the CLML instead of the LML?}
The DKL hyperparameter learning with the CLML is of particular practical significance. These experiments, in Section~\ref{sec: dklfull}, show that the CLML can be a promising drop-in replacement for the LML for learning many hyperparameters, especially for transfer learning and small datasets, and are our most substantial experiments involving DNNs. Future work in this area could have a substantial impact on the way we estimate hyperparameters in probabilistic models.

\paragraph{Do we expect to see these issues with the marginal likelihood arising in common practice?} 
Yes. For example, as above, we see clear improvements in using the CLML in place of the LML for deep kernel learning, which is a widespread application of the LML. More broadly, it is not safe to use the marginal likelihood as a proxy for generalization. Predicting the generalization of trained models is simply not the question the marginal likelihood is intended to answer. If we had a good prior, then these issues would be less severe. But if we are doing model selection, it is precisely because we are particularly uncertain which prior is reasonable.
Moreover, if we attempt to compare too many models, for example by optimizing the LML with respect to the parameters of the prior, we can easily see marginal likelihood overfitting. We note that even when there is a correlation between marginal likelihood and generalization, the correlation will often be adversely affected by the conceptual issues described in Figure~\ref{fig:mll_pitfalls}, as we see with DKL.

\paragraph{Are neural architecture search results addressed by optimizing the prior scale?} 

In Section~\ref{sec:cifar-exps}, we show that the marginal likelihood is correlated with BMA test accuracy once we optimize a layer-wise prior scale parameter.
At the same time, if we do not optimize the prior scale, this correlation can be negative, and if we only optimize a global prior scale shared across all layers, the correlation is negative if we restrict our attention to CNN models. It is important to note, however, that the amount of flexibility in the prior specification is completely arbitrary here: if we optimize too few parameters (a global prior scale) or too many parameters (a prior mean and variance separately for each weight), the correlation between the marginal likelihood and the generalization performance will break.
The specific choice of layer-wise prior scale optimization is not in any way suggested by our conceptual understanding of the marginal likelihood.
The fact that the marginal likelihood is only aligned with generalization for some subset of possible models, without a clear specification, is a major limitation for its use for model selection in practice. Moreover, even positive correlations with generalization are often adversely affected by the issues in Section~\ref{sec: pitfalls}.

\paragraph{Is prior sensitivity the main drawback of the marginal likelihood?} We do not consider prior sensitivity a drawback of the marginal likelihood. The marginal likelihood tells us how likely our data were generated from our prior. It \emph{should} be sensitive to the prior. 
Prior sensitivity can sometimes be problematic in using the marginal likelihood to predict generalization, but the marginal likelihood should not be used for that purpose. 

\vspace{5mm}

\subsection*{Acknowledgements}
We thank Sam Stanton, Marc Finzi, Rich Turner, Andreas Kirsch, Sebastian Farquhar, and Tom Minka for helpful discussions. This research is supported by NSF
CAREER IIS-2145492, NSF I-DISRE 193471, NIH R01DA048764-01A1, NSF IIS-1910266, NSF
1922658 NRT-HDR, Meta Core Data Science, Google AI Research, BigHat Biosciences, Capital
One, and an Amazon Research Award

\newpage
\appendix
\onecolumn

\vbox{%
\hsize\textwidth
\linewidth\hsize
\vskip 0.1in
\hrule height 4pt
  \vskip 0.25in
  \vskip -\parskip%
\centering
{\LARGE\bf
Appendix
\par}
\vskip 0.29in
  \vskip -\parskip
  \hrule height 1pt
  \vskip 0.09in%
}

\appendix

\section*{Appendix Outline}
This appendix is organized as follows: 
\begin{itemize}
\item In \autoref{sec:app-overfit-gp}, we demonstrate overfitting in Gaussian processes where the mean function is parameterized with a small multi-layer perceptron and we learn the parameters of the MLP by optimizing the LML.
\item In \autoref{sec:app_clml_details}, we provide details about the CLML approximation in our experiments and how it can be approximated directly in the function space in general. 
\item In \autoref{sec:fourier_model}, we provide additional details on the Fourier model.
\item In \autoref{sec:app-training-speed}, we demonstrate that faster training speed does not necessarily imply a better generalization performance for deep neural networks.
\item In \autoref{sec:app_density_estimation}, we provide analytical details and additional experiments for the density estimation example.
\item In \autoref{sec:app_calibration}, we discuss the difference between the joint and marginal predictive likelihoods and provide an example where CLML deviates from the test likelihood.
\item In \autoref{sec:app_cifar_search}, we provide the experimental details for neural architecture search as well as additional results for CIFAR-10 and CIFAR-100.
\item In \autoref{sec:app_gp}, we provide additional results on Gaussian processes that show that the CLML is more robust to misspecification than the LML. 
\item In \autoref{app:dkl}, we give details on the deep kernel learning experiments alongside additional results. 
\item Lastly, in \autoref{sec:ablation-clml}, we study the effect of the choice of $m$ on the CLML.

\end{itemize}

\section{Overfitting in Gaussian Processes}
\label{sec:app-overfit-gp}

In Figure \ref{fig:lml_overfitting_example_main} we provide a simple example in which LML optimization leads to severe overfitting. We generate a set of $100$ evenly spaced points from a GP prior with an RBF kernel with a lengthscale of $0.75$ and observation noise of $0.02$. 

We then use LML optimization to train two GP models on the first $50$ data points: the first model is a standard GP with constant mean and an RBF kernel, and the second is a GP with an RBF kernel, but where we have replaced the mean function with a small neural network. The mean function of the second model is a feed forward ReLU network with two hidden layers each with $50$ units. 

In Figure \ref{fig:lml_overfitting_example_main} (left) we see that by fitting the training data with a GP with constant mean and an RBF kernel we do not necessarily extrapolate well, but our prediction is reasonable and our uncertainty appears to be well-calibrated. In Figure \ref{fig:lml_overfitting_example_main} (right) we see that in training a GP with an overly flexible mean function we are able to overfit to the training data, and produce extrapolation predictions that are both incorrect, and very confident.
By building a model with an incredibly flexible prior, we are able to optimize out LML to concentrate heavily around a single solution. This model has a high likelihood of generating the data, but does not extrapolate well, similar to the effect presented in Figure \ref{fig:mll_pitfalls}(c).

\section{Details on the Conditional Marginal Likelihood}
\label{sec:app_clml_details}

Note that unlike the LML, the CLML depends on the ordering of the datapoints.
To remove this undesirable dependence, we can average the CLML over all possible orderings of the data:
\begin{equation}
    \label{eq:cmll_permutations}
    \frac 1 {n!} \sum_{\sigma \in S_{n}}
    \sum_{i=m}^n \log p(\D_{\sigma(i)} \vert \D_{\sigma(1)}, \ldots, \D_{\sigma(i)}, \M),
\end{equation}
where $S_n$ is the set of all the possible permutations of $n$ elements.
Using all the $n!$ permutations is prohibitively expensive in most practical scenarios, so we approximate Eq. \eqref{eq:cmll_permutations} as 
$\frac 1 {|\hat S|} \sum_{\sigma \in \hat S} \sum_{i=m}^n \log p(\D_{\sigma(i)} \vert \D_{\sigma(1)}, \ldots, \D_{\sigma(i)}, \M)$,
where $\hat S \subset S_n$ is a set containing several random permutations of the dataset.  When $\D$ is a large dataset such that $\D_{<m}$ and $\D_{\ge m}$ are both sufficiently large, a single permutation may suffice.

\textbf{Implementation.}\quad
For all experiments involving the Laplace approximation, we compute the conditional marginal likelihood as follows:

\begin{enumerate}
    \item We train a model on $80\%$ of the training data, and fit the LA approximation on the same subset of the data. 
    \item We tune a hyperparameter $T$ that we use to re-scale the Laplace posterior covariance matrix to ensure that it does not lead to very low BMA accuracies.
    We choose the value of $T$ that achieves the highest BMA accuracy (average over $20$ samples) on $5\%$ of the training data. 
    Our experimental results show that the optimal values of $T$ generally ranges between $0.1$ and $0.001$, so the LA posterior does not collapse on the MAP solution even as we use this re-scaling parameter. 
    \item Finally, we directly compute the CLML $p(\mathcal{D}_{\geq m}|\mathcal{D}_{<m})$ using the remaining $15\%$ of the training data. This quantity corresponds simply to the log predictive likelihood of the $15\%$ of the data approximated using a Bayesian model average of LA over $20$ samples. 
\end{enumerate}

It is important to note that in all our plots, we show the BMA test accuracy and log-likelihood of the model trained on the \textbf{entire} training data and for the Laplace approximation fit on the entire training data as well, and not just the $80\%$ subset that we condition the CLML on. 

\textbf{Function space.}\quad
Note that in the procedure described above, we approximate CLML purely in function space: the estimate only depends on the predictions made by the Bayesian model average and not the values of individual parameters.
The standard Laplace approximation of the LML is on the other hand quite sensitive to the number of parameters in the model.
Approximating the LML directly in function space is hard, because it would require approximating the integral over the prior with simple Monte Carlo, but
sampling from the prior over the weights of a neural network we will never randomly sample the parameters that are likely to generate the data.

\section{Details on the Fourier model}
\label{sec:fourier_model}

We can construct a model that achieves the best of both worlds: strong generalization performance both for small and large training dataset sizes $n$.
To do so, we consider the corrected model $\M_{9c}$, an order-9 model with a prior $a_d, b_d \sim \N(0, (1 / d^2)^2)$, following the data generating process.
We show the data fit and the learning curve for $\M_{9c}$ in Figure \ref{fig:learning-curves-sec5}(d) and (e) (bottom).
While the predictive distribution of $\M_{9c}$ is almost identical to that of $\M_9$ (see Figure \ref{fig:learning-curves-sec5}(e) (middle)), $\M_{9c}$ achieves the best marginal likelihood (comparable to $\M_3$) and the best CLML (comparable to $\M_9$) when evaluated on $100$ datapoints.
In the learning curve, $\M_{9c}$ provides comparable performance to $\M_3$ for small $n$, and comparable performance to $\M_9$ for large $n$. 

\section{Training speed of Deep Neural Networks}
\label{sec:app-training-speed}

\begin{figure}[h]
\centering
    \begin{tabular}{cc}
        \hspace{-0.2cm}\includegraphics[height=0.25\textwidth]{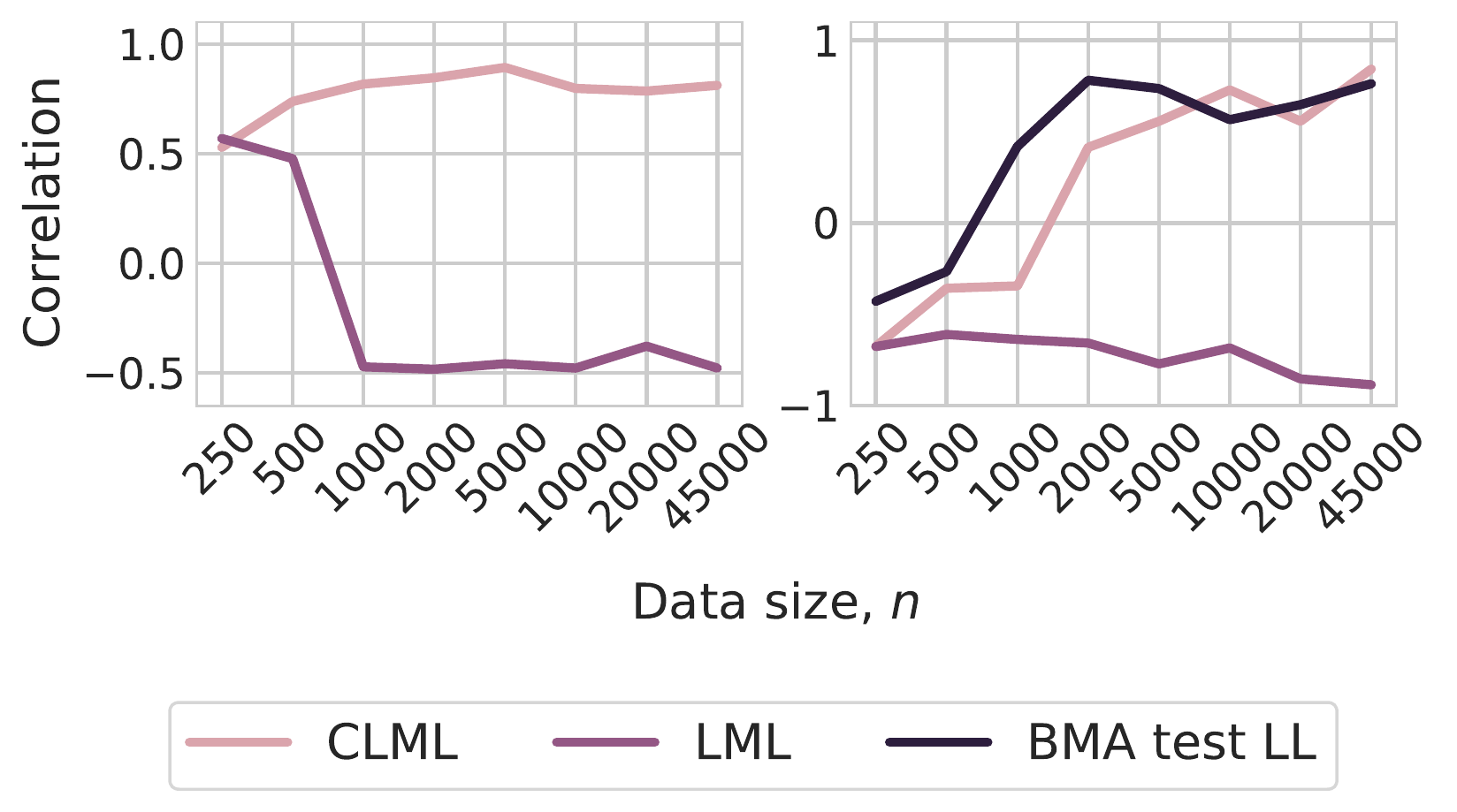}
        &
        \hspace{-0.2cm}\includegraphics[height=0.25\textwidth]{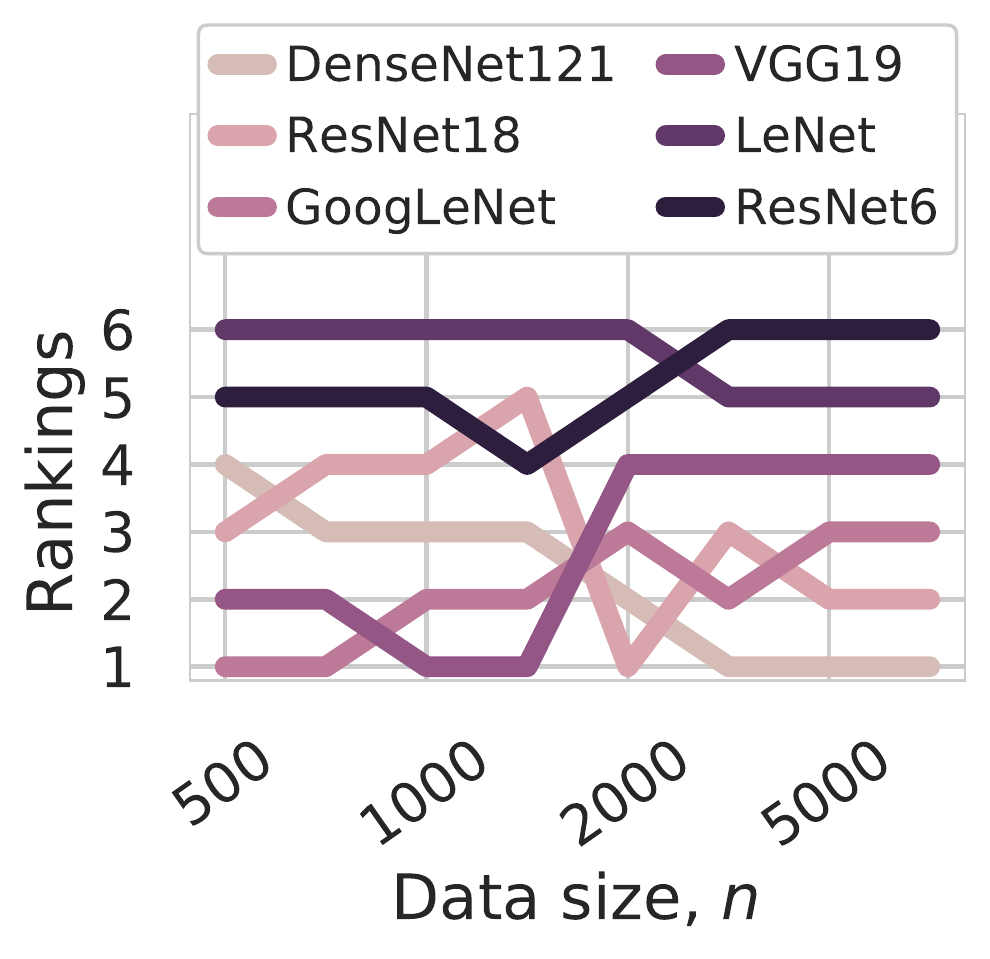}
      \\
        \hspace{-0.3cm}{\small \begin{tabular}{c}(a) Correlations with the BMA LL\\ and the inverse number of parameters
        \end{tabular}}
        & 
        \hspace{-0.cm}{\small (b) Rankings of DNNs} 
    \end{tabular}
\caption{
\textbf{Training speed, generalization, and LML for deep neural networks (DNN).}
\textbf{(a)}: \textbf{(Left)}: The correlation of the CLML and the LML with the BMA test log-likelihood. We show that the LML correlated positively with the BMA test log-likelihood (LL) for small sizes of the training data, but negatively for larger sizes, whereas CLML is correlated consistently positively with the BMA test log-likelihood for all sizes of the data. 
\textbf{(Right)}: The correlation of of the CLML, LML and the BMA test log-likelihood with the inverse number of parameters. The LML approximated with LA correlates negatively with number of parameters and assigns higher values to more constrained models. 
This negative correlation might reflect one of the sensitivity of the Laplace approximation to the number of parameters as discussed in Section~\ref{sec:la-in-dl}. 
\textbf{(b)}: Ranking of DNNs according to their Bayesian model average (BMA) test accuracy approximated with LA for different sizes of the CIFAR-10 training data. A lower ranking indicates a higher BMA test accuracy.
\texttt{VGG19} and \texttt{GoogLeNet}, in contrast with \texttt{ResNet-18} and \texttt{DenseNet121}, train faster but generalize worse for bigger sizes of the CIFAR-10 dataset.
}
\label{fig:nns-correlations}
\end{figure}

\textbf{Experimental details}\quad
We consider $6$ deep neural networks with architectures shown in Table~\ref{tab:neural-architectures} \citep{lecun1998gradient, simonyan2014very, szegedy2015going, he2016deep, huang2017densely}. We also consider $8$ different sizes of training datasets, $\{250, 500, 1000, 2000, 5000, 10,000, 20,000, 45,000\}$, each constructed by randomly sampling a subset of CIFAR-10.  To produce a MAP estimate, we train a neural network using hyperparameters found in Table \ref{tab:neural-hypers}.  All models are trained using SGD with weight decay coefficient $0.0005$, momentum coefficient of $0.9$, initial learning rate $0.1$, and learning rate drops by a factor of $10$ after $\frac{1}{2}$ and $\frac{3}{4}$ of the epochs.  
Data augmentations include random horizontal flips and crops.
We use the diagonal Laplace approximation to approximate the marginal likelihood and perform a Bayesian model average over $20$ samples to obtain the BMA test accuracy and log-likelihood. 
The CLML is computed using a $80\% - 20\%$ split of the training data as described in detail in Section~\ref{sec:app_clml_details}. 
We note that the BMA test accuracy and log-likelihood that we show in Figure~\ref{fig:nns-correlations} are computed with respect to all available training data and not just $80\%$ of it. 

\textbf{Discussion}\quad
Figure \ref{fig:nns-correlations} (b) shows the ranking of the models according to their generalization performance, where a lower ranking indicates a higher value of the BMA test accuracy. 
In particular, we see that \texttt{VGG19} and \texttt{GoogLeNet} train faster than \texttt{ResNet-18} and \texttt{DenseNet121} but generalize worse for bigger sizes of the CIFAR-10 dataset. 
This observation extends to many neural architectures that can perform better or worse depending on the size of the dataset \citep{dosovitskiy2020image}.
This proves that a faster training speed does not necessarily imply a better generalization performance.
Results in Figure \ref{fig:nns-correlations} (a) (left) are coherent with our conclusions from the Fourier example: the correlation of the BMA test log-likelihood with the LML is positive for small sizes of the training data and negative for higher sizes, whereas the correlation with CLML is consistently positive.
Finally, Figure \ref{fig:nns-correlations} (a) (right) shows that the LML approximated with LA heavily penalizes the number of parameters, which reflects the sensitivity of LA to the number of parameters as discussed in Sections~\ref{sec:la-in-dl}.

\begin{table}{t}
\parbox{.5\linewidth}{
        \centering
        \begin{tabular}{|l|l|}
            \hline
            Architecture & Number of parameters \\
            \hline
            \texttt{LeNet} & $62,006$ \\
            \hline
            \texttt{ResNet-6} & $609,354$ \\
            \hline
            \texttt{GoogLeNet} & $6,166,250$ \\
            \hline
            \texttt{DenseNet121} &  $6,956,298$\\
            \hline
            \texttt{VGG19} & $20,040,522$\\
            \hline
            \texttt{ResNet-18} & $11,173,962$ \\
            \hline
        \end{tabular}
        \caption{Neural architectures that we consider in Section~\ref{sec:speed}.}
        }
\hfill
\parbox{.5\linewidth}{
        \centering
        \begin{tabular}{|l|l|l|}
            \hline
            \# Samples & Epochs & Batch Size  \\
            \hline
            $250$ &  $900$ & $32$\\
            \hline
            $500$ &  $900$ & $32$\\
            \hline
            $1000$ &  $900$ & $32$ \\
            \hline
            $2000$ &  $600$ & $64$ \\
            \hline
            $5000$ &  $600$ & $64$\\
            \hline
            $10000$ &  $300$ & $128$ \\
            \hline
            $20000$ &  $300$ & $128$ \\
            \hline
            $45000$ &  $300$ & $128$\\
            \hline
        \end{tabular}
        \caption{Training hyperparameters for experiments that we consider for training neural networks in Section~\ref{sec:speed}.}
\label{tab:neural-hypers}
}
\label{tab:neural-architectures}
\end{table}

\section{The Density Estimation Example}
\label{sec:app_density_estimation}

\begin{figure*}[t]
\centering
    \begin{tabular}{ccc}
        \includegraphics[height=0.22\textwidth]{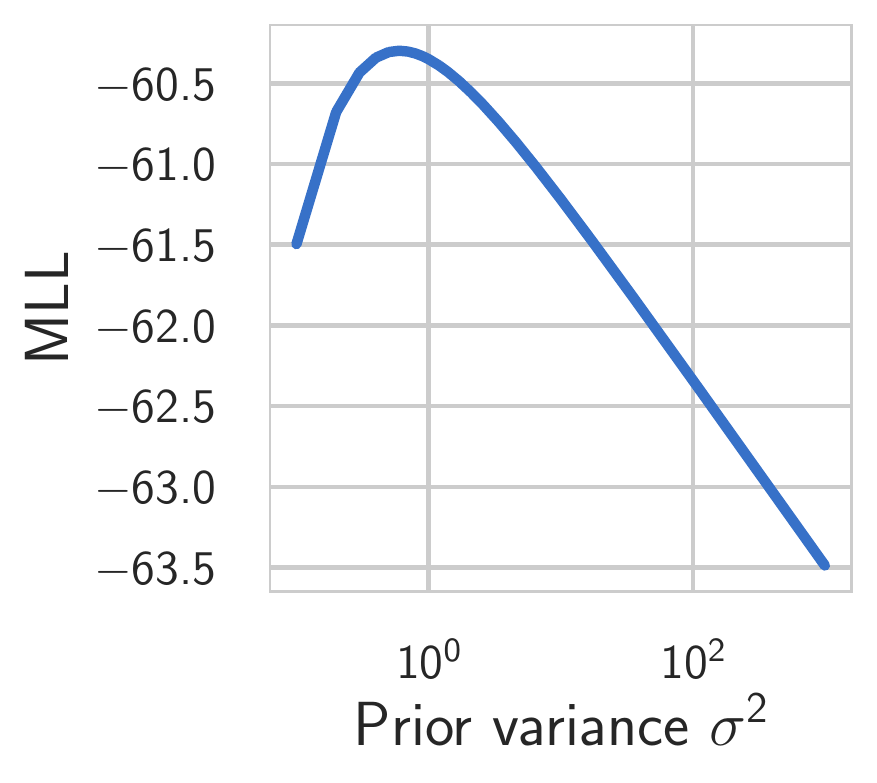}
        &
        \includegraphics[height=0.22\textwidth]{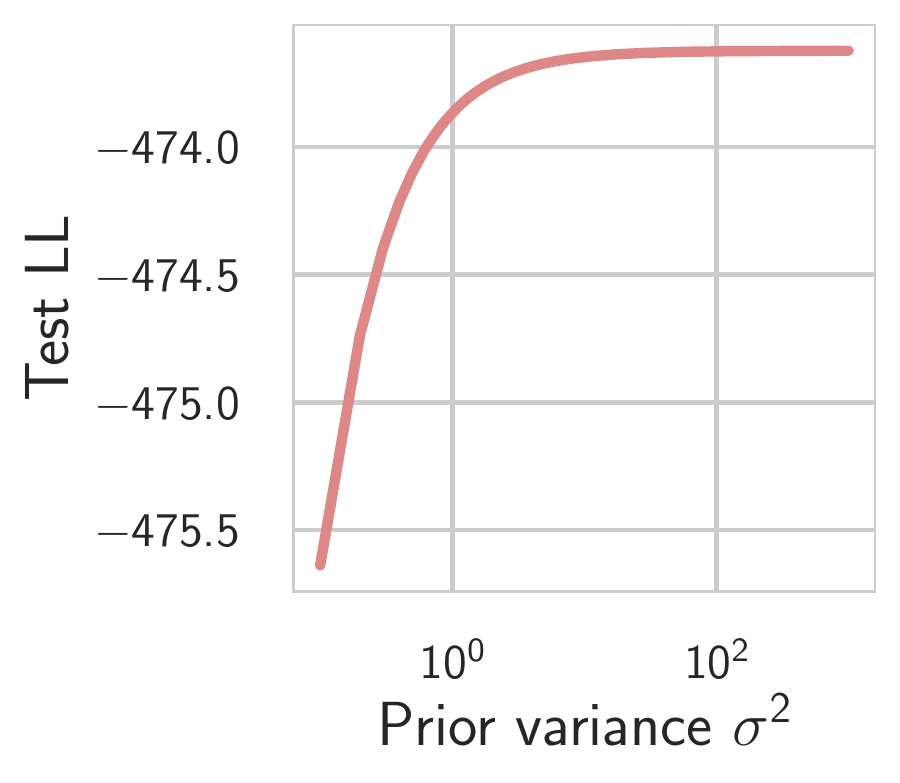}
        &
        \includegraphics[height=0.22\textwidth]{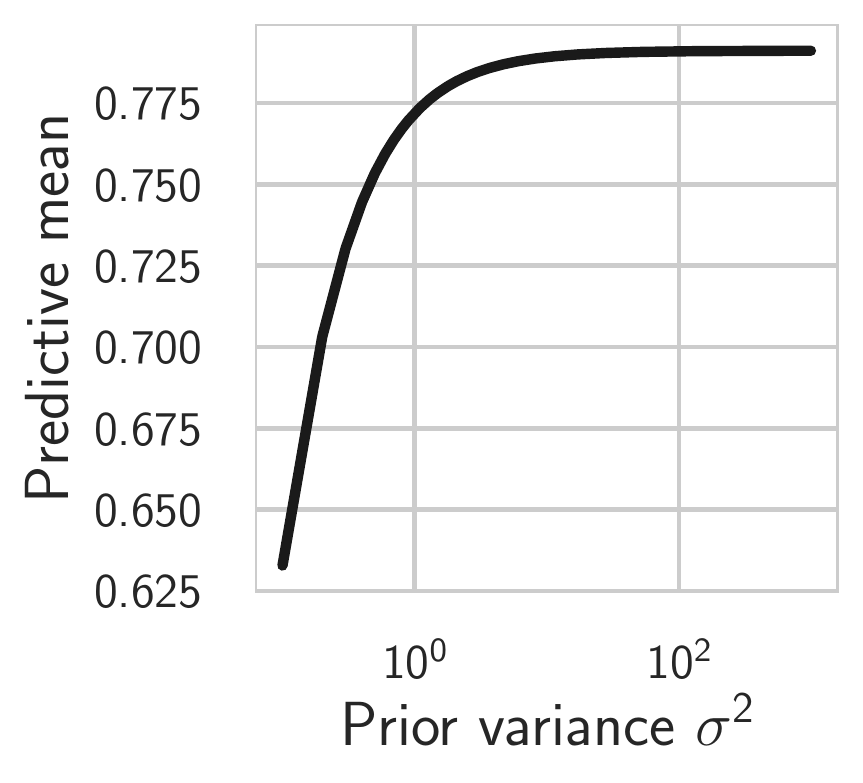}
        
        \\
        {\small (a) Marginal log-likelihood} & 
        {\small (b) Test log-likelihood} & 
        {\small (c) Predictive mean}
        \\[5mm]
        \multicolumn{3}{c}{
        \begin{tabular}{cc}
        \includegraphics[height=0.22\textwidth]{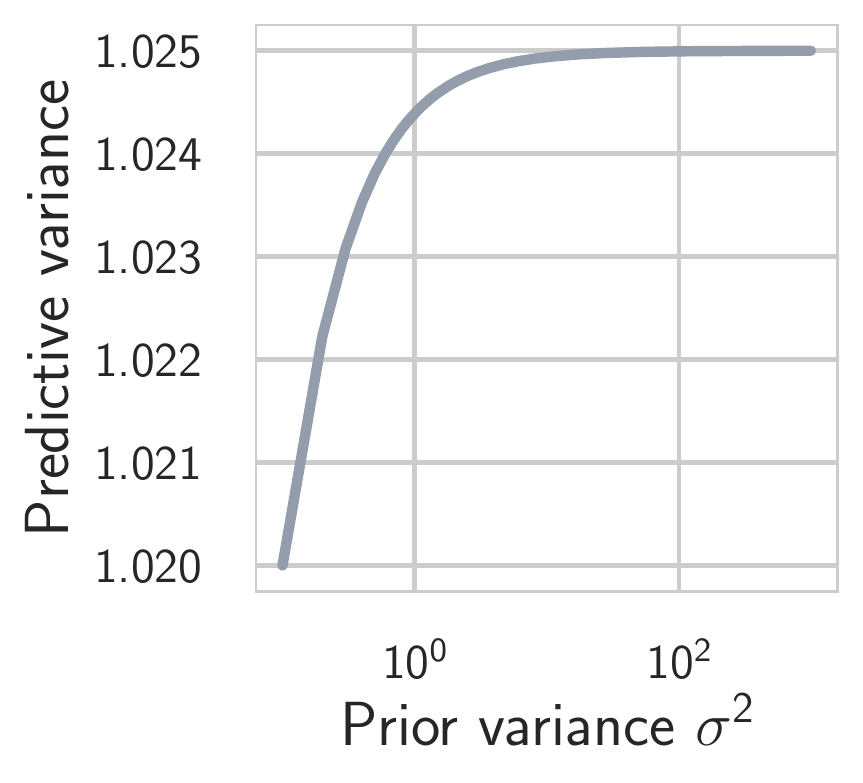}
        &
        \includegraphics[height=0.22\textwidth]{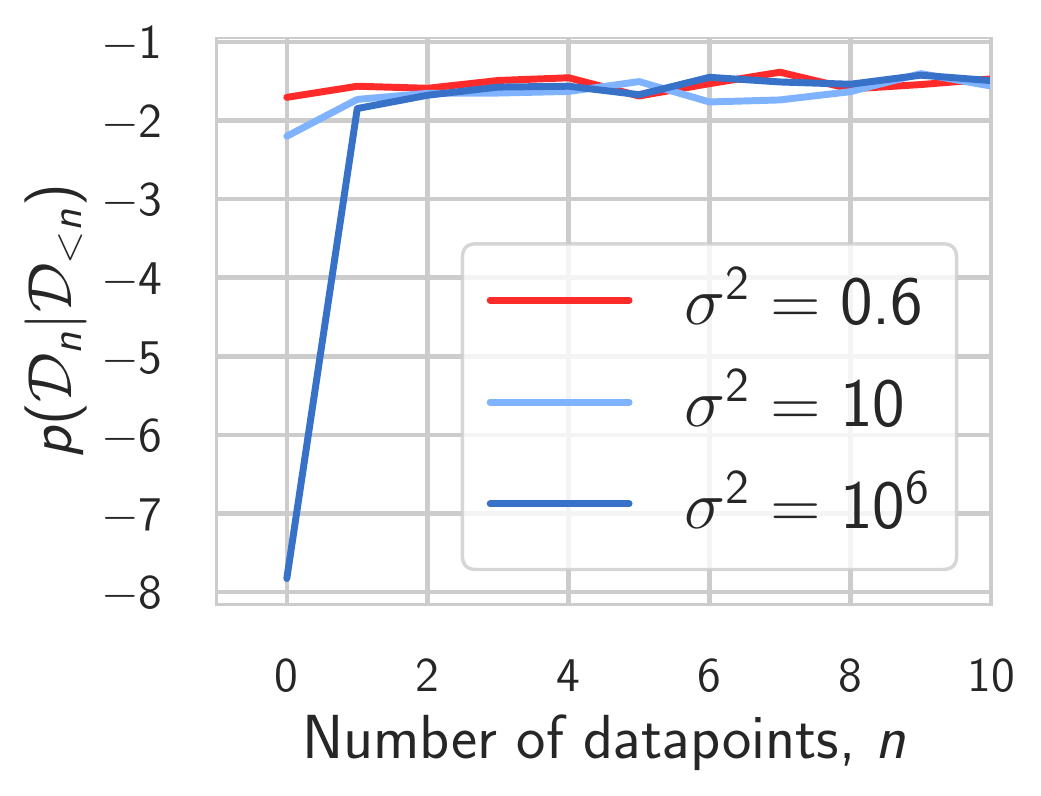}
        \\ 
        {\small (d) Predictive variance} & 
        {\small (e) Learning curves }
        \end{tabular}
        }
      
    \end{tabular}
\caption{
\textbf{Density estimation (details).}
\textbf{(a)}: Marginal log-likelihood, 
\textbf{(b)}: test log-likelihood and
\textbf{(c)}: mean and \textbf{(d)}: variance of the predictive distribution as a function of the prior variance $\sigma^2$.
The predictive distribution is virtually constant with respect to prior variance for $\sigma^2 > 10$, while marginal likelihood sharply decreases.
\textbf{(e)}: Learning curves for three different choices of prior standard deviation $\sigma^2$.
After observing $n = 5$ datapoints, the predictive distributions are almost indistinguishable between the different values of $\sigma^2$, but due to $\sigma^2 = 0.6$ providing the best fit for $n=1$ datapoint, marginal likelihood strongly prefers this choice.
}
\label{fig:gp_rq_details}
\end{figure*}

Consider the generative process where $x$ is generated using a Gaussian distribution $\mathcal{N}(u, 1)$ and the mean parameter is in turn generated
using a Gaussian distribution
$\mathcal{N}(\mu, \sigma^2)$.
Given dataset $D = \{x_i\}_{1}^{N}$, we can show that the marginal likelihood is equal to,
\begin{equation*}
    p(D | \sigma,  \mu ) = \mathcal{N} \left(\mu_N, I + \sigma^2 1_{N,N} \right),
\end{equation*}
where $\mu_N = \mu \times 1_N = \underbrace{\begin{bmatrix} \mu & \ldots & \mu \end{bmatrix}^T}_N$,
and $1_N \in \mathbb R^N$ is a column vector with all $N$ elements equal to $1$.
\begin{proof}
Indeed, we have $x_i = u + \epsilon_i$, where $\epsilon_i \sim \N(0, 1)$, and $u \sim \N(\mu, \sigma^2)$.
Thus, the observations $x_i$ are jointly Gaussian with a mean $\E x_i = \E(u + \epsilon_i) = \E u = \mu$.
The covariance structure is given by 
\begin{equation*}
\begin{split}
    \text{cov}(x_i, x_j) = \E(x_i - \mu)\cdot(x_j -\mu) = 
    \E (u + \epsilon_i - \mu)\cdot (u + \epsilon_j - \mu) = 
    \E (u - \mu)\cdot (u - \mu) + \E \epsilon_i \epsilon_j 
    \\
    = \text{cov}(u, u) + \text{cov}(\epsilon_i, \epsilon_j) =
    \sigma^2 + \delta_{ij},
\end{split}
\end{equation*}
where $\delta_{ij}$ is equal to $1$ if $i = j$ and $0$ otherwise.
Thus, we get $\D \sim \N(\mu_N, I + \sigma^2 1_{N,N})$.
\end{proof}

The posterior distribution is equal to,
\begin{equation}
\label{eq:app_density_post}
     p(u | D, \sigma,  \mu ) = \mathcal{N} \left(\frac{1}{1/\sigma^2 + N} \left(\sum_{i=1}^{N} x_i + \frac{1}{\sigma^2} \mu\right) \, , \, \frac{1}{1/\sigma^2 + N}\right).
\end{equation}
\begin{proof}
Let us denote $x = \begin{bmatrix}x_1 & \ldots & x_N \end{bmatrix}^T$.
We can write down the joint distribution over $x$ and $u$, following from the derivation marginal distribution of $x$ above:
\begin{equation*}
\begin{split}
    \begin{bmatrix} x \\ u  \end{bmatrix} \sim
     \N \left(
         \begin{bmatrix} \mu_N \\ \mu  \end{bmatrix},
         \begin{bmatrix}
            1_N 1_N^T \sigma^2 + I & 1_N \cdot \sigma^2 \\
            1_N^T \cdot \sigma^2 & \sigma^2
         \end{bmatrix}
     \right).
\end{split}
\end{equation*}
Using the properties of Gaussian distributions, we can compute the posterior $p(u \vert x)$ as a Gaussian conditional \citep[see e.g. Ch. 3 of][]{bishop2006pattern}:
\begin{equation}
\label{eq:app_density_post_derivation}
\begin{split}
    u~\vert~x \sim \N\bigg(
    \mu + 1_N^T \cdot \sigma^2 \cdot (1_N 1_N^T \sigma^2 + I)^{-1} \cdot (x - \mu_N),
    \sigma^2 - 1_N^T \cdot \sigma^2 \cdot (1_N 1_N^T \sigma^2 + I)^{-1} 1_N \cdot \sigma^2 
    \bigg).
\end{split}
\end{equation}
Now, note that $(1_N 1_N^T \sigma^2 + I)^{-1} = I - \frac{\sigma^2}{1 + N\sigma^2} 1_N 1_N^T$ which can be verified by direct multiplication.
Substituting the expression for the inverse in  Eq. \eqref{eq:app_density_post_derivation}, we recover Eq. \eqref{eq:app_density_post}.

\end{proof}

The predictive distribution is equal to,
\begin{equation}
\label{eq:app_density_pred}
    p(x^* |D, \sigma, \mu ) = \mathcal{N} \left(\frac{1}{1/\sigma^2 + N} \left(\sum_{i=1}^{N} x_i + \frac{1}{\sigma^2} \mu\right), 1 + \frac{1}{1/\sigma^2 + N}\right).
\end{equation}
\begin{proof}

Conditioned on $u$, the observations are Gaussian: $p(x^* | u) = \N(u, 1)$.
Furthermore, we have shown that the posterior $p(u \vert D, \sigma, \mu) = \N(\hat \mu, \hat \sigma^2)$ is also Gaussian, with the parameters $\hat \mu, \hat \sigma^2$ given by Eq. \eqref{eq:app_density_post}.
Then, the predictive distribution is simply $p(p(x^* |D, \sigma, \mu ) = \N(\hat \mu, \hat \sigma^2 + 1)$, recovering Eq. \eqref{eq:app_density_pred}.
\end{proof}

We see that as the variance of the prior mean $\sigma^2 \to +\infty$, both the predictive distribution and the posterior distribution do not depend on this hyperparameter, whereas the marginal likelihood depends on it. 
This is another example whereas the marginal likelihood is more sensitive to a hyperparameter that has little influence on the quality of future predictions.
Hence, the potential mismatch between marginal likelihood and generation.

\section{Neural Architecture Search Details}
\label{sec:app_cifar_search}
We investigate the correlation between the log marginal likelihood  (LML) and generalization in the context of image classification using the CIFAR-10 and CIFAR-100 datasets.
In particular, we consider two tasks: (1) model selection with fixed prior precision, and (2) tuning the prior precision then performing a similar model selection task. 

\textbf{Experimental details}\quad 
We reconstruct the neural architecture search experiments with convolutional (CNN) and residual (ResNet) networks for CIFAR-100 and CIFAR-10 from~\citet{immer2021scalable}. 
We use the same architectures: 

\begin{itemize}
    \item The CNNs consist of up to $5$ blocks of $3 × 3$ convolutions, followed by a ReLU activation function, and MaxPooling, except in the first layer. BatchNorm is replaced by the fixup initialization~\citep{zhang2019fixup} as in~\citet{immer2021scalable}. 
    The width (number of channels) of the first channel varies from $2$ to $32$ for both datasets. The last layer is a fully-connected layer to the class logit.
    \item ResNets of depths varying from $8$ to $32$ are used for CIFAR-10 and from $20$ to $101$ for CIFAR-100. The width varies from $16$ to $48$ for CIFAR-10 and from $32$ to $64$ for CIFAR-100. 
\end{itemize}

All models were trained for $250$ epochs with an SGD optimizer and an initial learning rate of $0.01$. 
The batch-size was fixed to $128$.
For experiments where the prior precision was optimized, we used online optimization where the prior precision was updated every $5$ epochs for $100$ iterations using an Adam optimizer with an initial learning rate equal to $1.0$. 

For all experiments in this section, we used the Kronecker Laplace approximation and computed the BMA test accuracy and log-likelihood by averaging over $20$ samples. 
The CLML was computed using a $80\% - 20\%$ split of the training data as described in detail in Section~\ref{sec:app_clml_details}. 
We note that the BMA test accuracy and log-likelihood that we show in all figures are computed using all available training data and not just $80\%$ of it that we condition CLML on. 

\begin{figure*}[t]
\centering
    \begin{tabular}{cc}
    \includegraphics[width=0.85\textwidth]{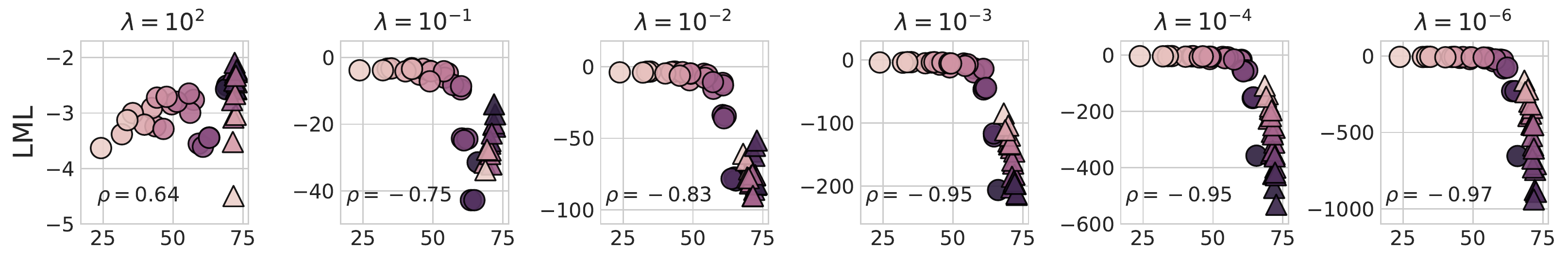} &
    \\
    \includegraphics[width=0.85\textwidth]{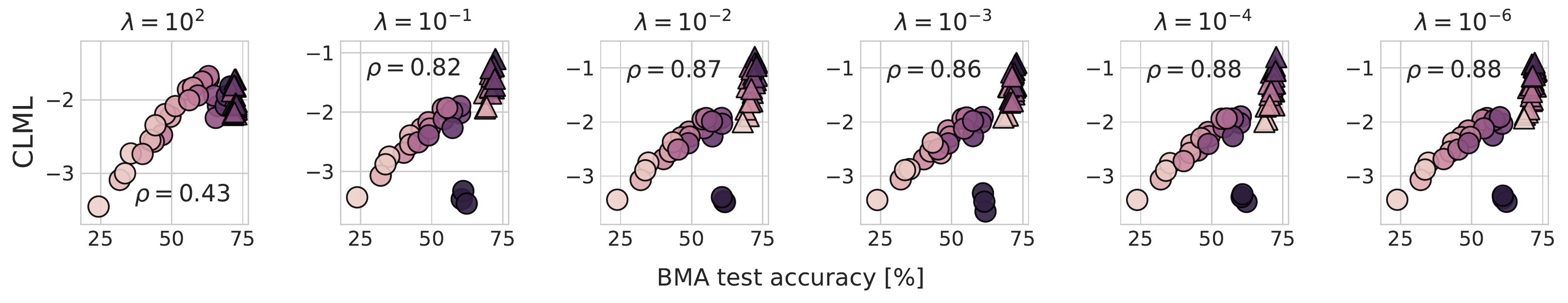} &
    \hspace{-0.1cm}\includegraphics[width=0.1\textwidth, trim={1.cm -2.0cm 0 0},clip]{figures/laplace_new/legend_colorbar.pdf}
    \end{tabular}
\caption{
\textbf{Neural architecture search for CIFAR-100.}
Visualization of the correlation between the model rankings according to different metrics for fixed prior precision $\lambda \in \{10^{2}, 10^{-1}, 10^{-2}, 10^{-3}, 10^{-4}, 10^{-6} \}$. We report the Spearman’s correlation coefficient $\rho$ in each figure.
\textbf{(Top row)}: Correlation between the BMA test accuracy and the log marginal likelihood (LML).
\textbf{(Top row)}: Correlation between the BMA test accuracy and the conditional log marginal likelihood (CLML).
The LML correlates positively with the BMA test accuracy for high values of the prior precision, and negatively for low values of the prior precision (vague priors). 
The CLML on the other hand is less sensitive to the value of the prior precision and consistently achieves a positive correlation with the BMA test accuracy. 
}
\label{fig:fixed-wd-cifar100-bma-tacc}
\end{figure*}

\begin{figure*}[t]
\centering
    \begin{tabular}{cc}
    \includegraphics[width=0.85\textwidth]{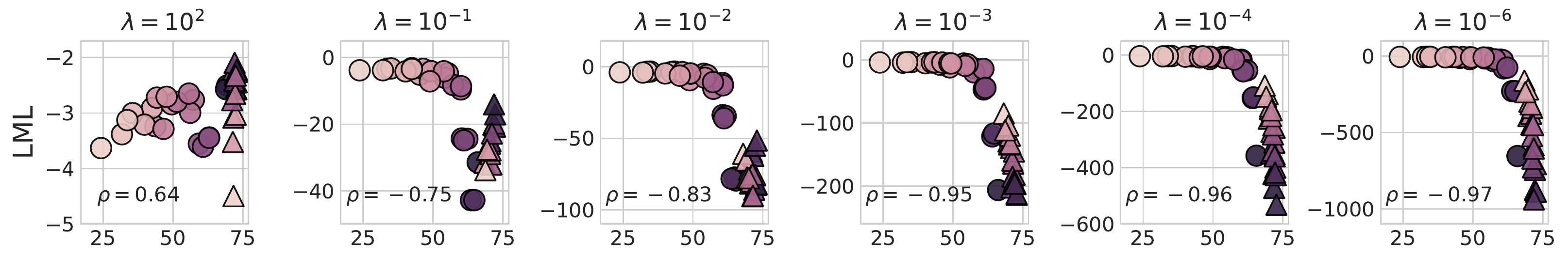} &
    \\
    \includegraphics[width=0.85\textwidth]{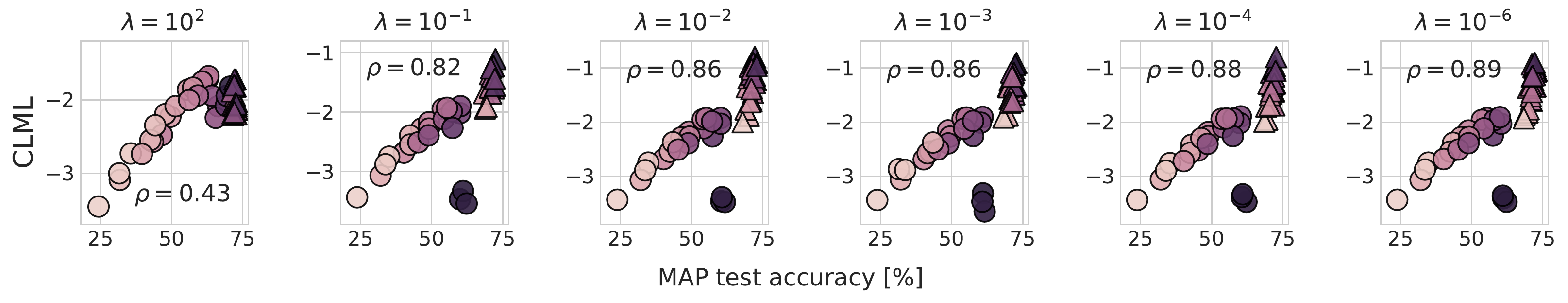} &
    \hspace{-0.1cm}\includegraphics[width=0.1\textwidth, trim={1.cm -2.0cm 0 0},clip]{figures/laplace_new/legend_colorbar.pdf}
    \end{tabular}
\caption{
\textbf{Neural architecture search for CIFAR-100.}
Visualization of the correlation between the model rankings according to different metrics for fixed prior precision $\lambda \in \{10^{2}, 10^{-1}, 10^{-2}, 10^{-3}, 10^{-4}, 10^{-6} \}$. We report the Spearman’s correlation coefficient $\rho$ in each figure.
\textbf{(Top row)}: Correlation between the maximum-a-posterior (MAP) test accuracy and the LML. 
\textbf{(Top row)}: Correlation between the MAP test accuracy and the CLML.
The LML correlates positively with the MAP test accuracy for high values of the prior precision, and negatively for low values of the prior precision, which correspond to vague priors.
The CLML on the other hand is less sensitive to the value of the prior precision and consistently achieves a positive correlation with the MAP test accuracy.
}
\label{fig:fixed-wd-cifar100-map-tacc}
\end{figure*}

\begin{figure*}[t]
\centering
    \begin{tabular}{cc}
    \includegraphics[width=0.85\textwidth]{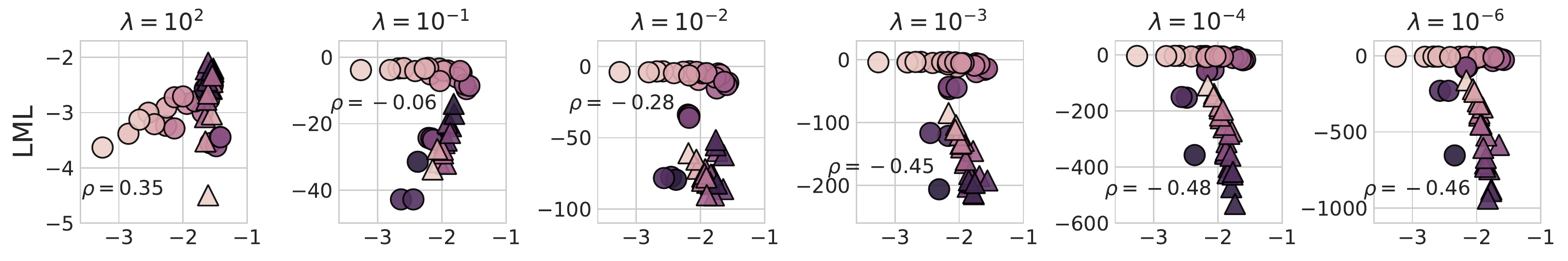} &
    \\
    \includegraphics[width=0.85\textwidth]{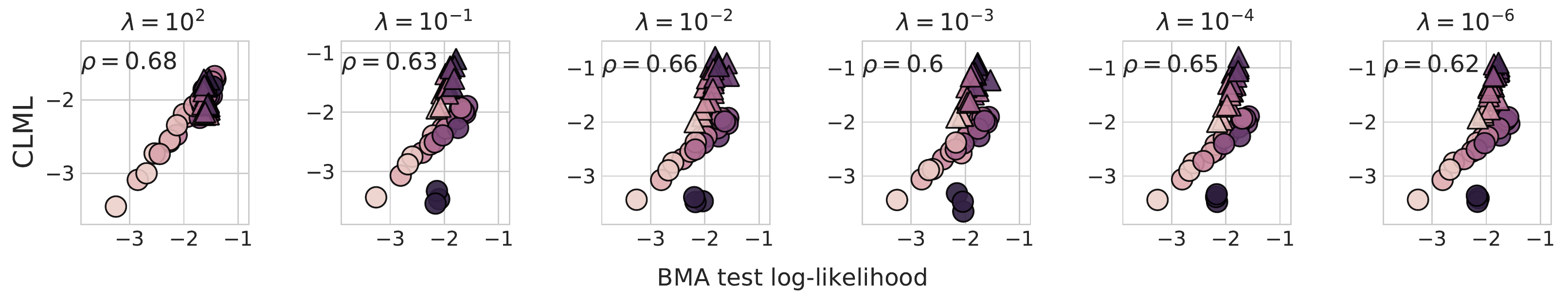} &
    \hspace{-0.1cm}\includegraphics[width=0.1\textwidth, trim={1.cm -2.0cm 0 0},clip]{figures/laplace_new/legend_colorbar.pdf}
    \end{tabular}
\caption{
\textbf{Neural architecture search for CIFAR-100.}
Visualization of the correlation between the model rankings according to different metrics for fixed prior precision $\lambda \in \{10^{2}, 10^{-1}, 10^{-2}, 10^{-3}, 10^{-4}, 10^{-6} \}$. We report the Spearman’s correlation coefficient $\rho$ in each figure.
\textbf{(Top row)}: Correlation between the BMA test log-likelihood and the log marginal likelihood (LML). 
\textbf{(Top row)}: Correlation between the BMA test log-likelihood and the conditional log marginal likelihood (CLML).
The LML correlates positively with the BMA test log-likelihood for high values of the prior precision, and negatively for low values of the prior precision (vague priors). 
The correlation shifts around $\lambda=10^{-1}$ as it remains positive for ResNets but becomes negative for CNNs. 
The CLML on the other hand is less sensitive to the value of the prior precision and consistently achieves a positive correlation with the BMA test log-likelihood. 
}
\label{fig:fixed-wd-cifar100-bma-tll}
\end{figure*}

\begin{figure*}[t]
\centering
    \begin{tabular}{cc}
    \includegraphics[width=0.85\textwidth]{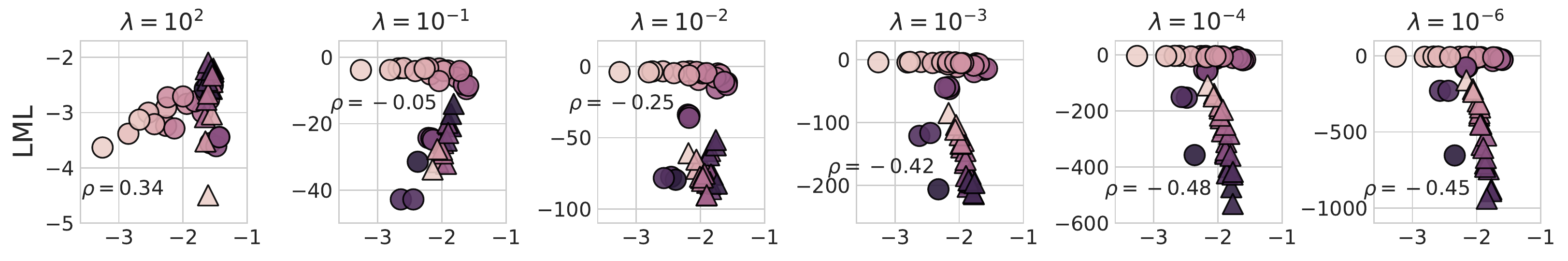} &
    \\
    \includegraphics[width=0.85\textwidth]{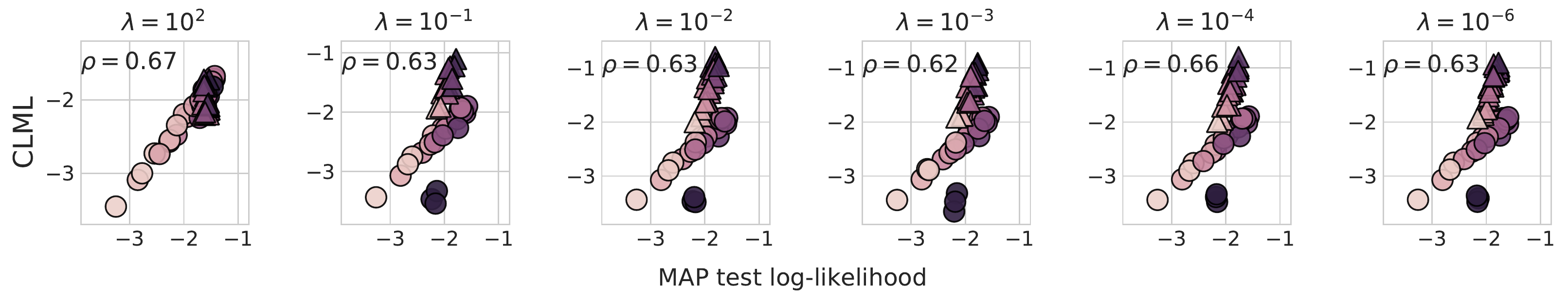} &
    \hspace{-0.1cm}\includegraphics[width=0.1\textwidth, trim={1.cm -2.0cm 0 0},clip]{figures/laplace_new/legend_colorbar.pdf}
    \end{tabular}
\caption{
\textbf{Neural architecture search for CIFAR-100.}
Visualization of the correlation between the model rankings according to different metrics for fixed prior precision $\lambda \in \{10^{2}, 10^{-1}, 10^{-2}, 10^{-3}, 10^{-4}, 10^{-6} \}$. We report the Spearman’s correlation coefficient $\rho$ in each figure.
\textbf{(Top row)}: Correlation between the MAP test log-likelihood and the log marginal likelihood (LML). 
\textbf{(Top row)}: Correlation between the MAP test log-likelihood and the conditional log marginal likelihood (CLML).
The LML correlates positively with the MAP test log-likelihood for high values of the prior precision, and negatively for low values of the prior precision (vague priors).
We can that the correlation shift occurs around $\lambda=10^{-1}$ as the correlation remains positive for ResNets but becomes negative for CNNs.
The CLML on the other hand is less sensitive to the value of the prior precision and consistently achieves a positive correlation with the MAP test log-likelihood.
}
\label{fig:fixed-wd-cifar100-map-tll}
\end{figure*}

\begin{figure*}[t]
\centering
    \begin{tabular}{cc}
    \includegraphics[width=0.85\textwidth]{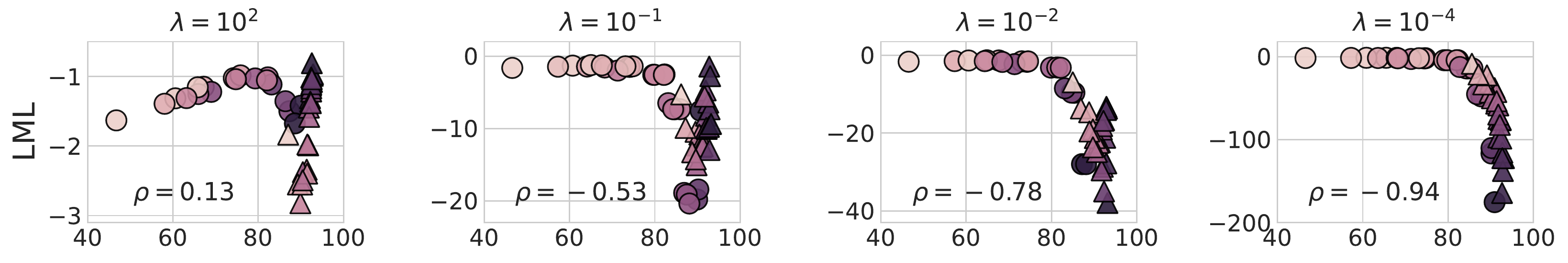} &
    \\
    \includegraphics[width=0.85\textwidth]{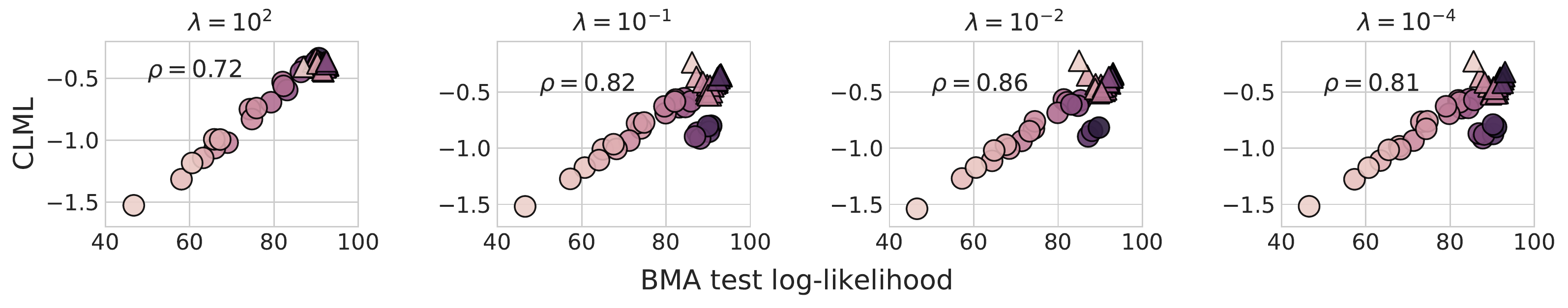} &
    \hspace{-0.1cm}\includegraphics[width=0.1\textwidth, trim={1.cm -2.0cm 0 0},clip]{figures/laplace_new/legend_colorbar.pdf}
    \end{tabular}
\caption{
\textbf{Neural architecture search for CIFAR-10.}
Visualization of the correlation between the model rankings according to different metrics for fixed prior precision $\lambda \in \{10^{2}, 10^{-1}, 10^{-2}, 10^{-4}\}$. We report the Spearman’s correlation coefficient $\rho$ in each figure. 
\textbf{(Top row)}: Correlation between the BMA test accuracy and the log marginal likelihood (LML).
\textbf{(Top row)}: Correlation between the BMA test accuracy and the conditional log marginal likelihood (CLML).
The LML correlates positively with the BMA test accuracy for high values of the prior precision, and negatively for low values of the prior precision (vague priors). 
The CLML on the other hand is less sensitive to the value of the prior precision and consistently achieves a positive correlation with the BMA test accuracy.
}
\label{fig:fixed-wd-cifar10-bma-tacc}
\end{figure*}

\begin{figure*}[t]
\centering
    \begin{tabular}{cc}
    \includegraphics[width=0.85\textwidth]{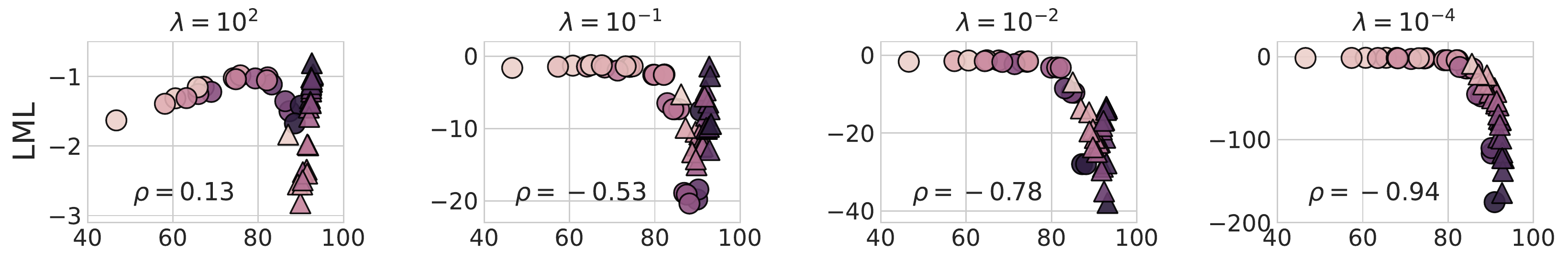} &
    \\
    \includegraphics[width=0.85\textwidth]{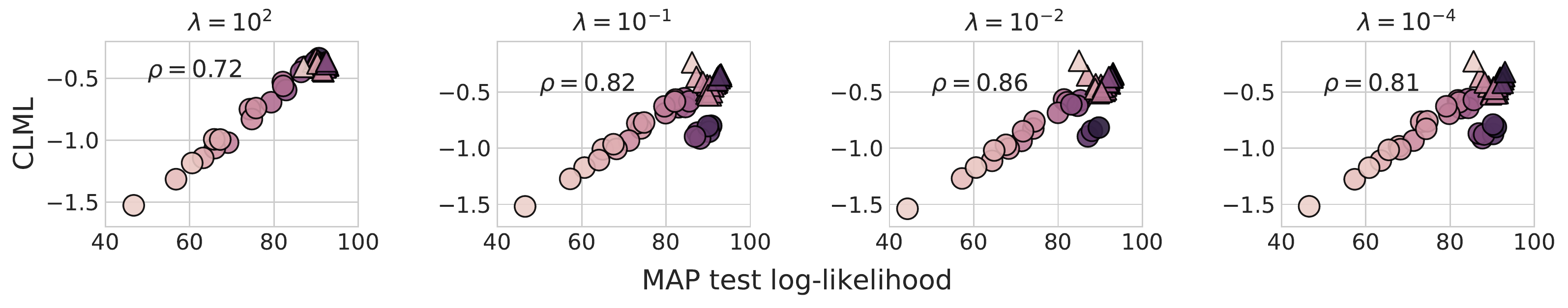} &
    \hspace{-0.1cm}\includegraphics[width=0.1\textwidth, trim={1.cm -2.0cm 0 0},clip]{figures/laplace_new/legend_colorbar.pdf}
    \end{tabular}
\caption{
\textbf{Neural architecture search for CIFAR-10.}
Visualization of the correlation between the model rankings according to different metrics for fixed prior precision $\lambda \in \{10^{2}, 10^{-1}, 10^{-2}, 10^{-4}\}$. We report the Spearman’s correlation coefficient $\rho$ in each figure.
\textbf{(Top row)}: Correlation between the maximum-a-posterior (MAP) test accuracy and the LML. 
\textbf{(Top row)}: Correlation between the MAP test accuracy and the CLML.
The LML correlates positively with the MAP test accuracy for high values of the prior precision, and negatively for low values of the prior precision, which correspond to vague priors.
The CLML on the other hand is less sensitive to the value of the prior precision and consistently achieves a positive correlation with the MAP test accuracy.
}
\label{fig:fixed-wd-cifar10-map-tacc}
\end{figure*}

\begin{figure*}[t]
\centering
    \begin{tabular}{cc}
    \includegraphics[width=0.85\textwidth]{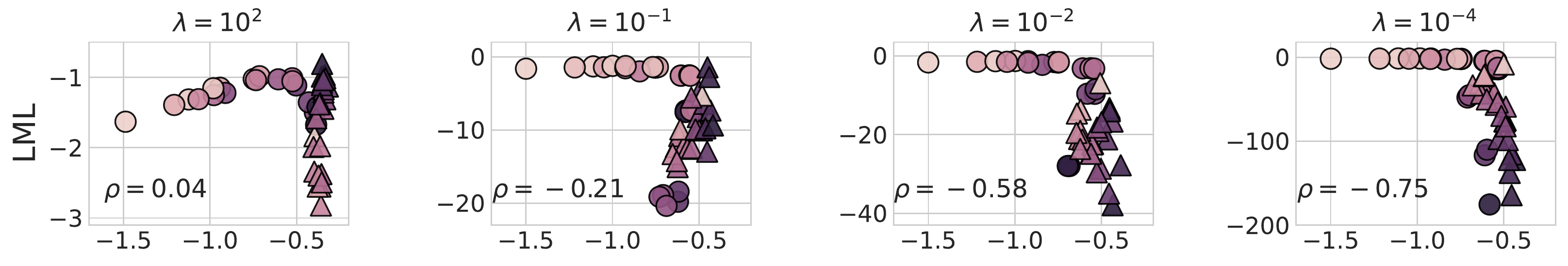} &
    \\
    \includegraphics[width=0.85\textwidth]{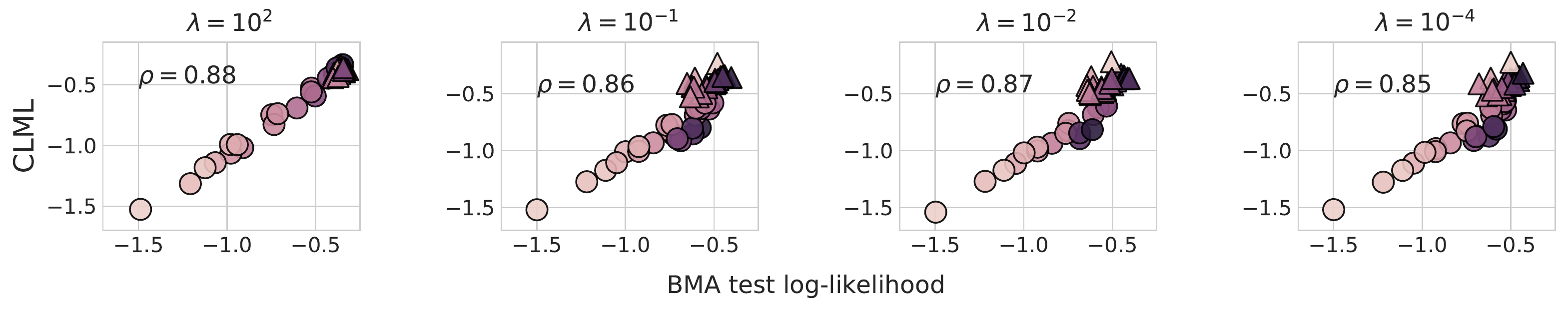} &
    \hspace{-0.1cm}\includegraphics[width=0.1\textwidth, trim={1.cm -2.0cm 0 0},clip]{figures/laplace_new/legend_colorbar.pdf}
    \end{tabular}
\caption{
\textbf{Neural architecture search for CIFAR-10.}
Visualization of the correlation between the model rankings according to different metrics for fixed prior precision$\lambda \in \{10^{2}, 10^{-1}, 10^{-2}, 10^{-4}\}$. We report the Spearman’s correlation coefficient $\rho$ in each figure.
\textbf{(Top row)}: Correlation between the BMA test log-likelihood and the log marginal likelihood (LML). 
\textbf{(Top row)}: Correlation between the BMA test log-likelihood and the conditional log marginal likelihood (CLML).
The LML almost does not correlate with the BMA test log-likelihood for high values of the prior precision, but shows a negative correlation for low values of the prior precision (vague priors). 
The correlation shifts around $\lambda=10^{-1}$ as it remains positive for ResNets but becomes negative for CNNs. 
The CLML on the other hand is less sensitive to the value of the prior precision and consistently achieves a positive correlation with the BMA test log-likelihood.
}
\label{fig:fixed-wd-cifar10-bma-tll}
\end{figure*}

\begin{figure*}[t]
\centering
    \begin{tabular}{cc}
    \includegraphics[width=0.85\textwidth]{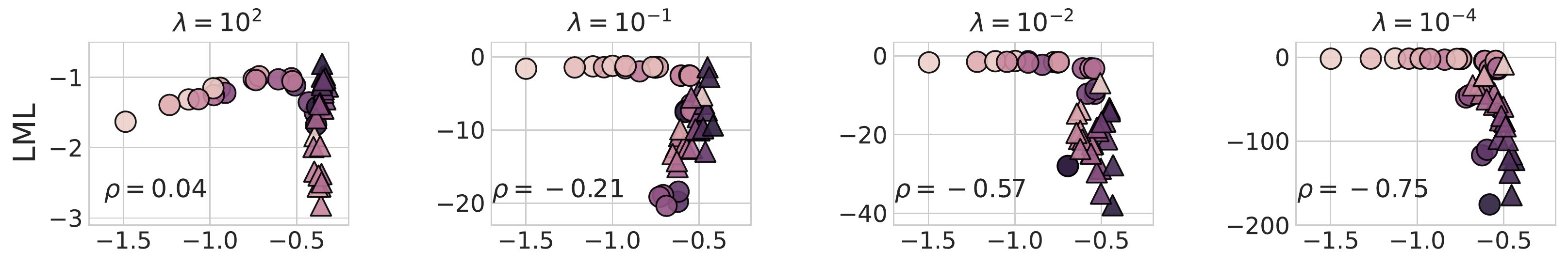} &
    \\
    \includegraphics[width=0.85\textwidth]{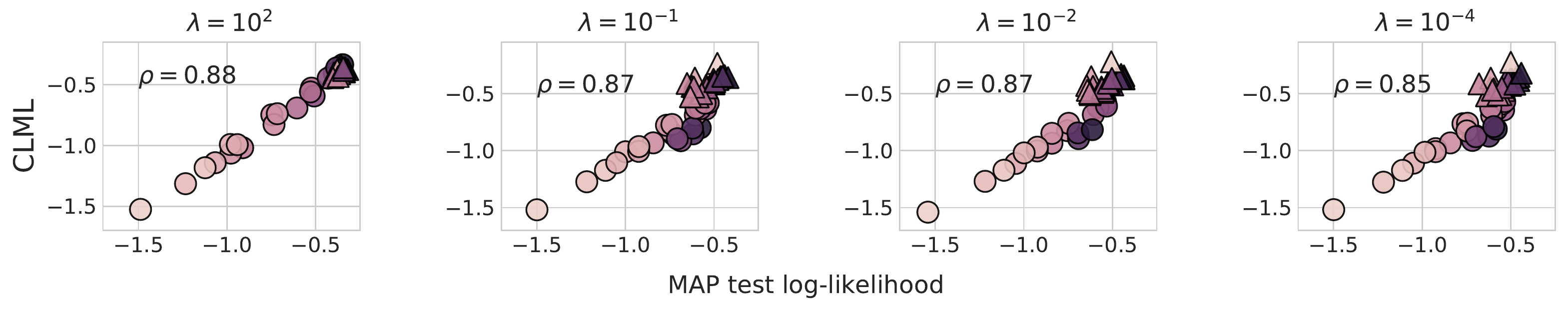} &
    \hspace{-0.1cm}\includegraphics[width=0.1\textwidth, trim={1.cm -2.0cm 0 0},clip]{figures/laplace_new/legend_colorbar.pdf}
    \end{tabular}
\caption{
\textbf{Neural architecture search for CIFAR-10.}
Visualization of the correlation between the model rankings according to different metrics for fixed prior precision $\lambda \in \{10^{2}, 10^{-1}, 10^{-2}, 10^{-4}\}$. We report the Spearman’s correlation coefficient $\rho$ in each figure.
\textbf{(Top row)}: Correlation between the MAP test log-likelihood and the log marginal likelihood (LML). 
\textbf{(Top row)}: Correlation between the MAP test log-likelihood and the conditional log marginal likelihood (CLML).
The LML almost does not correlate with the MAP test log-likelihood for high values of the prior precision, but shows a negative correlation for low values of the prior precision (vague priors). 
We can that the correlation shift occurs around $\lambda=10^{-1}$ as the correlation remains positive for ResNets but becomes negative for CNNs.
The CLML on the other hand is less sensitive to the value of the prior precision and consistently achieves a positive correlation with the MAP test log-likelihood. 
}
\label{fig:fixed-wd-cifar10-map-tll}
\end{figure*}

\textbf{Discussion}\quad
We visualize the correlation of the LML and the CLML in the top and bottom rows of  Figure~\ref{fig:fixed-wd-cifar100-bma-tacc}, respectively for CIFAR-100.   
We also report the Spearman’s correlation coefficient $\rho$ \citep{spearman1961proof}, which measures the correlation between the model rankings according to the BMA test accuracy and the LML/CLML.
We see that the LML correlates positively with the BMA test accuracy for high values of the prior precision, but negatively for lower values.
This can be understood in the light of what we discussed in Section~\ref{sec: notgen}; the LML penalizes low values of the prior precision because they correspond to diffuse priors.
A similar trend is observed in Figure~\ref{fig:fixed-wd-cifar100-map-tacc} for the MAP test accuracy, Figure~\ref{fig:fixed-wd-cifar100-bma-tll} for the BMA test log-likelihood, and  Figure~\ref{fig:fixed-wd-cifar100-map-tll} for the MAP test log-likelihood for the CIFAR-100 dataset. Similar results are obtained for CIFAR-10 in Figures~\ref{fig:fixed-wd-cifar10-bma-tacc}, ~\ref{fig:fixed-wd-cifar10-map-tacc}, ~\ref{fig:fixed-wd-cifar10-bma-tll}, ~\ref{fig:fixed-wd-cifar10-map-tll}.

\begin{figure}[h]
\centering
    \begin{tabular}{ccc}
        \hspace{-0.2cm}\includegraphics[height=0.2\textwidth]{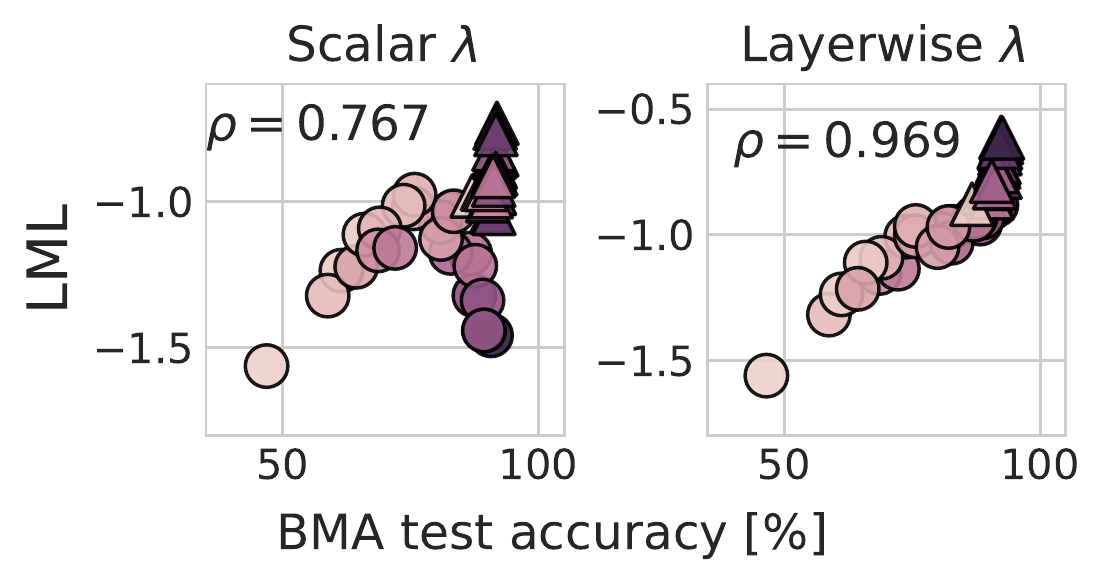}
        &
        \hspace{-0.cm}\includegraphics[height=0.2\textwidth]{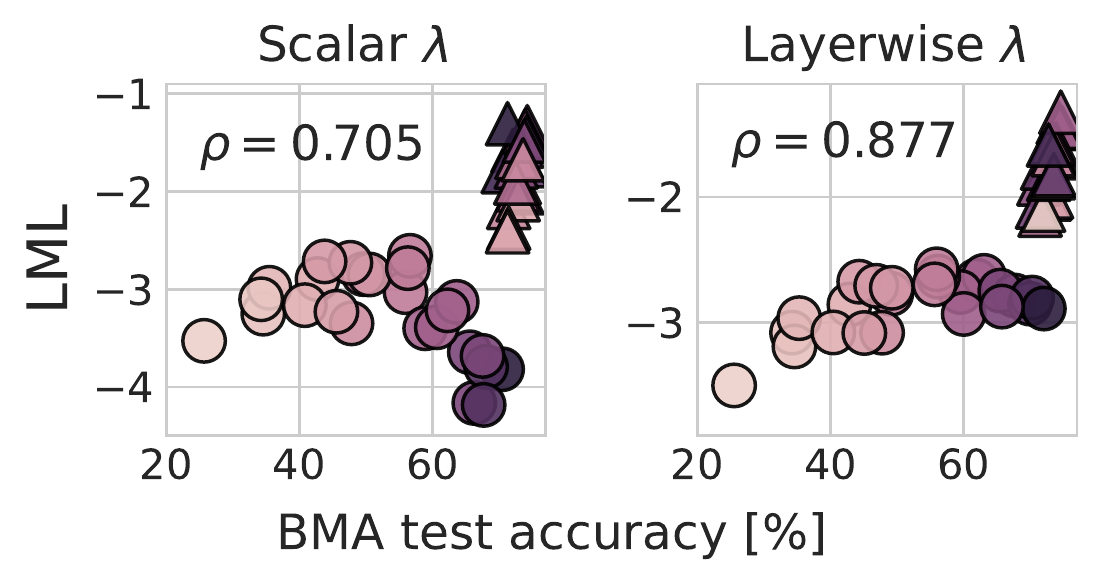}
        &    \hspace{-0.1cm}\includegraphics[width=0.1\textwidth, trim={1.cm -2.0cm 0 0},clip]{figures/laplace_new/legend_colorbar.pdf}
      \\
        \hspace{-0.3cm}{\small (a) CIFAR-10} & 
        \hspace{-0.cm}{\small (b) CIFAR-100} 
    \end{tabular}
\caption{
\textbf{Neural network hyperparameter optimization for CIFAR-10 and CIFAR-100.}
Correlation between the log marginal likelihood (LML) and the BMA test accuracy for \textbf{(Left)} optimized global prior precision, and \textbf{(Right)} optimized layerwise prior precision for CIFAR-10. 
We report the Spearman’s correlation coefficient $\rho$ in each figure.
We observe that the layerwise optimization further improves the correlation between the LML and the BMA test accuracy. 
}
\label{fig:app-la-optimized-precision}
\end{figure}

Figure~\ref{fig:app-la-optimized-precision} shows that optimizing the global and layerwise prior precision helps improve the correlation between LML and the BMA test accuracy.
To understand this effect, consider Figure~\ref{fig:learning-curves-sec5}(a):
two models with the same test performance can have very different values of the marginal likelihood depending on the prior variance. 
However, if we update the prior variance such that it maximizes the LML, we can expect that the final prior variance to be low without a major effect on the accuracy, therefore leading to a positive correlation between the LML and the BMA test accuracy. 

\begin{figure*}[h!]
\centering
    \begin{tabular}{cccc}
        \hspace{-0.3cm}\includegraphics[height=0.23\textwidth]{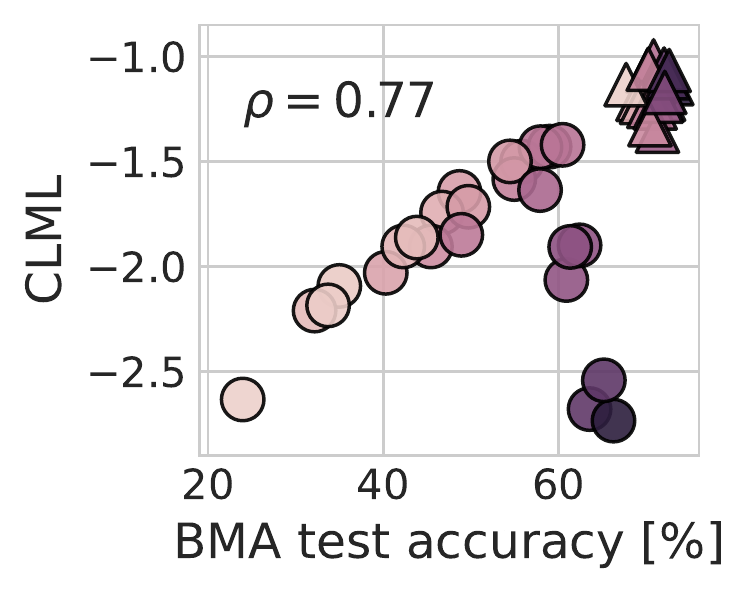}
        &
        \hspace{-0.3cm}\includegraphics[height=0.23\textwidth]{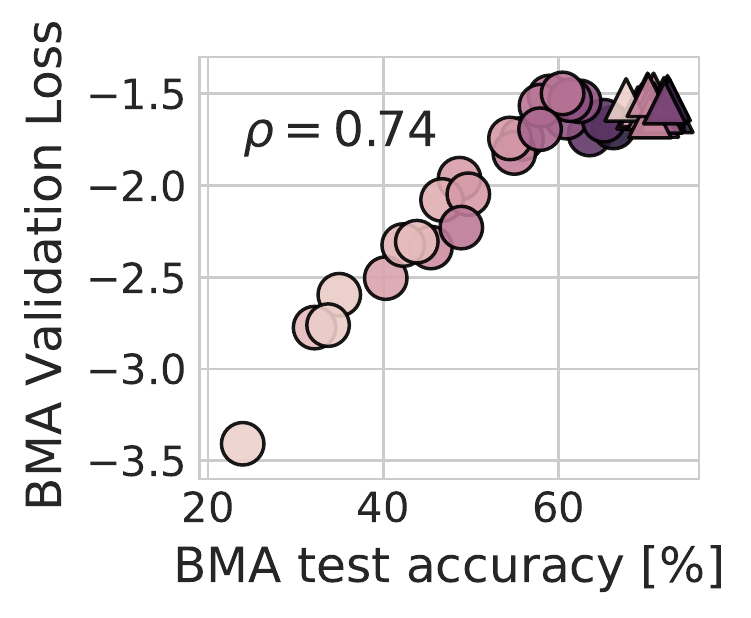}
        &
        \hspace{-0.3cm}\includegraphics[height=0.23\textwidth]{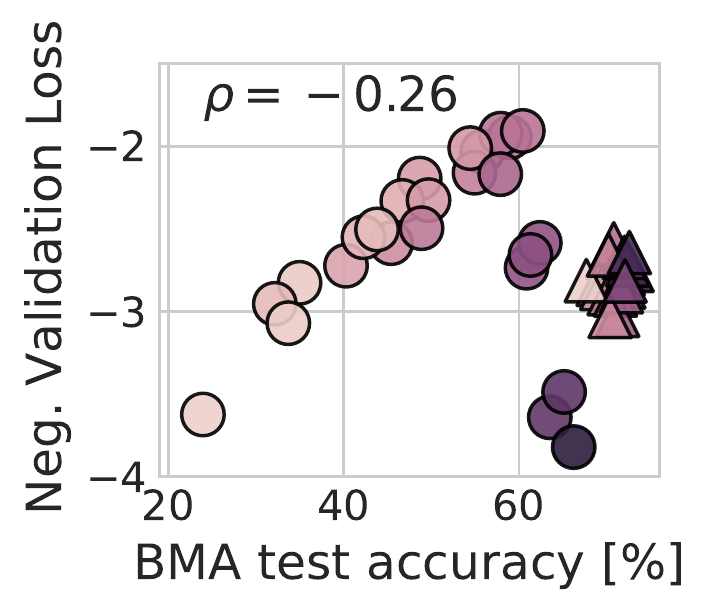}
        &
        \hspace{-0.35cm}\includegraphics[width=0.1\textwidth, trim={1.cm -2.0cm 0 0},clip]{figures/laplace_new/legend_colorbar.pdf}
      \\[-0.1cm]
        \hspace{-0.3cm}{\small (a) CLML vs BMA accuracy} & 
        \hspace{-0.3cm}{\small 
        \begin{tabular}{c}(b)  BMA validation loss\\ vs BMA accuracy
        \end{tabular} }
        &
        \hspace{-0.3cm}{\small 
        \begin{tabular}{c}(c)  Neg. of the validation loss\\ vs BMA accuracy
        \end{tabular} }
        \\ [-0.2cm]
    \end{tabular}
\caption{
\textbf{Neural hyperparameter optimization for CIFAR-100.}
The correlation (Spearman's $\rho$) between the model rankings and generalization. For panels \textbf{(a)}, \textbf{(b)}, and \textbf{(c)} we consider a fixed prior precision $\lambda = 10^{-1}$. 
\textbf{(a)}: Correlation between the BMA test accuracy and the CLML.
\textbf{(b)} Correlation between the BMA test accuracy and the BMA validation loss.
\textbf{(c)} Correlation between the BMA test accuracy and the negative of the validation loss.
While both the CLML and the BMA validation loss correlate positively with the BMA test accuracy, we observe that
the negative validation loss correlates positively with the BMA test accuracy for CNNs but not for ResNets, hence correlating negatively overall with the BMA test accuracy for CIFAR-100.  
}
\label{fig:clml-valid}
\end{figure*}

\textbf{Comparison to the validation loss}\quad \cref{fig:clml-valid} shows the correlation between the BMA test accuracy and: the CLML (a), the BMA validation loss (b), and the MAP validation loss (c).
The negative MAP validation loss correlates positively overall with the BMA test accuracy for CNNs, but not for ResNets, resulting in a negative total correlation for low prior precision values.
In contrast, the CLML and BMA validation loss exhibit a positive correlation with the BMA test accuracy.
This difference in the correlation factors of the BMA validation loss as opposed to the MAP validation loss might be due to the fact that the BMA solution tends to be generally less overconfident and better calibrated than MAP solution, hence increasing its correlation with the BMA test accuracy.

\section{Model Calibration and CLML}
\label{sec:app_calibration}

In this Section, we explain the difference between the CLML and the test (or validation) likelihood.
The CLML is a likelihood under the joint predictive distribution, i.e. it takes into account the dependence in the predictions on the different datapoints.
The test likelihood on the other hands treats datapoints as independent, and only evaluates the marginal predictive distributions.

Suppose the test set consists of two datapoints $\dataset = \{d_1, d_2\}$,
and suppose we have two parameter samples $w_1$ and $w_2$. Suppose for example that the predictive likelihoods are given by
$p(d_1 \vert w_1) = 0.1,~p(d_2 \vert w_1) = 0.9,~p(d_1 \vert w_2) = 0.9,~p(d_2 \vert w_2) = 0.1$.
Then the joint likelihood estimate will be given by
\begin{equation}
    \label{eq:app_joint}
    p(d_1, d_2) =
    \mathbb E_{w} p(d_1\vert w) \cdot p(d_2\vert w)
    \approx
    \frac 1 2 \left ( 
    p(d_1 \vert w_1) \cdot p(d_2 \vert w_1) + 
    p(d_1 \vert w_2) \cdot p(d_2 \vert w_2)
    \right )
    = 0.09.
\end{equation}
The average likelihood of a test datapoint will on the other hand be
\begin{equation}
    \label{eq:app_marginal}
    \begin{split}
    p(d_1) \cdot p(d_2) =
    \mathbb E_{w} p(d_1\vert w) &\cdot \mathbb E_{w} p(d_2\vert w)
    \approx
    \\
    &\frac 1 2 \left ( 
    p(d_1 \vert w_1) +  p(d_1 \vert w_2)
    \right) \cdot
    \frac 1 2 \left ( 
    p(d_2 \vert w_1) +  p(d_2 \vert w_2)
    \right)
    = 0.25.
    \end{split}
\end{equation}
On the other hand, if all samples provide equal likelihood
$p(d_1 \vert w_1) = p(d_2 \vert w_1) = p(d_1 \vert w_2) = p(d_2 \vert w_2) = 0.5$, both Eq.~\eqref{eq:app_joint} and Eq.~\eqref{eq:app_marginal} result in $0.25$.

In other words, the two configurations of the predictive distributions provide the same average test likelihood but different joint test likelihoods.
As the CLML evaluates the joint likelihood of held-out data $\dataset_{\ge m}$, it may not be aligned with the test likelihood.

The values of Eq.~\eqref{eq:app_joint} and Eq.~\eqref{eq:app_marginal} will be identical if $p(d_i \vert w_1) = p(d_i \vert w_2)$ for all $i \in \{1, 2\}$.
If each sample is tempered to optimize calibration, as in Figure \ref{fig:tempering}, then $p(d_i \vert w_j)$ should differ less across $w_j$, providing a higher correlation between the two measures.
Indeed if $p(d_i \vert w_1)\approx 0$ and $p(d_i \vert w_2)\approx 1$, the tempered distributions $p_T(d_i \vert w_1) > 0$ and $p_T(d_i \vert w_2)< 1$ would shift away from these extreme values, so that
and $|p_T(d_i \vert w_1) - p_T(d_i \vert w_2)| < |p(d_i \vert w_1) - p(d_i \vert w_2)|$.

\section{Extended Gaussian Process Results}
\label{sec:app_gp}

\begin{figure}[h]
\centering
    \begin{tabular}{cc}
        \includegraphics[height=0.25\textwidth]{figures/gp_rbf/hkl_exp.pdf}
        &
        \includegraphics[height=0.25\textwidth]{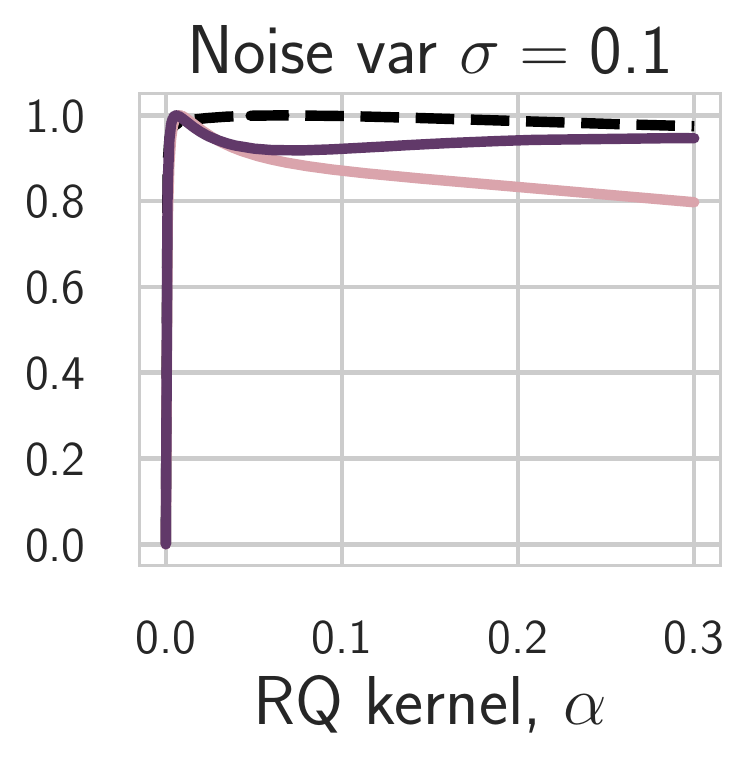}
        \includegraphics[height=0.25\textwidth]{figures/gp_rq/high_noise.pdf}
      \\
        {\small (a) Underfitting bias} & 
        {\small (b) Underfitting with the RQ kernel}
    \end{tabular}
\caption{
\textbf{LML for hyper-parameter tuning in Gaussian Processes.}
\textbf{(a):} The log-lengthscale learned by LML and CLML in a GP regression model averaged over $100$ datasets generated from a 
GP model with a lengthscale of $4$.
Unlike the train likelihood, LML has a bias towards underfitting, consistently overestimating the lengthscale, particularly for small $n < 20$.
\textbf{(b):} Test log-likelihood, LML and CLML as a function of the $\alpha$ hyper-parameter in the rational quadratic kernel.
While in panel with observation noise $\sigma=0.1$ the LML roughly captures the shape of the test likelihood curve,
with $\sigma=0.2$ the two curves are completely misaligned.}
\label{fig:gp_results}
\end{figure}

\textbf{GPs: RBF kernel.}\quad
In Figure \ref{fig:gp_results}(a) we illustrate the bias of LML towards underfitting.
We follow the experiment presented in \citet{wilson2015human}, and generate $100$ datasets from an RBF Gaussian process prior with a lengthscale of $l=4$.
The datapoints are located at positions $\{1, \ldots, 150\}$, the output scale is $1$, and the observation noise is $0.2$.
For each dataset and each $n \in \{1, \ldots, 150\}$, we fit a new GP model to the the first $n$ datapoints of the dataset:
we maximize the LML or CLML with respect to the lengthscale of the RBF kernel, using the ground truth value $l=4$ as the initialization.
We plot the learned lengthscales averaged over the datasets as a function of $n$ in \ref{fig:gp_results}(a).
This experiment illustrates a unique quality of marginal likelihood that distinguishes it from conventional maximum likelihood training:
while low lengthscales would lead to a better fit of the training data, marginal likelihood has a \textit{bias towards underfitting} in the data space.
Indeed, LML consistently selects lengthscales that are larger than the the lengthscale that was used to generate the data, especially for small $n$.
We note that CLML does not remove this bias, and provides a very similar curve.

\textbf{GPs: RQ kernel.}\quad
Above, we have seen how marginal likelihood can over-estimate the lengthscale of an RBF kernel leading to underfitting in data space.
Here, we construct a more extreme example of this behaviour using the rational quadratic (RQ) kernel (see \citet{rasmussen10gpml}):
$k_{RQ}(x_1, x_2) = \big ( 1 + \frac {\|x_1 - x_2\|^2}{2 \alpha l^2} \big )^{-\alpha}$.
The hyperparameters are the lengthscale $l$ and $\alpha$; lower values of $\alpha$ correspond to higher prior correlations, while as $\alpha \rightarrow \infty$ the kernel approaches the RBF kernel with lengthscale $l$.

We generate the data from a GP with an RQ kernel with hyperparameters $\hat \alpha = 0.05$, $\hat l = 0.5$, and observation noise standard deviation $\hat \sigma = 0.1$.
The dataset is shown in Appendix Figure \ref{fig:gp_rq_details}.
We then evaluate the LML and CLML as a function of $\alpha$ and compare them to the true BMA predictive likelihood of test data.
For this experiment we set the lengthscale $l = \hat l$ to its ground truth value, and we consider two values of the observation noise standard deviation: ground truth $\sigma = \hat \sigma = 0.1$ and over-estimated noise $\sigma = 2 \cdot \hat \sigma = 0.2$.
We show the results in Figure \ref{fig:gp_results}(b).
For the ground-truth noise scale, both LML and CLML provide an adequate representation of the test likelihood, although they both peak at a lower $\alpha$ value than the test likelihood surface.
However, for $\sigma=0.2$ the marginal likelihood is completely misaligned with the test log-likelihood: LML peaks at $\alpha \approx 0$, and then sharply decreases, while test LL is the lowest near $\alpha = 0$, and increases with $\alpha$.
The CLML does a much better job of tracking the test LL curve.

In Figure \ref{fig:gp_rq_details} (a), (b) we show the fit of the model with over-estimated observation noise $\sigma = 0.2$ for the $\alpha$ parameter chosen by maximizing the marginal likelihood and CML respectively.
For the CML, we condition on $m = 45$ datapoints (the training dataset size is $n = 50$), and we average the results over $20$ random orderings of the data.

In panel (c) of Figure \ref{fig:gp_rq_details} we show the learning curve averaged over $100$ random orderings of the data.
While for large $n$ the $\alpha=0.3$ model generalizes better, the small-$n$ terms in the marginal likelihood decomposition dominate, so that marginal likelihood prefers the simpler $\alpha=0.001$ model.

\begin{figure*}[t]
\centering
    \begin{tabular}{ccc}
       \includegraphics[height=0.17\textwidth]{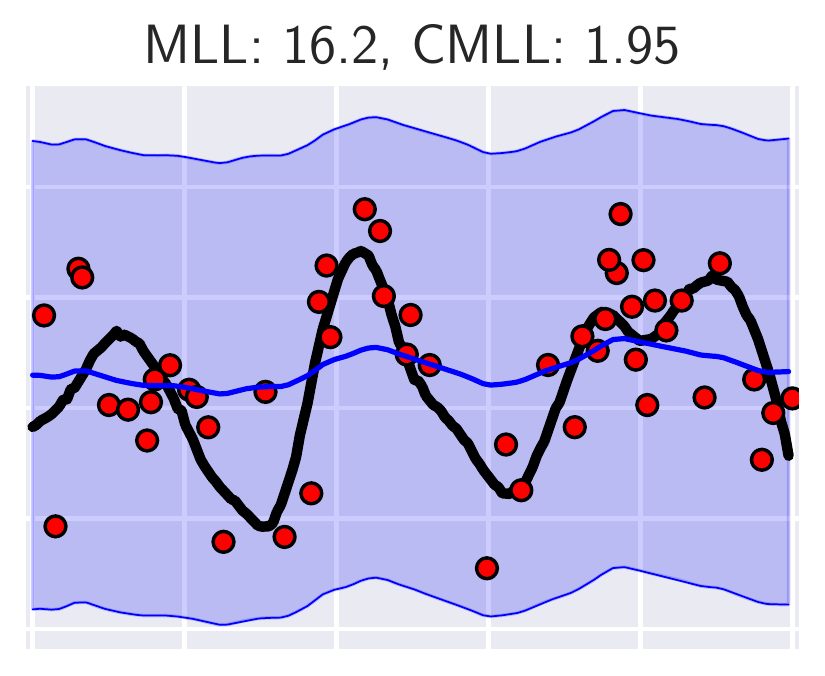}
        &
        \includegraphics[height=0.17\textwidth]{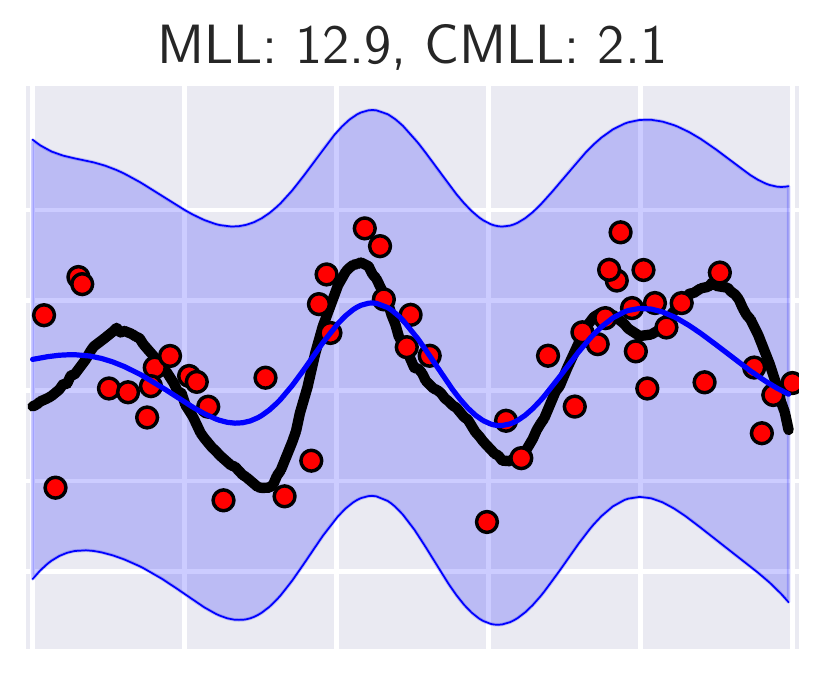}
        &
        \includegraphics[height=0.17\textwidth]{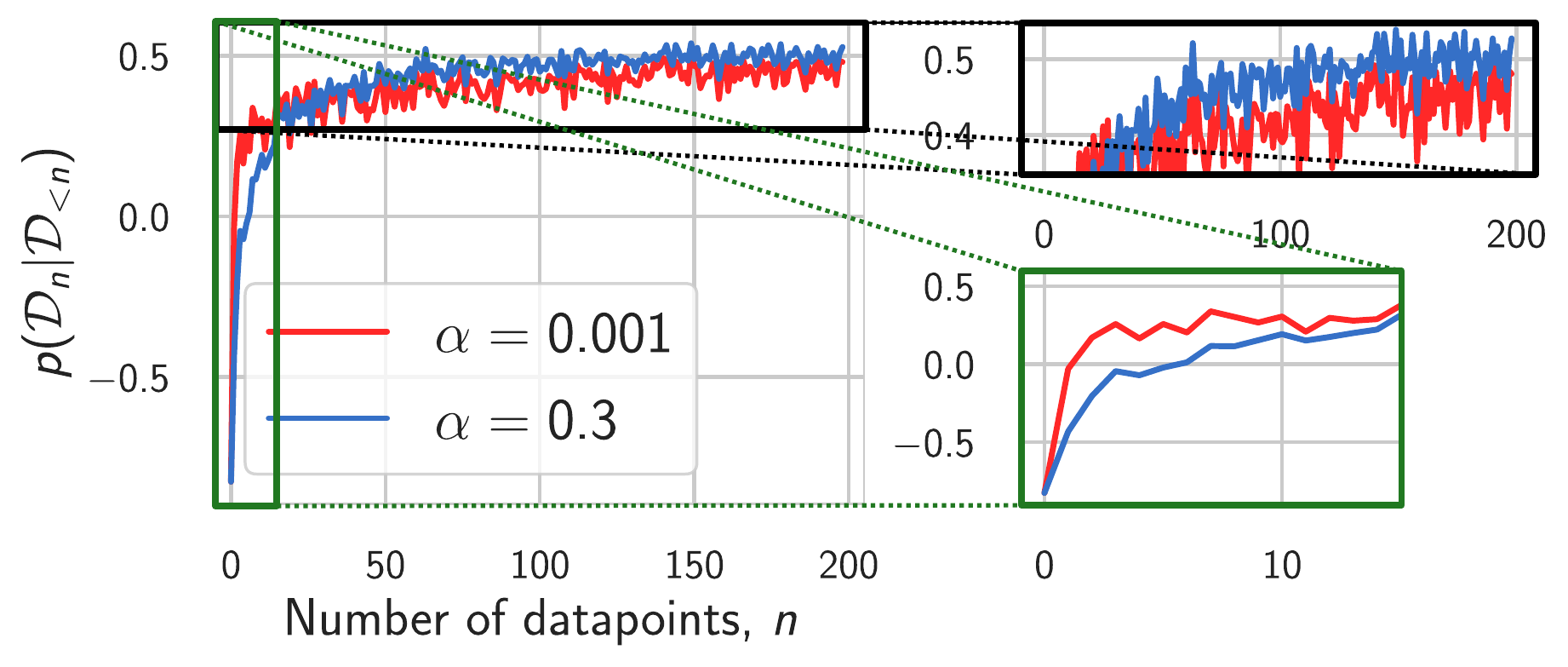}
      \\
        \hspace{-0.3cm}{\small (a) LML-optimized $\alpha$} & 
        \hspace{-0.cm}{\small (b)  CLML-optimized $\alpha$} &
        \hspace{-0.cm}{\small (c) Learning curves }
    \end{tabular}
\caption{
\textbf{Gaussian process: RQ kernel (details).}
The data fit and learning curves for the GP models with the RQ kernel with different $\alpha$ parameters.
\textbf{(a)}: The $\alpha$ parameter maximizing the marginal likelihood;
\textbf{(b)}: the $\alpha$ parameter maximizing the CML.
In each panel, the red dots show the training data, the black line shows the true latent function, the blue line shows the predictive mean, and the shaded region shows the $2\sigma$-region under the predictive distribution.
\textbf{(c)}: learning curves for small $\alpha$ providing the best marginal likelihood, and high $\alpha$ providing the best generalization.
While the larger $\alpha$ value generalizes better when the training set size is $n \ge 50$, marginal likelihood prefers small $\alpha$ values, as the model with a small $\alpha$ generalizes better on small datasets with $n \le 15$.
}
\label{fig:gp_rq_details}
\end{figure*}

\section{Deep Kernel Learning Details}\label{app:dkl}

\subsection{UCI Regression}

For the UCI regression datasets we use a DKL model with a fully-connected ReLU architecture of $[\mathcal{D}, 50, 50, 2]$, where $\mathcal{D}$ is the dimensionality of the data, and train using random subsets of the full UCI datasets ranging in size from $100$ to $700$ training points. We use the Bayesian Benchmarks library\footnote{\url{https://github.com/hughsalimbeni/bayesian_benchmarks}} to obtain the datasets, with a modification to ensure test data are not included in the normalization statistics.
Models are trained using the closed form LML and CLML forms known for Gaussian process regression.

\begin{figure}
    \centering
    \includegraphics[width=0.5\linewidth]{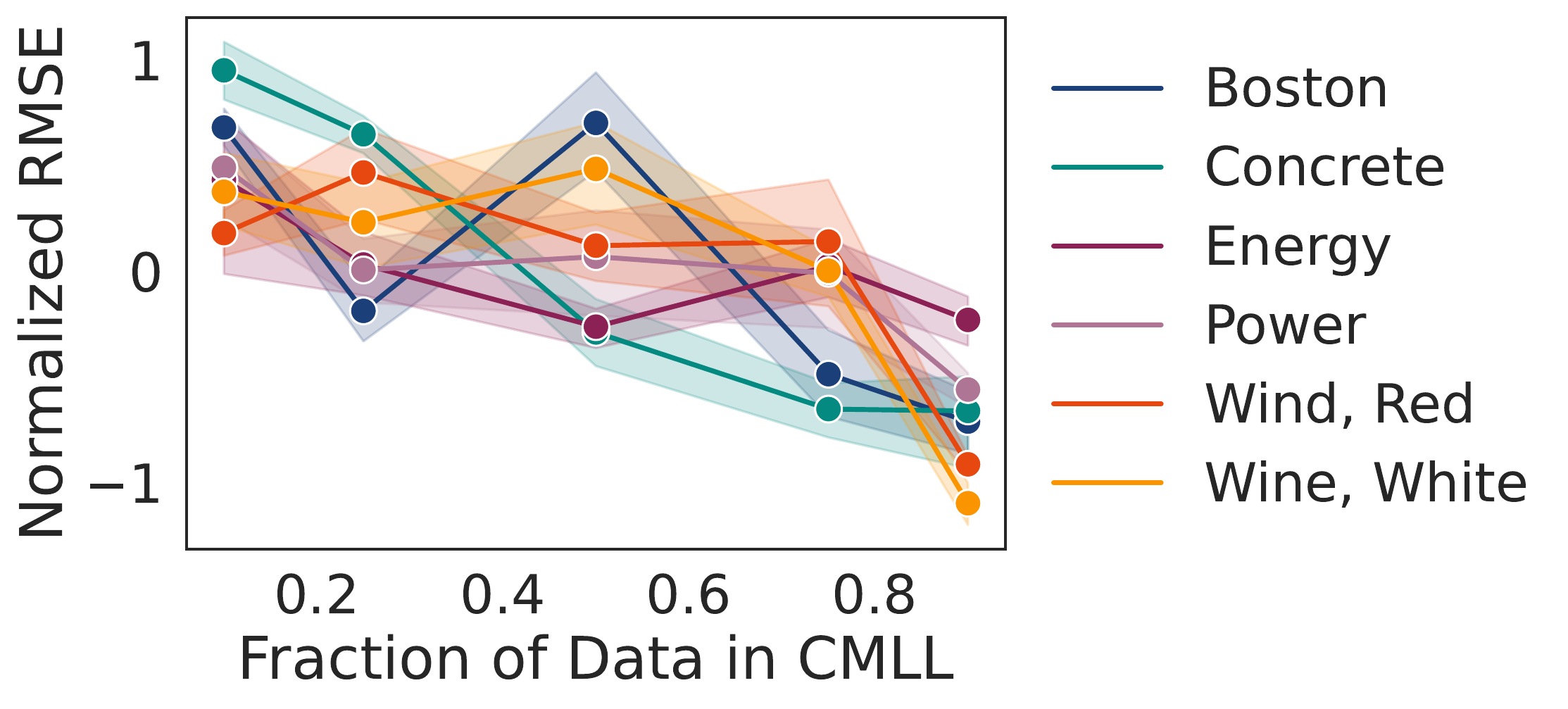}
    \caption{Effect of the parameter $m$ in the performance of DKL on several UCI regression datasets. The general trend is such that setting $m$ to be a larger, so that the CLML is conditioned on a larger fraction of the available training data, leads to better performance.}
    \label{fig:dkl-m}
\end{figure}

In figure \ref{fig:dkl-m} we show how the RMSE (normalized by dataset) varies for CLML optimization as a function of the fraction of data used used to condition the conditional marginal likelihood. As a general trend, the performance of CLML optimization increases as we use a larger fraction of the available data to condition on and a smaller fraction to compute the likelihood.

\begin{figure}
    \centering
    \begin{tabular}{cc}
        \hspace{-0.3cm}\includegraphics[width=0.4\linewidth]{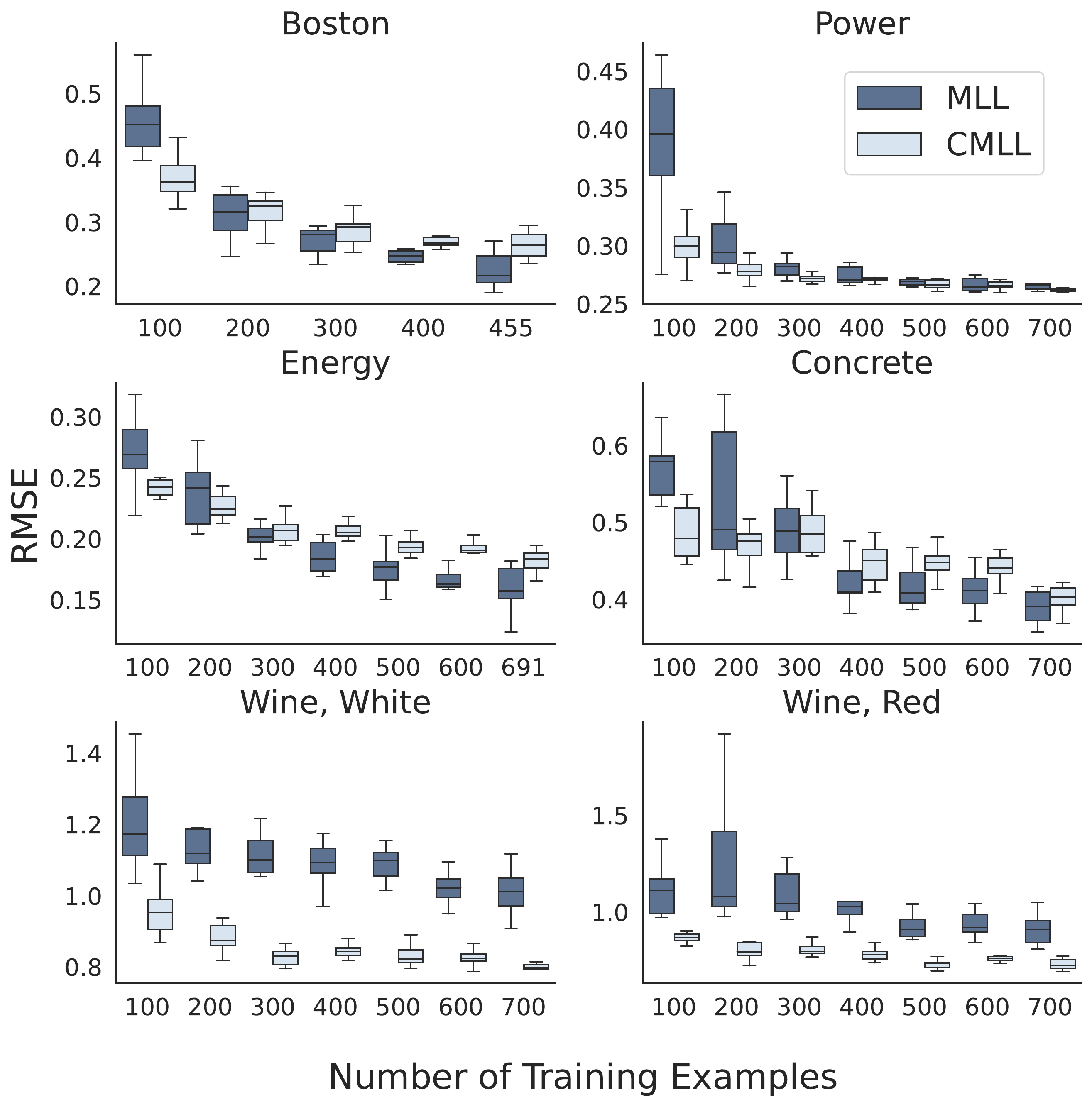}
        &
        \includegraphics[width=0.3\linewidth]{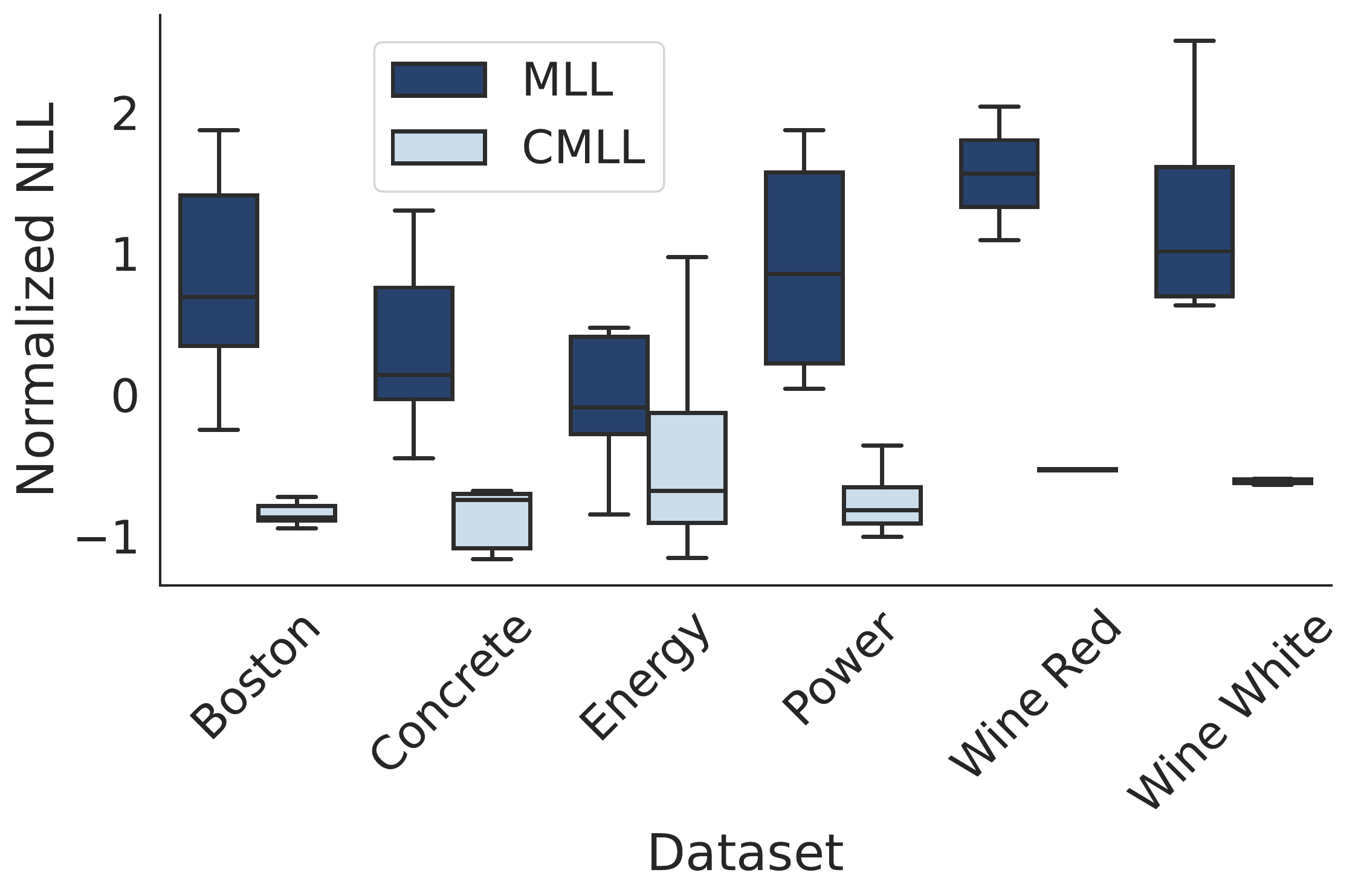}
    \end{tabular}
    \caption{
    \textbf{Left:} RMSE and \textbf{right:}
     NLL values normalized by dataset for DKL models trained on UCI regression tasks when trained with $N=300$ datapoints over $10$ independent initializations. For this training set size not only does CLML optimization lead to better accuracy on the test set, but also to better test likelihoods in limited data settings.}
    \label{fig:dkl-nlls}
\end{figure}

In Figure \ref{fig:dkl-nlls} we show the negative log likelihoods (normalized by dataset) of the $N=100$ models on the UCI regression problems. While there is some variance in the relative gap in performance between CLML and LML optimization, in all cases we see that for very restricted train set sizes CLML not only produces more accurate predictions, but is more performant in terms of NLL.

\subsection{DKT Transfer Learning}

Tables \ref{tbl:qmul} and \ref{tbl:omni-emnist} give the numerical results accompanying Figure \ref{fig:dkl_results}. From these tables we see that CLML optimization with the same model configuration consistently outperforms LML optimization, and in both experiments leads to the highest performing model. For full experimental details see \citet{patacchiola2020bayesian}.

\begin{table}
    \centering
    \begin{tabular}{c c c}
     Method & Model & MSE \\
     \hline
     \multirow{2}{*}{CLML} & DKT + RBF & $\mathbf{0.066  \pm  0.08}$ \\
       & DKT + Spectral  & $0.076 \pm 0.05$ \\
        \hline
      \multirow{2}{*}{LML} & DKT + RBF   &  $0.12\pm 0.04 $\\
       & DKT + Spectral   & $0.10\pm 0.01$ \\\hline
    \end{tabular}
    \caption{CLML and LML optimization of deep kernel transfer models on the QMUL head pose trajectory task of \citet{patacchiola2020bayesian}. In this limited data regime the focus on test performance of CLML leads to stronger performance.}
    \label{tbl:qmul}
\end{table}

\begin{table}
    \centering
    \begin{tabular}{c c c}
     Method & Model & Accuracy \\
     \hline
     \multirow{3}{*}{CLML} & DKT + CosSim & $75.34 \pm 0.35$ \\
       & DKT + BNCosSim  & $\mathbf{76.03 \pm 0.57}$ \\
       & DKT + Linear  & $75.64 \pm 0.38$ \\
        \hline
      \multirow{3}{*}{LML} & DKT + CosSim & $73.06 \pm 2.36$ \\
       & DKT + BNCosSim  & $75.06 \pm 1.10$ \\
       & DKT + Linear  & $75.97 \pm 0.70$ \\\hline
    \end{tabular}
    \caption{CLML and LML optimization of deep kernel transfer models on a transfer learning task in which the training data is from the Omniglot dataset, and the test data from the EMNIST dataset. In this transfer learning setting we should be more focused on the test performance, as opposed to the alignment of our model with the training data, which is a focus more closely aligned with the biases of CLML than LML optimization.}
    \label{tbl:omni-emnist}
\end{table}

\section{Choice of $m$ for the Conditional Marginal Likelihood}
\label{sec:ablation-clml}

The hyperparameter $m$ --- the number of datapoints that we condition on in CLML --- has an important effect on the conditional marginal likelihood.
Indeed, if we set $m=0$, we recover the marginal likelihood.
Setting $m=n-1$, we recover leave-one-out cross-validation likelihood for the BMA model, assuming we average the CLML over all possible orderings of the data.
Generally, we find that CLML works best for relatively large values of $m$.
However, setting $m << n$ (for example, in the architecture search experiments in Section \ref{sec:cifar-exps}) allows us to estimate CLML without averaging over multiple orderings.

In Figure \ref{fig:dkl-m} we show the effect of $m$ on the final RMSE for Deep Kernel Learning models trained with CLML.
We find that larger values of $m$ lead to better performance, but the results are relatively stable with respect to $m$.

\clearpage
\bibliography{sample}

\end{document}